\documentclass{article}

\usepackage[final, nonatbib]{neurips_data_2022}

\usepackage[utf8]{inputenc} 
\usepackage[T1]{fontenc}    
\usepackage{hyperref}       
\usepackage{url}            
\usepackage{booktabs}       
\usepackage{amsfonts}       
\usepackage{nicefrac}       
\usepackage{microtype}      
\usepackage{xcolor}         
\usepackage{amsmath,bm}
\usepackage{multirow}
\usepackage{array}
\usepackage{makecell}
\usepackage{graphicx}
\usepackage[small]{caption}
\usepackage{diagbox}
\usepackage{tabularx}
\usepackage{adjustbox}
\usepackage{colortbl}
\usepackage{bbding}
\usepackage{transparent}
\usepackage{subcaption}

\usepackage[noabbrev,nameinlink]{cleveref}
\crefname{property}{property}{Property}
\creflabelformat{property}{(#1)#2#3}
\creflabelformat{equation}{#1#2#3}

\newcolumntype{x}[1]{>{\centering\arraybackslash\hspace{0pt}}p{#1}}

\makeatletter
\newcommand{\thickhline}{%
    \noalign {\ifnum 0=`}\fi \hrule height 1.3pt
    \futurelet \reserved@a \@xhline
}

\title{\textsc{OpenFWI}: Large-scale Multi-structural \\ Benchmark Datasets for Full Waveform Inversion}

\author{%
  Chengyuan Deng$^{1, 2, }$\thanks{Equal contribution} \\
  \And 
  Shihang Feng$^{1, *}$\\
  \And 
  Hanchen Wang$^{1}$\\
  \And 
  Xitong Zhang$^{1, 3}$\\
  \And 
  Peng Jin$^{1,4}$\\
  \And 
  Yinan Feng$^{1}$\\
  \And 
  Qili Zeng$^{1}$\\
  \And 
  Yinpeng Chen$^{5}$\\
  \And 
  Youzuo Lin$^{1}$\\
   \And
\textnormal{$^1$Los Alamos National Laboratory \
$^2$Rutgers University \
$^3$Michigan State University} \\
$^4$The Pennsylvania State University \
$^5$Microsoft Research \\
\texttt{\{charles.deng, shihang, hanchen.wang, xitongz, pjin, ynf, ylin\}@lanl.gov}\\
\texttt{qili.zeng.cs@gmail.com, yichen@microsoft.com}
}

\begin{document}

\maketitle

\begin{abstract}
Full waveform inversion~(FWI) is widely used in geophysics to reconstruct high-resolution velocity maps from seismic data. The recent success of data-driven FWI methods results in a rapidly increasing demand for open datasets to serve the geophysics community. We present \textsc{OpenFWI}, a collection of large-scale multi-structural benchmark datasets, to facilitate diversified, rigorous, and reproducible research on FWI. In particular, \textsc{OpenFWI} consists of $12$ datasets ($2.1$TB in total) synthesized from multiple sources. It encompasses diverse domains in geophysics (interface, fault, CO$_2$ reservoir, etc.), covers different geological subsurface structures (flat, curve, etc.), and contains various amounts of data samples (2K - 67K). It also includes a dataset for 3D FWI. Moreover, we use~\textsc{OpenFWI} to perform benchmarking over four deep learning methods, covering both supervised and unsupervised learning regimes. Along with the benchmarks, we implement additional experiments, including physics-driven methods, complexity analysis, generalization study, uncertainty quantification, and so on, to sharpen our understanding of datasets and methods. The studies either provide valuable insights into the datasets and the performance, or uncover their current limitations. We hope \textsc{OpenFWI} supports prospective research on FWI and inspires future open-source efforts on AI for science. All datasets and related information (including codes) can be accessed through our website at \texttt{\url{https://openfwi-lanl.github.io/}}

\end{abstract}

\begin{figure*}[h!]
\centering
\includegraphics[width=1.0\columnwidth]{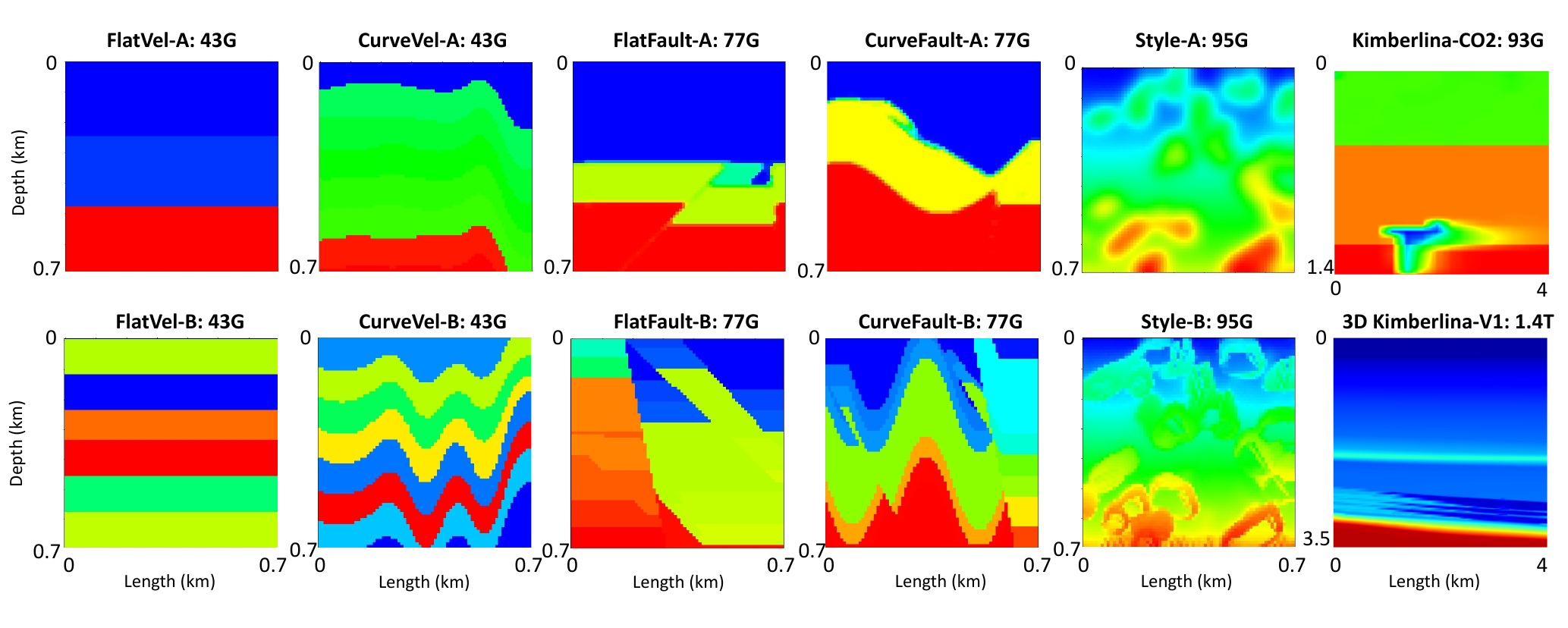}
\caption{\textbf{Gallery of \textsc{OpenFWI}}, which contains one example of velocity maps from each dataset in \textsc{OpenFWI}.}
\label{fig:gallery}
\end{figure*}

\section{Introduction}
\label{sec:Intro}
Understanding subsurface velocity structures is critical to a myriad of subsurface applications, such as carbon sequestration, reservoir identification, subsurface energy exploration, earthquake early warning, etc~\cite{Virieux-2009-Overview}. They can be reconstructed from seismic data with full waveform inversion~(FWI), which is governed by partial differential equations (PDEs) and can be formulated as a non-convex optimization problem. FWI has been intensively studied in the paradigm of \textit{physics-driven} approaches~\cite{fichtner2010full, zhang2012wave,ma2012image, zhang2013double,feng2019transmission+,feng2021mpi, lin2014acoustic, lin2015quantifying, hu2009simultaneous, guitton2012blocky,chen2020multiscale}. Negative complications of these approaches include high computation consumption, cycle-skipping, and ill-posedness issues. 

With the advance in deep learning techniques, researchers have been actively exploring data-driven solutions for complicated FWI problems~\cite{Deep-2021-Adler, willard2020integrating, zhu2019applications, mehta2019high, yang2019deep}. Recently, data-driven approaches have witnessed exploration for FWI, especially on network architectures such as multilayer perceptron (MLP)~\cite{araya2018deep, kim2018geophysical}, encoder-decoder based convolutional neural networks (CNNs)~\cite{yang2019deep, wu2019inversionnet, feng2021multiscale, wang2018velocity, feng2022intriguing}, recurrent networks~\cite{richardson2018seismic, fabien2020seismic, adler2019deep}, generative adversarial networks (GANs)~\cite{zhang2020data, wang2022seismic, mosser2020stochastic}, etc. \cite{zeng2021inversionnet3d} extended data-driven FWI from 2D to 3D. UPFWI \cite{Jin-2021-Unsupervised} leverages the governing acoustic wave equation to shift the learning paradigm from supervised to unsupervised.  \cite{adler2021deep} provides a detailed survey on purely deep learning-based FWI and \cite{Lin2022} gives a thorough overview of physics-guided data-driven FWI approaches.

Data is the oxygen for data-driven approaches, and public datasets figure prominently in developing cutting-edge machine learning algorithms. However, the FWI community currently experiences a lack of large public datasets. The existing seismic datasets~\cite{wang2020velocity,liu2021deep,araya2018deep,yang2019deep,ren2021building,geng2022deep} have not been released to the public. As a result, it is difficult to perform fair comparisons among different methods.

\begin{table}[h!]
\centering
\caption{\textbf{Existing datasets for data-driven FWI.} The top row corresponds to our~\textsc{OpenFWI} dataset. The
symbols \Checkmark and {\XSolidBrush} indicate that the dataset has or does not have the corresponding feature, respectively.} 
\vspace{0.5em}
\label{table:existing_dataset}
\begin{adjustbox}{width=1\textwidth}
\renewcommand{\arraystretch}{1.5}
\begin{tabular}{c|c|c|c|c|c|c|c|c|c} 
\hline
\multirow{2}{*}{Dataset}    & \multirow{2}{*}{Public} & \multirow{2}{*}{Multi-scale} & \multicolumn{2}{c|}{Domains} & \multicolumn{5}{c}{Geological Structures}                           \\ 
\cline{4-10}
                            &                         &                               & 2D & 3D                       & Interface & Fault & Salt body & CO$_2$~storage & Natural structure  \\ 
\hline
\textsc{OpenFWI}          & \Checkmark               & \Checkmark           & \Checkmark              & \Checkmark   & \Checkmark   & \Checkmark   & \transparent{0.2}{\XSolidBrush}   & \Checkmark   & \Checkmark  \\
Wang and Ma~\cite{wang2020velocity}                 & \transparent{0.2}{\XSolidBrush}                 & \transparent{0.2}{\XSolidBrush}           & \Checkmark              & \transparent{0.2}{\XSolidBrush} & \transparent{0.2}{\XSolidBrush} & \transparent{0.2}{\XSolidBrush} &  \transparent{0.2}{\XSolidBrush}  & \transparent{0.2}{\XSolidBrush} &   \Checkmark\\
Liu \textit{et al.}~\cite{liu2021deep}                 & \transparent{0.2}{\XSolidBrush}                  & \transparent{0.2}{\XSolidBrush}           & \Checkmark              & \transparent{0.2}{\XSolidBrush} & \Checkmark   & \Checkmark   & \Checkmark & \transparent{0.2}{\XSolidBrush} &   \transparent{0.2}{\XSolidBrush}  \\
Araya-Polo \textit{et al.}~\cite{araya2018deep}                 & \transparent{0.2}{\XSolidBrush}                   & \transparent{0.2}{\XSolidBrush}              & \Checkmark              & \transparent{0.2}{\XSolidBrush} & \Checkmark   & \transparent{0.2}{\XSolidBrush} &\Checkmark  & \transparent{0.2}{\XSolidBrush} &   \transparent{0.2}{\XSolidBrush}  \\
Yang and Ma~\cite{yang2019deep}                 & \transparent{0.2}{\XSolidBrush}                   & \transparent{0.2}{\XSolidBrush}             & \Checkmark              & \transparent{0.2}{\XSolidBrush} & \Checkmark   & \transparent{0.2}{\XSolidBrush} & \Checkmark  & \transparent{0.2}{\XSolidBrush} &   \transparent{0.2}{\XSolidBrush} \\
Ren \textit{et al.}~\cite{ren2021building}                 & \transparent{0.2}{\XSolidBrush}                 & \transparent{0.2}{\XSolidBrush}            & \transparent{0.2}{\XSolidBrush}            & \Checkmark   & \Checkmark   & \Checkmark   & \Checkmark & \transparent{0.2}{\XSolidBrush} &  \transparent{0.2}{\XSolidBrush}   \\
Geng \textit{et al.}~\cite{geng2022deep}                 & \transparent{0.2}{\XSolidBrush}                  & \transparent{0.2}{\XSolidBrush}             & \transparent{0.2}{\XSolidBrush}            & \Checkmark   & \Checkmark   & \Checkmark   & \transparent{0.2}{\XSolidBrush} & \transparent{0.2}{\XSolidBrush} & \transparent{0.2}{\XSolidBrush}  \\
\hline
\end{tabular}
\end{adjustbox}
\end{table}

Here, we present \textsc{OpenFWI}, the first large-scale collection of open-access multi-structural seismic FWI datasets based on our knowledge. It contains $12$ datasets, each pairs seismic data with velocity maps for different subsurface structures. Examples of velocity maps are shown in \Cref{fig:gallery}. A comparison between \textsc{OpenFWI} datasets and other existing datasets for data-driven FWI is listed in~\Cref{table:existing_dataset}. In contrast to previous datasets, our \textsc{OpenFWI} datasets are open-source, covering both 2D and 3D scenarios, capturing more geological structures on multiple scales. We emphasize our datasets have the following favorable characteristics:
\begin{itemize}
    \item \textit{Multi-scale: }\textsc{OpenFWI} covers multiple scales of datasets, in terms of the number of data samples and the file size. The smallest 2D dataset has $15$K data samples while the largest one contains $60$K samples. Four of the 2D datasets take $43$GB of space each, which supports training without massive computational power. The 3D dataset occupies $1.4$TB of space, therefore is usually trained in the distributed setting, further expediting the development of scalable algorithms for deep learning-based FWI. 
    
    \item \textit{Multi-domain: }\textsc{OpenFWI} empowers the research on both 2D and 3D scenarios of FWI. The datasets include velocity maps that are representative of realistic subsurface applications, such as time-lapse imaging, subsurface carbon sequestration, geologic faults detection, etc.
    \item \textit{Multi-subsurface-complexity: } \textsc{OpenFWI} encompasses a wide range of subsurface structures from simple to complex, such as interfaces, faults, CO$_2$ storages and natural structures from natural images. The complexity is primarily measured by Shannon entropy. 
    It supports researchers to start with moderate datasets and refine their methods for more challenging ones.
\end{itemize}

\begin{figure*}[t]
\centering
\includegraphics[width=1.0\columnwidth]{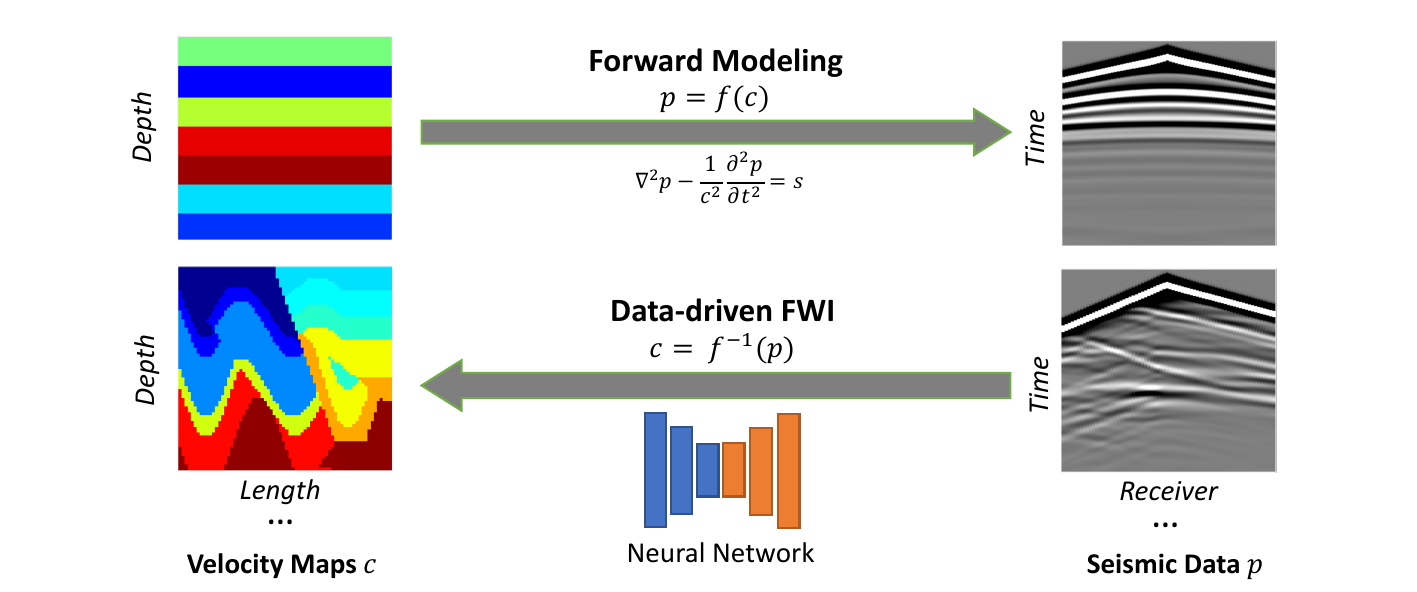}
\caption{\textbf{Schematic illustration of data-driven FWI and forward modeling.} Neural networks are employed to infer velocity maps from seismic data while forward modeling is to calculate the seismic data using governing wave equations with velocity map provided. }
\label{fig:fwi}
\end{figure*}

\textsc{OpenFWI} enables fair comparison among different methods over multiple datasets. We evaluate three representative methods (InversionNet~\cite{wu2019inversionnet}, VelocityGAN~\cite{zhang2020data}, and UPFWI~\cite{Jin-2021-Unsupervised}) over 2D datasets, and assess InversionNet3D~\cite{zeng2021inversionnet3d} on the 3D Kimberlina-V1 dataset. We hope these results provide a baseline for future work. For attempts on reproducibility, please refer to the resources listed in Section 1 of the supplementary materials, and the licenses therein.

\textsc{OpenFWI} also facilitates other related studies, such as complexity analysis, uncertainty quantification, generalization and so on. Limited by space, we briefly summarize the results of these studies and provide details in the supplementary materials. In particular, good generalizability is considered an important property of data-driven FWI, as a utopian method is expected to learn the physics rules of inversion, thus induces small errors when tested with unseen data. However, our empirical study shows existing methods suffer non-negligible degradation in terms of generalization, and it is related to the complexity of subsurface structures of the target datasets.

The rest of the paper is organized as follows: \Cref{sec:background} introduces the physics background of FWI. \Cref{sec:dataset} presents the datasets' properties concerned by domain interests. It follows in \Cref{sec:benchmark} to briefly introduce four deep learning methods for benchmarking, and demonstrate the inversion performance on each dataset. In \Cref{sec:discussion}, we initiate a discussion on the complexity of subsurface structure, the generalization performance, and uncertainty quantification, then move forward to future challenges. Finally, \Cref{sec:conclusion} concludes the paper.

\section{Seismic FWI and Forward Modeling}
\label{sec:background}
\Cref{fig:fwi} provides a concise illustration of 2D data-driven FWI and the relationship between velocity maps and the seismic data therein. The governing equation of the acoustic wave forward modeling in an isotropic medium with a constant density is as follows:
\begin{equation}
\nabla^2p-\frac{1}{c^2}\frac{\partial ^2p}{\partial t ^2}=s,
\label{Acoustic}
\end{equation}
where $\nabla^2=\frac{\partial ^2}{\partial x ^2}+\frac{\partial ^2}{\partial y ^2}+\frac{\partial ^2}{\partial z ^2}$, $c$ is velocity map, $p$ is pressure field and $s$ is source term. Velocity map $c$ depends on the spatial location $(x,y,z)$ while pressure field $p$ and source term $s$ depend on the spatial location and time $(x,y,z,t)$. In this study, we focus on controlled source methods, thus the source term $s$ is given. Forward modeling of acoustic wave propagation entails calculating the pressure field $p$ by \Cref{Acoustic} given velocity ${c}$. For simplicity, we denote the forward modeling problems expression as ${p} = f({c})$,

where $f(\cdot)$ represents the highly nonlinear forward mapping.
Data-driven FWI leverages neural networks to learn the inverse mapping as~\cite{adler2021deep}: ${c} = f^{-1}({p})$.

\section{\textsc{OpenFWI} Datasets and Domain Interests}
\label{sec:dataset}

\textsc{OpenFWI} datasets contain diverse subsurface structures covering multiple domains, thus supporting the study motivated by geophysics domain interests. The basic information and physical meaning of all datasets in \textsc{OpenFWI} is summarized in \Cref{tab:summary} and \Cref{tab:physics}, including 11 2D datasets and one for 3D FWI.\par 

The datasets are divided into four groups: ``\textit{Vel Family}'', ``\textit{Fault Family}'', ``\textit{Style Family}'' and ``\textit{Kimberlina Family}'',
to address five potential topics below. The first three families cover two versions: \textbf{easy (A)} and \textbf{hard (B)}, in terms of the complexity of subsurface structures. Details on the measurement of dataset complexity can be found in \Cref{sec:model-complexity}. 

Domain interests supported by \textsc{OpenFWI} datasets include:

 \begin{itemize}
     \item \textbf{Interfaces} that outline the subsurface structures and bound the velocity properties of rock layers~\cite{sen2006seismic}. To detect the interfaces, ``\textit{Vel Family}'' provides velocity maps comprised of flat and curved layers that have clear interfaces. The velocity value within the layers gradually increases with depth in version A and is randomly distributed in version B. 
     \item \textbf{Faults} caused by shifted rock layers can trap fluid hydrocarbons and form reservoirs~\cite{araya2017automated}. Fault detection is crucial for identifying, characterizing, and locating the reservoirs. ``\textit{Fault Family}'' includes discontinuity caused by the faults in the velocity maps, which enables the fault identification. Version B presents more discontinuities and severer velocity changes than version A. 
     \item \textbf{Field data} from different survey areas with high diversity and complexity, which have a significant effect on the inversion accuracy~\cite{li2021elastic}. ``\textit{Style Family}'' enriches the diversity of the dataset by generating the velocity maps from diversified natural images, which enables the inversion of field data in general cases. Version B has the high-resolution velocity maps while those in version A are smoothed by a Gaussian filter and the corresponding seismic data contains fewer events.
     \item \textbf{CO$_2$ storage}, one of the most promising methods to achieve significant reductions in atmospheric CO$_2$ emissions~\cite{lumley20104d} by injecting CO$_2$ into the reservoirs for long-term storage. The ``\textit{Kimberlina Family}'' has two datasets simulated with high fidelity through a geologic carbon sequestration (GCS) reservoir~\cite{wagoner20093d}. ``\textit{Kimberlina-CO$_2$}'' describes the spatial and temporal migration of the supercritical CO$_2$ plume within the reservoir, which is accompanied by timestamps within a time frame of $200$ years, and can be used for CO$_2$ storage problems, such as leakage detection and measurement.
     \item \textbf{3D seismic techniques} that attract increasing attention as 3D surveys have been widely implemented since~\cite{wang20193d}. The ``\textit{3D Kimberlina-V1}'' dataset is the first large-scale public 3D FWI dataset. It is generated by multiple institutions~\cite{Development-2021-Alumbaugh} and supported under the US Department of Energy (DOE)-SMART Initiative~\cite{SMART-2019-DOE}. It is designed and specified for the development of such techniques (not restricted to FWI). It contains a large amount of high-resolution 3D velocity maps and seismic data. 
\end{itemize}

Remarkably, the velocity maps are generated from three sources: math functions, natural images, and geological reservoirs. This property enhances the diversity and generality of the velocity maps significantly. The details of the velocity map and seismic data generation pipeline are elaborated in Section 2 and Section 3 of the supplementary materials, respectively. Moreover, we provide thorough instructions on the data format, loading, and all necessary information in Section 4 of the supplementary materials.

\renewcommand{\arraystretch}{1.25}
\begin{table}[ht]
\caption{\textbf{Dataset summary}. Explanation of data size: Velocity maps follow (depth $\times$ width $\times$ length); seismic data represents (\#source $\times$ time $\times$ \#receiver in width $\times$ \#receiver in length).}
\vspace{0.5em}
\small
\centering
\begin{tabularx}{\textwidth}{c|c|cccc}
\thickhline

Group & Dataset     & \multicolumn{1}{c}{Size}  & \multicolumn{1}{c}{\#Train/\#Test}  & \multicolumn{1}{c}{Seismic Data Size}   & Velocity Map Size \\ \thickhline

\multirow{2}{*}{\shortstack{Vel \\ Family}}  & FlatVel-A/B   & \multicolumn{1}{c}{$43$GB} & \multicolumn{1}{c}{$24$K / $6$K}    & \multicolumn{1}{c}{$5 \times 1000\times 1 \times 70$} & $70 \times 1 \times 70$   \\

 &  CurveVel-A/B  & \multicolumn{1}{c}{$43$GB} & \multicolumn{1}{c}{$24$K / $6$K}   & \multicolumn{1}{c}{$5 \times 1000 \times 1\times 70$} & $70 \times 1 \times 70$   
\\

\hline
 \multirow{2}{*}{\shortstack{Fault \\ Family}}  & FlatFault-A/B   & \multicolumn{1}{c}{$77$GB}  & \multicolumn{1}{c}{$48$K / $6$K} & \multicolumn{1}{c}{$5 \times 1000\times 1 \times 70$}  & $70 \times 1 \times 70$    \\

  & CurveFault-A/B & \multicolumn{1}{c}{$77$GB}  & \multicolumn{1}{c}{$48$K / $6$K} & \multicolumn{1}{c}{$5 \times 1000 \times 1\times 70$}  & $70 \times 1 \times 70$    \\

\hline
\multirow{1}{*}{ \shortstack{Style Family}}  & Style-A/B  & \multicolumn{1}{c}{$95$GB} & \multicolumn{1}{c}{$60$K / $7$K}  & \multicolumn{1}{c}{$5 \times 1000\times 1 \times 70$} & $70 \times 1 \times 70$  \\  

\hline
\multirow{2}{*}{ \shortstack{Kimberlina \\ Family}}   &  Kimberlina-CO$_2$   & \multicolumn{1}{c}{$96$GB}  & \multicolumn{1}{c}{$15$K / $4430$}   & \multicolumn{1}{c}{$9 \times 1251\times 1\times 101$}  & $141 \times 1 \times 401$    \\ 
  & \multirow{1}{*}[-2pt]{3D Kimberlina-V1}  & \multicolumn{1}{c}{$1.4$TB}      & \multicolumn{1}{c}{$1664$ / $163$}     & \multicolumn{1}{c}{$25\times5001\times40\times40$}      &   $ 350\times400\times400$        \\ \thickhline
\end{tabularx}

\label{tab:summary}
\vspace{-1em}
\end{table}

\begin{table}
\centering
\caption{\textbf{Physical Meaning of \textsc{OpenFWI} dataset}}
\vspace{0.5em}
\renewcommand{\arraystretch}{1.5}
\label{tab:physics}
\begin{adjustbox}{width=1\textwidth}
\begin{tabular}{c|c|c|c|c|c|c|c|c} 
\thickhline
Dataset                & \begin{tabular}[c]{@{}c@{}}Grid\\Spacing\end{tabular} & \begin{tabular}[c]{@{}c@{}}Velocity Map\\Spatial Size\end{tabular} & \begin{tabular}[c]{@{}c@{}}Source\\~Spacing\end{tabular} & \begin{tabular}[c]{@{}c@{}}Source Line\\Length\end{tabular} & \begin{tabular}[c]{@{}c@{}}Receiver Line\\Spacing\end{tabular} & \begin{tabular}[c]{@{}c@{}}Receiver Line\\Length\end{tabular} & \begin{tabular}[c]{@{}c@{}}Time\\~Spacing\end{tabular} &  \begin{tabular}[c]{@{}c@{}}Recorded\\Time\end{tabular}  \\ 
\thickhline
“Vel, Fault and Style” Family & 10 $m$                                                & 0.7 $\times$ 0.7 $km^2$                                           & 140 $m$                                                  & 0.7 $km$                                                    & 10 $m$                                                    & 0.7 $km$                                                 & 0.001 $s$                                                                                            & 1 $s$                                                   \\
Kimberlina-CO$_2$      & 10 $m$                                                & 1.4~$\times$~4~$km^2$                                             & 400 $m$                                                  & 3.6 $km$                                                    & 40 $m$                                                    & 4 $km$                                                   & 0.002~$s$                                                                                            & 2.5 $s$                                                 \\
3D Kimberlina-V1       & 10~$m$                                                & 3.5~$\times$~4 $\times $~4 $km^3$                                 & 800~$m$                                                  & (4 $km$, 4 $km$)                                            & 100 $m$                                                   & (4 $km$, 4 $km$)                                         & 0.001~$s$                                                                                             & 5 $s$                                                   \\
\thickhline
\end{tabular}
\end{adjustbox}
\end{table}

\section{\textsc{OpenFWI} Benchmarks}
\label{sec:benchmark}

\subsection{Deep Learning Methods for FWI}
We introduce four deep learning-based methods, InversionNet, VelocityGAN, and UPFWI for 2D FWI as well as InversionNet3D for 3D FWI, and report the inversion results as the initial benchmark. As mentioned above, UPFWI is an unsupervised learning method while the rest fall in the classical supervised learning regime. We provide a summary of each method separately as follows.

\textbf{InversionNet} \cite{wu2019inversionnet} proposed a fully-convolutional network to model the seismic inversion process. With the encoder and the decoder, the network was trained in a supervised scheme by taking 2D (time $\times$ \# of receivers) seismic data from multiple sources as the input and predicting 2D (depth $\times$ length) velocity maps as the output.

\textbf{VelocityGAN} \cite{zhang2020data} employed a GAN-based model to solve FWI. The generator is an encoder-decoder structure performing like the InversionNet, while the discriminator is a CNN designed to classify the real and fake velocity maps. It further used network-based deep transfer learning to improve the model's robustness and generalization.

\textbf{UPFWI} \cite{Jin-2021-Unsupervised} connected the forward modeling and a CNN in a loop to achieve unsupervised learning without the ground truth velocity maps for training. The velocity maps are predicted by CNN from the seismic data and then fed into the differentiable forward modeling to reconstruct the seismic data. Eventually, the loop is closed by calculating the loss between the input seismic data and the reconstructed ones.

\textbf{InversionNet3D} \cite{zeng2021inversionnet3d} extended InversionNet into 3D domain. In order to reduce the memory footprint and improve computational efficiency (i.e., two of the most challenging barriers in 3D inversion), the network utilized group convolution in the encoder and employed a partially reversible architecture via invertible layers based on additive coupling \cite{dinh2015a}.

\subsection{Inversion Benchmarks}
This section demonstrates the baseline results. We show the performance of three 2D deep learning methods in \Cref{table:OpenFWI_Benchmark2D} and InversionNet3D for 3D FWI separately in \Cref{tab:k3d-perf}. The network architectures of these methods and the hyper-parameters are provided in Section 5 of the supplementary materials. We consider three metrics: mean absolute error (MAE), rooted mean squared error (RMSE) and structural similarity (SSIM) \cite{wang2004image}. MAE and RMSE both capture the numerical difference between the predicted and true velocity maps. SSIM measures the perceptual similarity between two images.

\subsubsection{2D FWI Benchmarks}
\begin{figure*}[h!] 
\centering
\includegraphics[width=1.0\columnwidth]{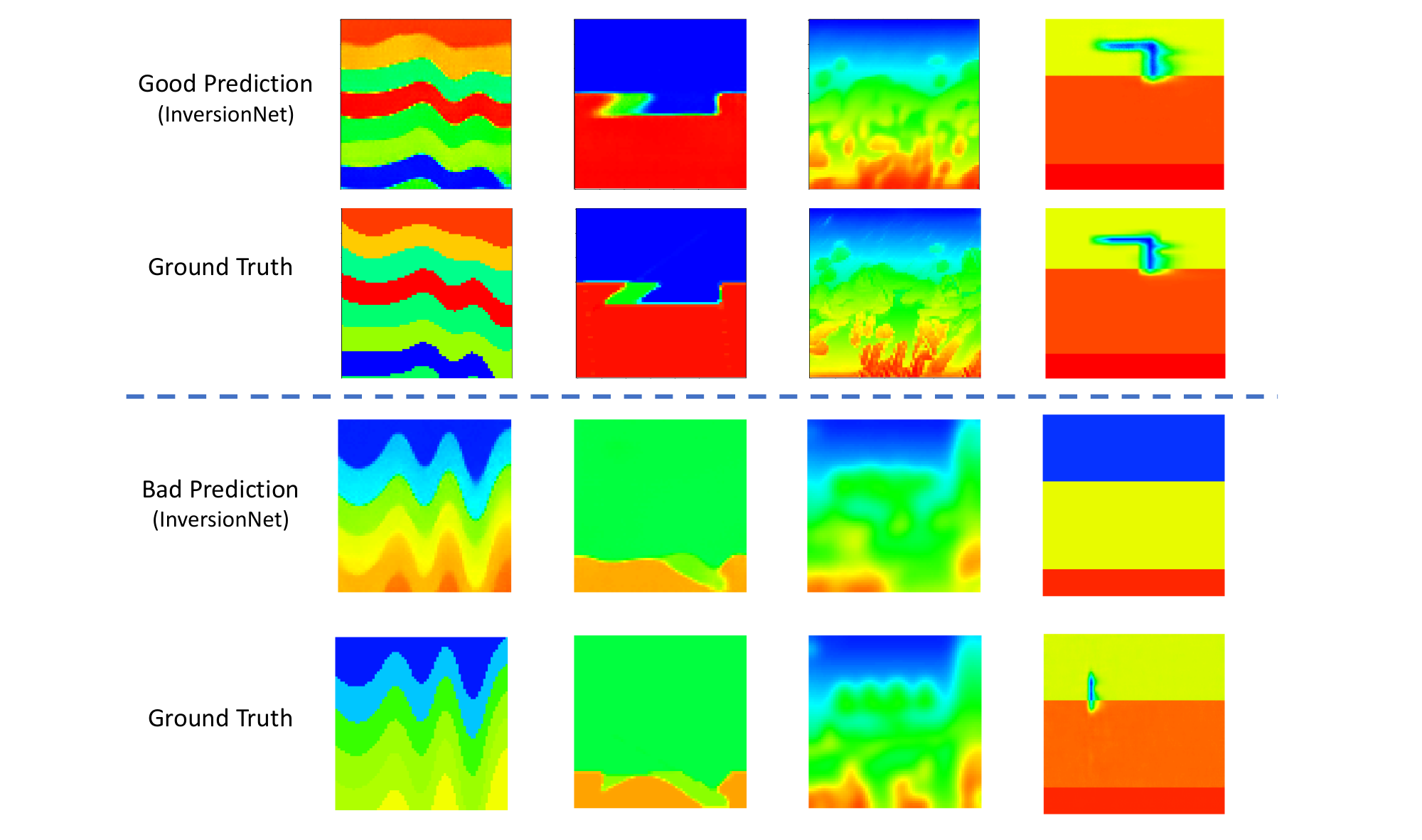}
\caption{First two rows: Illustration of good predicted velocity maps by InversionNet and ground truth on four datasets~(from left to right): CurveVel-B, FlatFault-A, Style-B, and Kimberlina-CO$_2$. Last two rows: Illustration of bad predicted velocity maps by InversionNet and ground truth on four datasets~(from left to right): CurveVel-A, CurveFault-A, Style-A, and Kimberlina-CO$_2$.}
\label{fig:2Dinv_maps}
\end{figure*}

 The training parameters are identical for all 2D datasets, and the model architecture only varies a little when training using the Kimberlina-CO$_2$ dataset, noticing that its data has different input and output shapes. Two most commonly used loss functions, $\ell_1$-norm and $\ell_2$-norm, are adopted as the metrics in InversionNet and VelocityGAN while UPFWI uses a combination of $\ell_1$-norm, $\ell_2$-norm and perceptual loss as in~\cite{Jin-2021-Unsupervised}. All the experiments are implemented on NVIDIA Tesla P100 GPUs. \Cref{table:OpenFWI_Benchmark2D} shows the inversion performance of three models on all 2D datasets, and \Cref{tab:time} shows the estimated training time by each method on \textsc{OpenFWI} datasets. Note that UPFWI is not evaluated on Kimberlina-CO$_2$ because of its high computational cost. The 
examples of inverted velocity maps and the ground truth are demonstrated in \Cref{fig:2Dinv_maps}, where we show both successful inversion results and those unpromising. The details of training configuration and more inversion results can be found in Section 6 and 7 of the supplementary materials, respectively.

From \Cref{table:OpenFWI_Benchmark2D}, we observe that all three methods perform well on simple datasets such as FlatVel-A and FlatFault-A. However, there exists considerable space for improvement on difficult datasets (CurveFault-B, Style-B, etc.). Notably, VelocityGAN outperforms InversionNet on the majority of datasets by a small margin and shows comparable results on the rest. It is worth mentioning that it would take much more training time for VelocityGAN to obtain better results than InversionNet. The performance of the UPFWI velocity maps is lower than the supervised methods to a small degree because of the limited frequency band in seismic data~\cite{schuster2017seismic}. The noticeable performance degradation for CurveFault-B indicates additional improvement on the UPFWI method would be needed.

\begin{table}
\footnotesize
\renewcommand{\arraystretch}{1.3}
\centering
\caption{\textbf{Quantitative results} of three benchmarking methods on 2D FWI datasets. }
\vspace{0.5em}
\begin{adjustbox}{width=1\textwidth}
\begin{tabular}{c|cc|ccc|ccc|ccc} 
\thickhline
\multirow{2}{*}{Dataset}                                                      & \multicolumn{2}{c|}{\multirow{2}{*}{Loss  }}                  & \multicolumn{3}{c}{InversionNet}                     & \multicolumn{3}{c}{VelocityGAN}                      & \multicolumn{3}{c}{UPFWI}                                                    \\ 
\cline{4-12}
                                                                              & \multicolumn{2}{c|}{}                                         & MAE$\downarrow$ & RMSE$\downarrow$ & SSIM$\uparrow$  & MAE$\downarrow$ & RMSE$\downarrow$ & SSIM$\uparrow$  & MAE$\downarrow$         & RMSE$\downarrow$        & SSIM$\uparrow$           \\ 
\thickhline
\multirow{2}{*}{FlatVel-A}                                                    & \begin{tabular}[c]{@{}c@{}}$\ell_1$\\\end{tabular} &          & 0.0131          & 0.0211           & \textbf{0.9895} & 0.0118          & 0.0178           & \textbf{0.9916} & \multirow{2}{*}{0.0621} & \multirow{2}{*}{0.1233} & \multirow{2}{*}{\textbf{0.9565}}  \\
                                                                              &                                                    & $\ell_2$ & 0.0111          & 0.0180           & 0.9887          & 0.0605          & 0.0783           & 0.9453          &                         &                         &                          \\ 
\hline
\multirow{2}{*}{FlatVel-B}                                                    & $\ell_1$                                           &          & 0.0351          & 0.0876           & \textbf{0.9461} & 0.0329          & 0.0807           & 0.9521          & \multirow{2}{*}{0.0677} & \multirow{2}{*}{0.1493} & \multirow{2}{*}{\textbf{0.8874}}  \\
                                                                              &                                                    & $\ell_2$ & 0.0417          & 0.0909           & 0.9402          & 0.0328          & 0.0787           & \textbf{0.9556} &                         &                         &                          \\ 
\hline
\multirow{2}{*}{CurveVel-A}                                                   & $\ell_1$                                           &          & 0.0685          & 0.1273           & 0.8074          & 0.0482          & 0.1034           & 0.8624          & \multirow{2}{*}{0.0805} & \multirow{2}{*}{0.1411} & \multirow{2}{*}{\textbf{0.8443}}  \\
                                                                              &                                                    & $\ell_2$ & 0.0690          & 0.1202           & \textbf{0.8223} & 0.0510          & 0.0976           & \textbf{0.8758} &                         &                         &                          \\ 
\hline
\multirow{2}{*}{CurveVel-B}                                                   & $\ell_1$                                           &          & 0.1497          & 0.2891           & \textbf{0.6727}          & 0.1268          & 0.2618           & \textbf{0.7111} & \multirow{2}{*}{0.1777} & \multirow{2}{*}{0.3179} & \multirow{2}{*}{\textbf{0.6614}}  \\
                                                                              &                                                    & $\ell_2$ & 0.1624          & 0.2801           & 0.6661 & 0.1428          & 0.2611           & 0.6962          &                         &                         &                          \\ 
\hline
\multirow{2}{*}{FlatFault-A}                                                  & $\ell_1$                                           &          & 0.0172          & 0.0426           & {0.9766}          & 0.0868          & 0.1485           & 0.9313          & \multirow{2}{*}{0.0876} & \multirow{2}{*}{0.2060} & \multirow{2}{*}{\textbf{0.9340}}  \\
                                                                              &                                                    & $\ell_2$ & 0.0174          & 0.0362           & \textbf{0.9798} & 0.0319          & 0.0531           & \textbf{0.9798} &                         &                         &                          \\ 
\hline
\multirow{2}{*}{FlatFault-B}                                                  & $\ell_1$                                           &          & 0.1055          & 0.1741           & \textbf{0.7208} & 0.0925          & 0.1600           & 0.7476          & \multirow{2}{*}{0.1416} & \multirow{2}{*}{0.2220} & \multirow{2}{*}{\textbf{0.6937}}  \\
                                                                              &                                                    & $\ell_2$ & 0.1106          & 0.1723           & 0.7186          & 0.0946          & 0.1553           & \textbf{0.7552} &                         &                         &                          \\ 
\hline
\multirow{2}{*}{CurveFault-A}                                                 & $\ell_1$                                           &          & 0.0260          & 0.0650           & 0.9566          & 0.0258          & 0.0606           & 0.9613          & \multirow{2}{*}{0.0500} & \multirow{2}{*}{0.0966} & \multirow{2}{*}{\textbf{0.9495}}  \\
                                                                              &                                                    & $\ell_2$ & 0.0280          & 0.0602           & \textbf{0.9592} & 0.0216          & 0.0505           & \textbf{0.9687} &                         &                         &                          \\ 
\hline
\multirow{2}{*}{CurveFault-B}                                                 & $\ell_1$                                           &          & 0.1646          & 0.2477           & \textbf{0.6163}          & 0.1571          & 0.2427           & 0.5996          & \multirow{2}{*}{0.3452}      & \multirow{2}{*}{0.5010}      & \multirow{2}{*}{\textbf{0.3941}}       \\
                                                                              &                                                    & $\ell_2$ & 0.1669          & 0.2412           & 0.6053 & 0.1583          & 0.2336           & \textbf{0.6033} &                         &                         &                          \\ 
\hline
\multirow{2}{*}{Style-A}                                                      & $\ell_1$                                           &          & 0.0625          & 0.1024           & 0.8859          & 0.0612          & 0.1000           & \textbf{0.8883} & \multirow{2}{*}{0.1429}      & \multirow{2}{*}{0.2342}      & \multirow{2}{*}{\textbf{0.7846}}  \\
                                                                              &                                                    & $\ell_2$ & 0.0610          & 0.0989           & \textbf{0.8910} & 0.0645          & 0.1025           & 0.8882          &                         &                         &                          \\ 
\hline
\multirow{2}{*}{Style-B}                                                      & $\ell_1$                                           &          & 0.0689          & 0.1614           & 0.6314          & 0.0697          & 0.1108           & 0.6953          & \multirow{2}{*}{0.1702}      & \multirow{2}{*}{0.2609}      & \multirow{2}{*}{\textbf{0.6102}}  \\
                                                                              &                                                    & $\ell_2$ & 0.0586          & 0.0893           & \textbf{0.7599} & 0.0649          & 0.0979           & \textbf{0.7249} &                         &                         &                          \\ 
\hline
\multirow{2}{*}{\begin{tabular}[c]{@{}c@{}}Kimberlina-CO$_2$\end{tabular}} & $\ell_1$                                           &          & 0.0061          & 0.0374           & \textbf{0.9872} & 0.0122          & 0.0574           & \textbf{0.9716} & \multirow{2}{*}{$\backslash$}      & \multirow{2}{*}{$\backslash$}      & \multirow{2}{*}{$\backslash$}       \\
                                                                              &                                                    & $\ell_2$ & 0.0098          & 0.0400           & 0.9798          & 0.0119          & 0.0387           & 0.9527          &                         &                         &                          \\
\thickhline
\end{tabular}
\end{adjustbox}
\vspace{0.5em}
\label{table:OpenFWI_Benchmark2D}
\vspace{-1em}
\end{table}

\begin{table}
\centering
\caption{\textbf{Training time} by each benchmarking method on \textsc{OpenFWI} datasets. Notice that the training of UPFWI and InversionNet3D occupied 32 GPUs, the rest used a single GPU.}
    \vspace{0.5em}
\begin{adjustbox}{width=1.0\textwidth}
\label{tab:time}
\begin{tabular}{c|ccccc} 
\thickhline
             & Vel Family & Fault Family & Style Family & Kimberlina-CO$_2$ & 3D Kimberlina-V1  \\ 
\hline
InversionNet & 2h         & 4h           & 5.5h         & 3.5h              & 5.5h              \\
VelocityGAN  & 8.6h           & 16h             & 30h             & 32h               & N.A.              \\
UPFWI        & 30h        & 60h          & 60h          & N.A.              & N.A.              \\
\thickhline
\end{tabular}
\end{adjustbox}
\end{table}

\subsubsection{3D FWI Benchmarks}

\begin{figure*}[h!]
\begin{adjustbox}{width=1\textwidth}
\begin{minipage}{0.3\linewidth}
    \centering
    \vspace{20pt}
    \includegraphics[height=3.5cm]{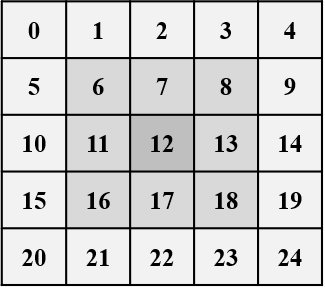}
    \captionof{figure}{\textbf{Spatial Placement of Sources.} Each source is the input seismic data of one channel.} 
    \label{fig:k3d-sources}
\end{minipage} 
\hspace{10pt}
\begin{minipage}{0.70\linewidth}
    \centering
    \renewcommand{\arraystretch}{1.5}
    \captionof{table}{\textbf{Quantitative results} of InversionNet3D on 3D Kimberlina-V1 dataset with different channel selection strategies of seismic input. }
    \begin{adjustbox}{width=1\textwidth}
    \begin{tabular}{c|l|c|c|c}
    \hline
    Training Loss                             & \multicolumn{1}{c|}{Selected Channels}     & MAE $\downarrow$            & RMSE $\downarrow$           & SSIM $\uparrow$           \\ \hline
    \multirow{3}{*}{$\ell_1$} & {[}1,  2,  14,  15,  16,  20,  23,  24{]}  & 0.0108	&	0.0286 &	0.9838          \\
                                      & {[}6, 7, 8, 11, 13, 16, 17, 18{]}          & 0.0105	& 	0.0276 &	0.9838          \\
                                      & {[}0, 2, 4, 10, 14, 20, 22, 24{]}          & 0.0107 &		0.0282 &	0.9835          \\ \cline{1-5} 
    \multirow{3}{*}{$\ell_2$} & {[}1,  2,  14,  15,  16,  20,  23,  24{]}  & 0.0154	&	0.0306 &	0.9482          \\
                                      & {[}6, 7, 8, 11, 13, 16, 17, 18{]}          & 0.0152	&	0.0302 &	0.9476          \\
                                      & {[}0, 2, 4, 10, 14, 20, 22, 24{]}          & 0.0158	&	0.0312 &	0.9427          \\ \cline{1-5} 
                                    
    \end{tabular}
    \end{adjustbox}
    \label{tab:k3d-perf}
\end{minipage}%
\vspace{0.5em}
\end{adjustbox}
\end{figure*}

Kimberlina 3D-V1 is a recently generated experimental dataset, on which only the performance of InversionNet3D \cite{zeng2021inversionnet3d} has been reported. In \Cref{tab:k3d-perf} we include the performance of InversionNet3Dx1, the shallowest version of the network, on three-channel distributions, one of which is randomly selected and the other two are symmetrical.~\Cref{fig:k3d-sources} explains the serial number allocation of $25$ sources (channels) in the seismic data. Compared to $\ell_1$ loss, $\ell_2$ loss leads to a degradation on SSIM of 3\%. More details and analysis can be found in \cite{zeng2021inversionnet3d}.

\section{Ablation Study}

In this section, we conduct intensive ablation studies including subsurface complexity analysis, generalization test, and uncertainty quantification. Each study brings insights on sharpening our understanding of \textsc{OpenFWI}. Moreover, We discover the current limitation of generalizability is closely related to the subsurface complexity. Limited by space, other additional experiments are described in the supplementary materials.

\subsection{Velocity Map Complexity Analysis}
\label{sec:model-complexity}
Recall that the first step of data generation is to synthesize velocity maps from different priors, simulating various geological subsurface structures (interfaces, layers, faults, etc). Therefore, the velocity maps encompass different levels of complexity. We employ three standard metrics: Shannon entropy, spatial information, and gradient sparsity index to compare the relative model complexity of all 2D datasets. The spatial information captures the average boundary magnitude, and the gradient sparsity index measures the percentage of non-smooth pixels. Their math formulation is presented in Section 8 of the supplementary materials, where we also include numerical results and illustrations. \par

\begin{figure*}[h!] 
\centering
\includegraphics[width=1.0\columnwidth]{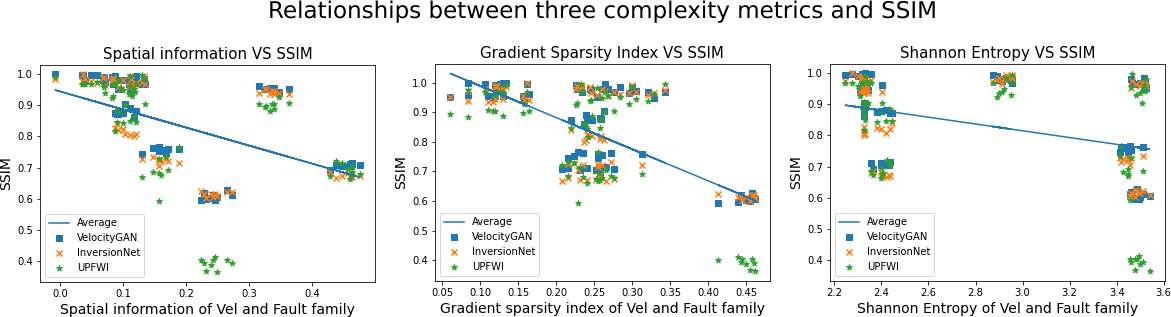}
\caption{From left to right: three complexity metrics (spatial information, gradient sparsity index, Shannon entropy) versus SSIM. Three 2D benchmark methods (InversionNet, VelocityGAN and UPFWI) are colored in blue, orange and green, respectively. The blue line is obtained from the linear regression on the average SSIM.}
\label{fig:complexity}
\end{figure*}

Our aim is to explore the connection between geological subsurface and performance. Therefore we demonstrate their relationship with three complexity metrics and the SSIM of three 2D benchmark methods on eight datasets in the Vel and Fault family. The reason for selecting these two families is that they follow the same generation strategy. The scatter plots and the line plots obtained from linear regression can be found in \Cref{fig:complexity}, which indicates that the inversion performance is negatively related to the velocity map complexity, corresponding to the numerical results in \Cref{table:OpenFWI_Benchmark2D}. The conclusion is not surprising due to a straightforward intuition: complex velocity maps should be more difficult to be inverted from the seismic data.\par

\subsection{Generalization Study}
\begin{figure*}[h]
\centering
\includegraphics[width=0.9\columnwidth]{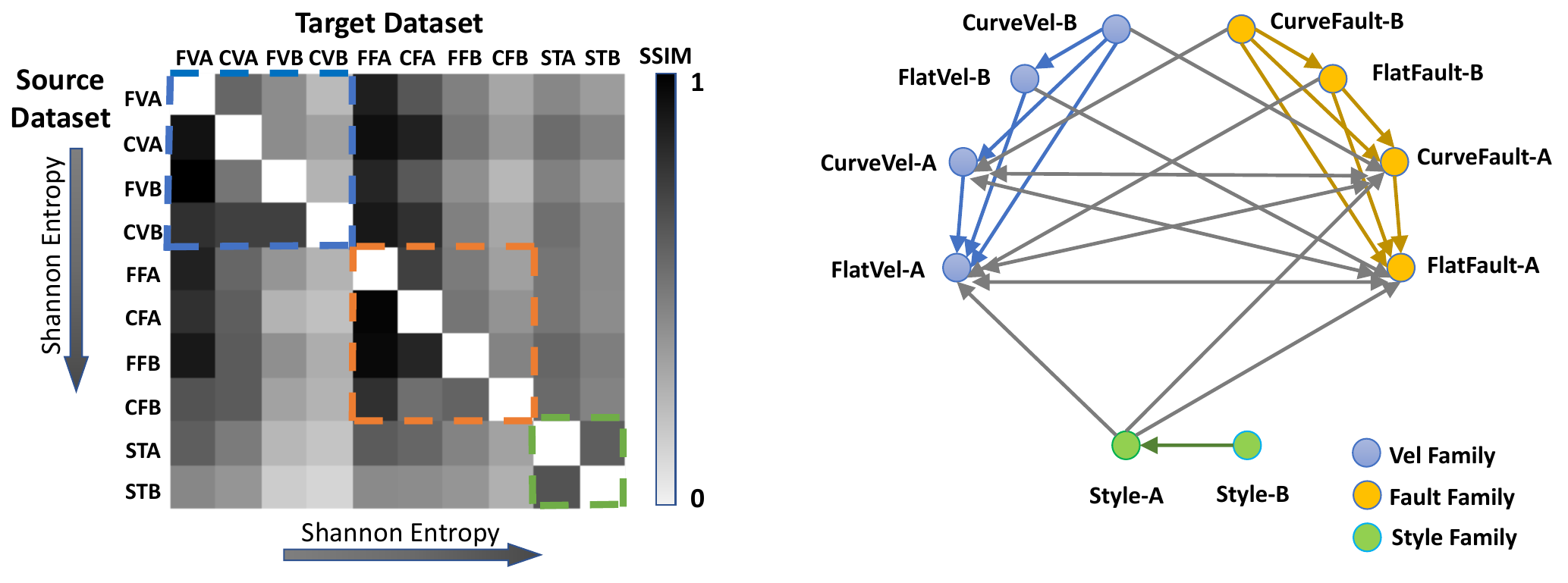}
\caption{\textbf{Heatmap~(Left) and graph (Right) of the generalization performance~.} ``FVA'' is short for ``FlatVel-A'', same applies to the rest datasets. The arrow ``$X\rightarrow Y$'' implies the SSIM metric is above $0.6$ for model trained on $X$ and tested on $Y$.}
\label{fig:gen_test}
\vspace{-1em}
\end{figure*}

We perform pair-wise generalization tests across $10$ datasets in the ``\textit{Vel}'', ``\textit{Fault}'' and ``\textit{Style}'' families. Specifically, we select the best-trained models by VelocityGAN on each dataset (\cite{zhang2020data} claims that it shows better generalization results than InversionNet) and tested with the rest $9$ datasets. The generalization performance is measured by the SSIM metric, and we obtain a $10 \times 10$ matrix illustrated in the heatmap of \Cref{fig:gen_test}, darker color indicates better generalization. We extract the relationship between these ten datasets based on the generalization performance, shown on the right of \Cref{fig:gen_test}. The results are analyzed in two-fold: \textit{intra-domain} and \textit{cross-domain}.

\textbf{Intra-domain:} Focusing on the 3 diagonal blocks on the heatmap (enclosed with dashed rectangles of different colors for each data family) in \Cref{fig:gen_test}, we observe that the lower-triangle entries always have larger values than those in the upper triangle, implying that generalization from harder datasets to easier ones is more promising than the other way.

\textbf{Cross-domain:} When the source dataset is fixed, the generalization drops on the target dataset as the complexity increases. From the graph, we also observe that the degree of nodes on datasets with ``A'' is always higher than those with ``B''. The Style-B dataset has no incoming or outcoming edges to datasets in other families, thus can be regarded as a challenging dataset for generalization. More discussions on the generalization study are given in Section 9 of the supplementary materials.

\subsection{Uncertainty Quantification}
\begin{figure*}[t]
\centering
\includegraphics[width=1.0\textwidth]{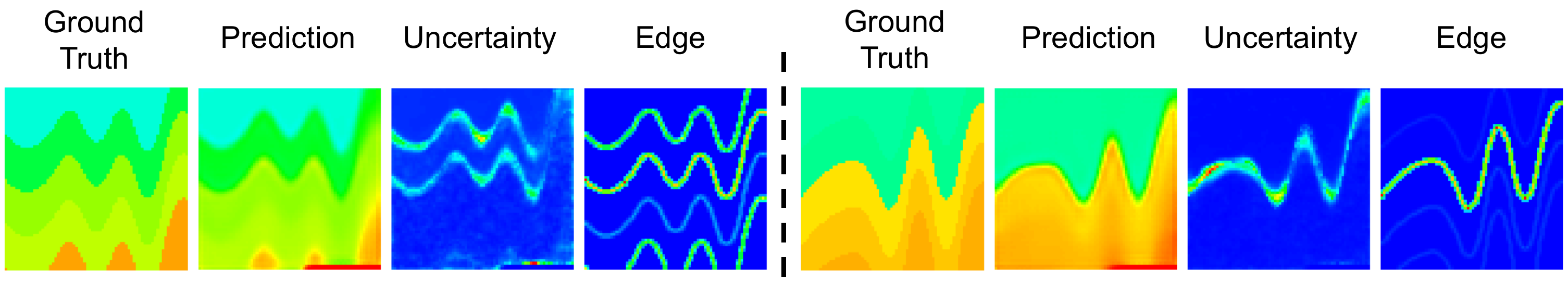}
\caption{\textbf{Uncertainty visualization}. The uncertainty is higher on the boundaries compared with other regions.}
\label{fig:uncertainty1}
\end{figure*}

We conduct experiments on CurveVel-A to quantify uncertainty in InversionNet as a case study. As shown in Figure~\ref{fig:uncertainty1}, the uncertainty on boundaries is higher than in other regions, which implies the prediction around the boundaries is more sensitive. We also observe that the uncertainty increases gradually when increasing the noise levels. Moreover, the uncertainty values of cross datasets are much higher than training and testing on the same dataset, which indicates that domain shifts lead to an increase in uncertainty. Experiment details and more results are provided in Section 10 of the supplementary materials.

\subsection{Additional Experiments}
We have conducted more experiments including the robustness test, the comparison between physics-driven methods and data-driven methods, the comparison between InversionNet and InversionNet3D and a demonstration of choosing a dataset for the target in the real scenario. All above tasks answer for major concerns in the data-driven FWI community. Limited by the space, we briefly present our findings from these experiments, more details are provided in Section 11, 12, 13, and 14 of supplementary material, respectively.

\begin{itemize}
    \item \textbf{Robustness test}: Models are trained on 2D clean datasets but tested on noisy seismic data over multiple noise levels. Not surprisingly, degradation appears as the noise increases. We also find InversionNet is the most sensitive model.
    \item \textbf{Comparison between data-driven methods and physics-driven methods}: We compare two methods with respect to accuracy and computational cost. The inversion results of Data-driven methods are better by a large margin, and faster when the ratio between the number of training and test samples is less than 62.
    \item \textbf{Comparison between 3D simulation and 2D slices}: We train an InversionNet with 2D velocity/seismic data slices of the 3D Kimberlina-V1 dataset and compare with the InversoinNet3D benchmark. The results are comparable, though InversionNet3D slightly performs better (0.9652 compared to 0.9838).
    \item \textbf{Choosing a dataset in the real scenario}: We choose a real velocity map in~\cite{huang2018full} and generate its seismic data, then apply all twenty models trained across ten \textsc{OpenFWI} datasets. Only for this case, the best model trained using $\ell_1$ loss
 is from the FlatVel-B dataset and the best model trained using $\ell_2$ loss
 is from the Style-A dataset.  
\end{itemize}

\section{Discussion}
\label{sec:discussion}

\subsection{Future Challenges}
In light of the results demonstrated so far, we envisage four future challenges for data-driven FWI as listed below, where \textsc{OpenFWI} should be able to empower the related studies.

\textbf{Inversion for complex velocity maps:} The deteriorated performance on datasets with high subsurface complexity necessitates more advanced methods, especially those without reliance on more data.

\textbf{Generalization of data-driven methods:} The field data is usually different from the training dataset and thus good generalization is crucial for the data-driven FWI in field applications. However, the existing methods suffer non-negligible degradation on generalization. We expect more robust methods to handle data from different domains.

\textbf{Computational efficiency:} Based on our experience, UPFWI and InversionNet3D suffer from the high computational cost, which limits their potential applications. Especially for InversionNet3D, the training data is down-sampled with several channels, which may lead to the loss of information and affect its performance. More efficient algorithms are expected for these directions.

\textbf{Passive seismic imaging:} The benchmark results in this paper mainly cover the controlled/active seismic source imaging problems, but passive seismic problems is also a big sub-field. How to solve the passive imaging issues using data-driven and FWI methods requires further study and development. We conduct a preliminary test on event picking for passive data, which can be found in Section 15 of the supplementary material, to serve as a kick-off experiment for future studies.

\subsection{Broader Impact}
\textbf{Data-driven FWI:} FWI is a typical scientific problem being studied with physics-driven approaches for decades, with the rapid development of deep learning, we have seen a myriad of data-driven approaches. \textsc{OpenFWI} embraces this junction and brings the community with the potential of: (1) \textit{Unified Evaluation}, (2) \textit{Further Improvement} and (3) \textit{Re-producibility and Integrity}, which are essential as the study on this topic evolves. We also envision \textsc{OpenFWI} supporting domain experts attempting to explore deep learning methods with a smooth beginning, and machine learning professionals pursuing further improvement on the current limitations. \par

\textbf{Future Developments:} We plan to maintain \textsc{OpenFWI} meticulously by releasing new datasets, and new benchmarks and serving the community with follow-up questions. There will be workshops with future updates about \textsc{OpenFWI}, and data competitions with more challenging data/tasks at the appropriate junction. We also appreciate any feedback from both the geophysics and machine learning communities on improving \textsc{OpenFWI}.\par

\textbf{AI for science:}~Scientific machine learning~(SciML) is demonstrating its great potential in various disciplines including geoscience. Compared to other fields in machine learning (such as computer vision and natural language processing), serious data challenges remain - sparse direct measurements, unbalanced data distribution, inevitable noise, etc.  Our effort would hopefully shed some light on how to overcome those data challenges for SciML to enable exciting progress in typical science-rich and data-starved scientific fields.

\section{Conclusion}
\label{sec:conclusion}
In this paper, we introduced \textsc{OpenFWI}, an open-source platform containing twelve datasets and benchmarks on four deep learning methods. The released datasets have various scales, encompass diverse domains in geophysics, and have simulated multiple scenarios of subsurface structures. The current benchmarks showed promising results on some datasets, while the rest may need further improvement. In addition, we also include complexity analysis, generalization study, and uncertainty quantification to demonstrate the favorable properties of our datasets and benchmarks. Last, we discussed existing challenges that can be studied with these datasets and conceived the future advancement as \textsc{OpenFWI} evolves.

\section*{Acknowledgement}

This work was funded by the Los Alamos National Laboratory~(LANL) - Laboratory Directed Research and Development program under project number 20210542MFR and by the U.S. Department of Energy~(DOE) Office of Fossil Energy’s Carbon Storage Research Program via the Science-Informed Machine Learning to Accelerate Real Time Decision Making for Carbon Storage~(SMART-CS) Initiative. The first author would like to thank Ms. Mier Chen for valuable inputs on UI/UX design of \textsc{OpenFWI}.

\bibliographystyle{unsrt}  
\bibliography{references}  
\clearpage

\appendix

Appendix arrangement:
\begin{itemize}
    \item \Cref{sup:license} highlights the public resources to support the reproducibility and describes the licenses of the \textsc{OpenFWI} data and released code.
    \item \Cref{sup:velocity} illustrates the generation pipeline of velocity maps.
    \item \Cref{sup:forward} lists the configurations of seismic forward modeling.
    \item \Cref{sup:format} introduces the format of dataset files, the practice of naming, and other details.
    \item \Cref{sup:network} shows the network architecture design and specifies the parameters involved.
    \item \Cref{sup:training} provides all parameters and configurations for training to guarantee reproducibility.
    \item \Cref{sup:inversion} contains more illustrations of predicted velocity maps for all datasets.
    \item \Cref{sup:complexity} introduces the subsurface complexity metrics with concrete numerical results and illustrations.
    \item \Cref{sup:gen} demonstrates more details on the generalization test and analysis of the results.
    \item \Cref{sup:uncertainty} conducts the case study of uncertainty quantification.
    \item \Cref{sup:robust} includes the robustness analysis and how to improve model robustness.
    \item \Cref{sup:method-compare} compares inversion results and computational cost of physics-driven methods and data-driven methods.
    \item \Cref{sup:inv2d_3d} studies if predicting 2D slices by InversionNet can have comparable results as InversionNet3D.
    \item \Cref{sup:field} elaborates the test strategy in the real-world situation.
   \item \Cref{sup:passive} establishes a connection from \textsc{OpenFWI} to passive geological inference and its potential applications.
    \item \Cref{sup:dis} has more discussion on previous versions of datasets and current limitations of \textsc{OpenFWI}.
\end{itemize}

\section{\textsc{OpenFWI} Public Resources and Licenses}
\label{sup:license}
First and foremost, the reproducibility of \textsc{OpenFWI} benchmarks is guaranteed by a number of public resources, listed below. Remarkably, we have a group (link available below) where any related discussion is welcome. Our team also promises to maintain the platform and support further developments based on the community feedback.

\begin{itemize}
\item \textbf{Website: } \url{https://openfwi-lanl.github.io}
\item \textbf{Dataset URL: } \url{ https://openfwi-lanl.github.io/docs/data.html#vel}
\item \textbf{Github Repository:} \url{https://github.com/lanl/openfwi}
\item \textbf{Pretrained Models:} \url{https://tinyurl.com/bddzkxfz}
\item \textbf{Tutorial: }\url{ https://openfwi-lanl.github.io/tutorial/}
\item \textbf{Google Group: } \url{https://groups.google.com/g/openfwi}
\end{itemize}

The codes are released on Github under \textbf{OSS} license and \textbf{BSD-3} license, as required by the Los Alamos National Lab and the Department of Energy, U.S.A. We also attach the Creative Commons Attribution-NonCommercial-ShareAlike 4.0 International License to the data.

\section{Velocity Map Generation}
\label{sup:velocity}
\begin{figure*}[h]
\centering
\includegraphics[width=1.0\columnwidth]{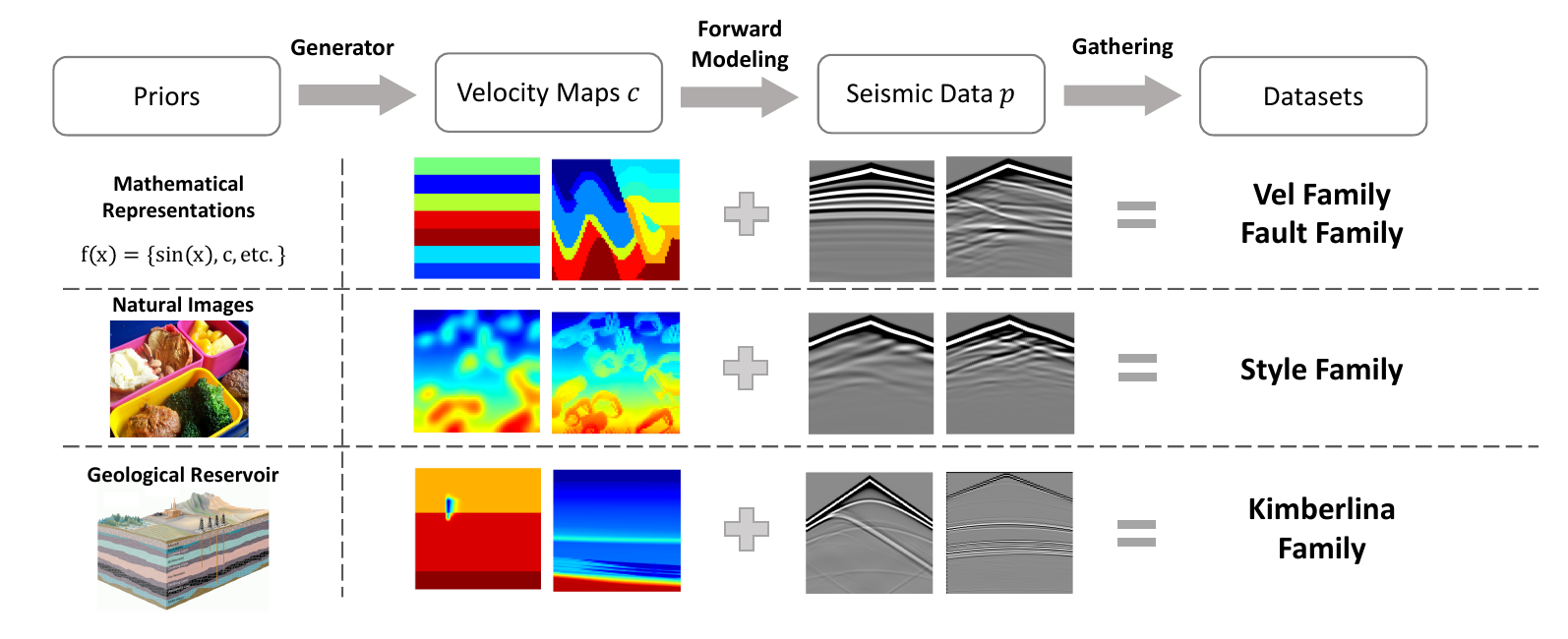}
\caption{\textbf{Data generation pipelines.} The velocity maps $c$ are built from three different priors, then the seismic data $p$ are generated via forward modeling. Velocity maps $c$ and seismic data $p$ are collected to construct the four families of datasets. }
\label{fig:generation}
\end{figure*}

In this section, we introduce the data generation pipelines. Essentially, the data generation follows two steps: (1) synthesizing velocity maps $c$ and (2) generating seismic data $p$ via forward modeling. In the first step, we generate the velocity maps from three different prior information: mathematical representations, natural images, and geological reservoir, which notably contributes to the dataset diversity. In the second step, the seismic data is obtained from the synthetic velocity maps via forward modeling. \Cref{fig:generation} illustrates the data generation process with three priors elaborated below.

\textbf{Mathematical Representations:} The sediments are usually deposited as flat-lying layers so that the velocity map $c_0$ is initialized as a combination of multiple flattened layers with random width. To mimic the curved folds and faults caused by the geological activities, ``\textit{Vel Family}'' and ``\textit{Fault Family}'' are generated by recursively applying the following equations to $c_0$, respectively:
    \begin{align}
        &c_i(x,y)=c_{i-1}(x,a_i\sin(2\pi k_i x));~i>0,\label{eq:vel_gen}\\
        &c_i(x,y)=\left\{
        \begin{array}{lr}
            c_{0}(x+s_i,a_i\sin(2\pi k_i x)+s'_i),&y\geq f_i(x) \\
            c_{i-1}(x,y),&y<f_i(x)
        \end{array}
        \right.;~i>0,\label{eq:fault_gen}
    \end{align}
    where $s_i$, $s'_i$, $k_i$ and $a_i$ are random variables in $i$-th iteration. $f_i(\cdot)$ is a random linear function for fault simulation. Pixels that are out of map's boundaries are filled by the nearest available values. 
   
\textbf{Natural Images:} ``\textit{Style Family}'' is obtained with a style transfer network $R(\cdot)$  trained with the Marmousi model~\cite{brougois1990marmousi} as the style image and the natural images $n$ from COCO dataset~\cite{lin2014microsoft} as the content images\footnote{For the training, we use the Github package at https://github.com/kewellcjj/pytorch-multiple-style-transfer}. The details of the network can be referred to~\cite{feng2021multiscale}. The velocity maps are constructed using
\begin{equation}
  c=(1-\alpha)R(n)+{\alpha}c_{b},
\end{equation}
where $c_{b}$ is an 1D background velocity map with linearly increasing velocity value, and $\alpha$ is the weight between the style-transferred velocity map $R(n)$ and $c_{b}$. 

\textbf{Geological Reservoir:} ``\textit{Kimberlina Family}'' is built under DOE’s National Risk Assessment Program (NRAP) based on a potential CO$_2$ reservoir at the Kimberlina site in the southern San Joaquin Basin, CA, USA~\cite{Wagoner2009}. The hydrologic-state models $h$ are produced with reservoir-simulation scenario~\cite{Development-2021-Alumbaugh} and then converted into velocity maps $c$ following the petrophysical transformation methods $P(\cdot)$~\cite{yang2019assessment}:
\begin{equation}
  c=P(h).
\end{equation}

\section{Seismic Forward Modeling Details}
\label{sup:forward}
We follow the forward modeling algorithm at~\url{https://csim.kaust.edu.sa/files/SeismicInversion/Chapter.FD/lab.FD2.8/lab.html}.
The seismic data is simulated using finite difference methods~\cite{moczo2007finite} with the absorbing boundary condition~\cite{engquist1977absorbing} and the Ricker wavelet is the source function. The original code in the link is written in MATLAB. To increase its computational efficiency and its compatibility with the neural network, we change its scheme to 2-4 (2nd-order accuracy in time and 4th-order in space) and rewrite it in Python. Sources and receivers are evenly distributed on the surface. The details of the forward modeling configuration are listed in~\Cref{tab:forward}, including the grid spacing of the velocity maps, the central frequency of the source wavelet, and so on. Below is an example of the acquisition geometry setting in the MATLAB script for the 1st source in  ``Vel, Fault and Style'' datasets:

 \begin{verbatim}
nz=70; nx=70;
dx=10; nbc=120; nt=1001; dt=0.001;
freq=15; s=ricker(freq,dt); isFS=false; 
coord.sx = 1*dx; coord.sz = 1*dx;
coord.gx=(1:nx)*dx; coord.gz=ones(size(coord.gx))*dx;
\end{verbatim}

\begin{table}
\centering
\caption{\textbf{Seismic Forward Modeling Configuration}}
\vspace{0.5em}
\renewcommand{\arraystretch}{1.5}
\label{tab:forward}
\begin{adjustbox}{width=1\textwidth}
\begin{tabular}{c|c|c|c|c|c|c|c|c|c} 
    \specialrule{.15em}{.05em}{.05em} 
{Dataset}                        & \begin{tabular}[c]{@{}c@{}}{Grid }\\{Spacing}\end{tabular} & \begin{tabular}[c]{@{}c@{}}{Source }\\{Frequency}\end{tabular} & \begin{tabular}[c]{@{}c@{}}{Source}\\{~Spacing}\end{tabular} & \begin{tabular}[c]{@{}c@{}}{Source }\\{Numbers}\end{tabular} & \begin{tabular}[c]{@{}c@{}}{{Receiver }}\\{{Spacing}}\end{tabular} & \begin{tabular}[c]{@{}c@{}}{Receiver }\\{Numbers}\end{tabular} & \begin{tabular}[c]{@{}c@{}}{Time}\\{~Spacing}\end{tabular} & \begin{tabular}[c]{@{}l@{}}{Time }\\{Steps}\end{tabular} &\begin{tabular}[c]{@{}c@{}}{Boundary }\\{Grids}\end{tabular} \\ 
    \specialrule{.15em}{.05em}{.05em} 

``Vel, Fault and Style'' Family & 10 $m$                                                                      & 15 $Hz$                                                                   & 175 $m$                                                                    & 5                                                                          & 10 $m$                                                                                         & 70                                                                           & 0.001 $s$                                                                & 1,001  &120                                                                  \\
Kimberlina-CO$_2$                          & 10 $m$                                                                      & 10 $Hz$                                                                   & 400 $m$                                                                    & 9                                                                          & 40 $m$                                                                                         & 101                                                                          & 0.002~$s$                                                                & 1,251   &120                                                                 \\
3D Kimberlina-V1                        & 10~$m$                                                                      & 15 $Hz$                                                                   & 800~$m$                                                                    & 5$\times$5                                                                     & 100 $m$                                                                                        & 40$\times$40                                                                     & 0.001~$s$                                                                & 5,001   &100                                                                 \\
    \specialrule{.15em}{.05em}{.05em} 
\end{tabular}
\end{adjustbox}
\end{table}

\section{\textsc{OpenFWI} Datasets: Illustration, Format, Naming, Loading}
\label{sup:format}
This section contains instructions on the usage of all datasets. We emphasize that all velocity maps and seismic data are saved as ``.npy'' files, therefore, could be conveniently accessed through Python. Other details may vary as datasets have different sizes, functions, and generation backgrounds. For the rest of this section, we go through each dataset carefully. 

\subsection{Vel Family}

\begin{figure*}[h!]
\centering
\includegraphics[width=1\columnwidth]{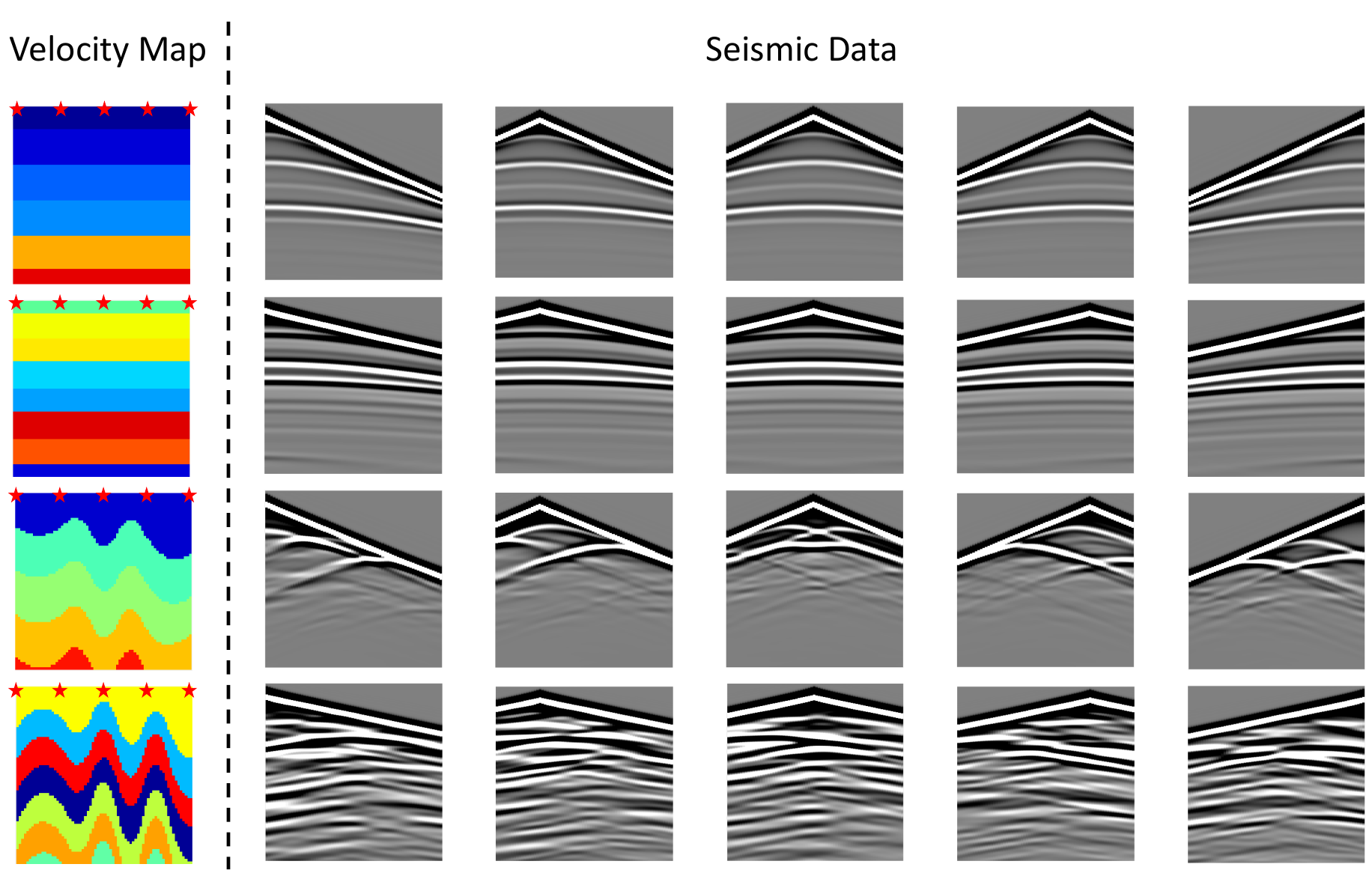}
\caption{\textbf{Example of velocity maps and seismic data in ``\textit{Vel Family}''}. The left part shows one velocity map from each dataset, and the right demonstrates all five channels of the seismic data. The red stars in the velocity map indicate the location of sources while the receivers are distributed all over the surface.}
\label{fig:vel_family}
\end{figure*}

The seismic data and velocity maps are saved in two folders,\texttt{\{./data\}} and \texttt{\{./model\}}, respectively. The naming of files follows the format of \texttt{\{data\}\{n\}} for seismic data and \texttt{\{model\}\{n\}} for velocity maps, \texttt{n} denotes the index of a file (starting from 1). Notice that for the same \texttt{n}, data and model become a pair. Each file contains $500$ samples. The training set and testing set are split as 24K/6K, the corresponding training data are \texttt{\{./data1-48.npy\}} (paired with \texttt{\{./model1-48.npy\}}), 
and the rest are testing data. Examples are shown in ~\Cref{fig:vel_family}.

\subsection{Fault Family}
\begin{figure*}[h!]
\centering
\includegraphics[width=1\columnwidth]{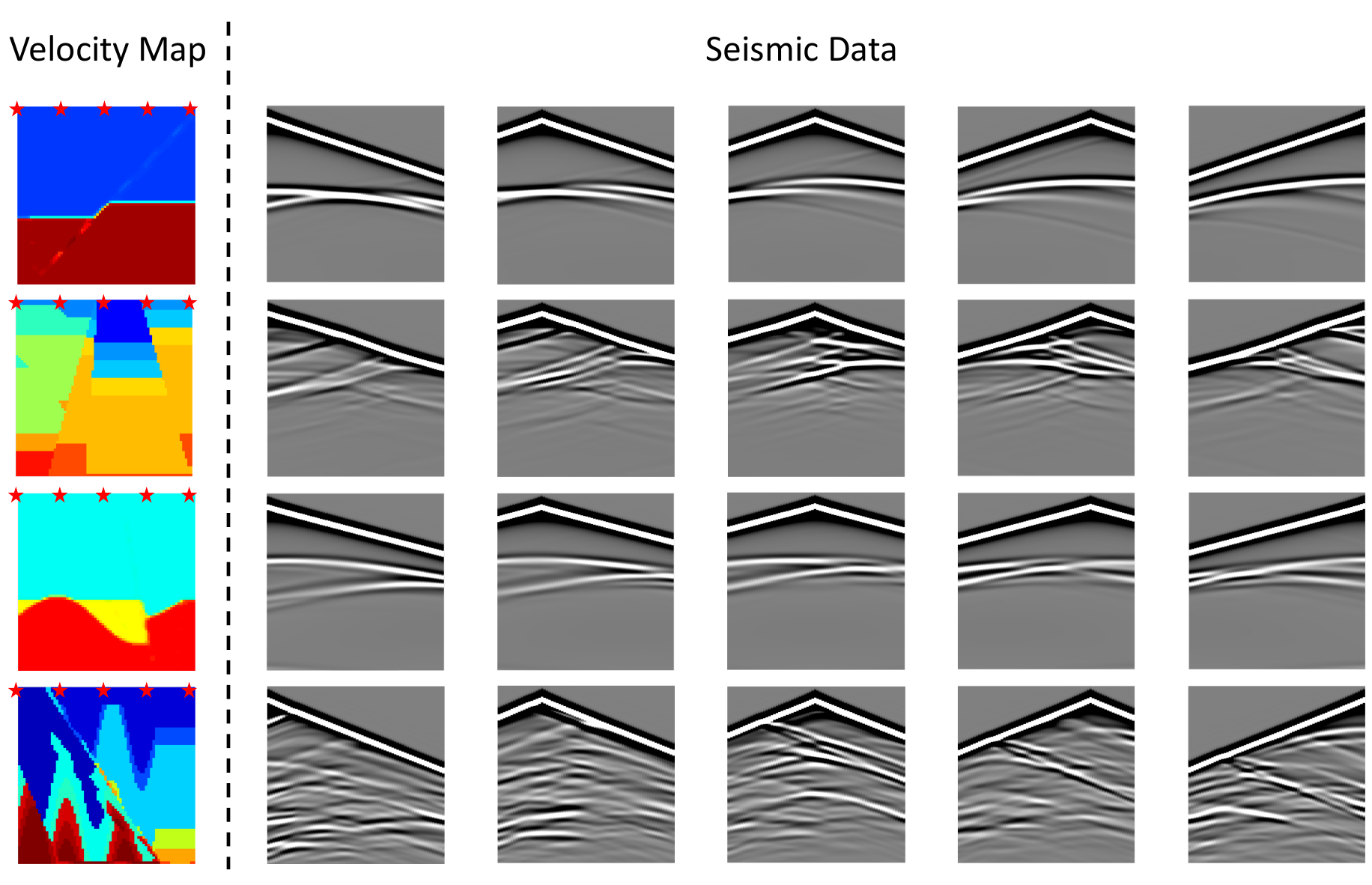}
\caption{\textbf{Example of velocity maps and seismic data in ``\textit{Fault Family}''}. The left part shows one velocity map from each dataset, and the right demonstrates all five channels of the seismic data. The red stars in the velocity map indicate the location of sources while the receivers are distributed all over the surface.}
\label{fig:fault_family}
\end{figure*}

The naming of files can be described as \texttt{\{vel|seis\}\_\{n\}\_1\_\{i\}.npy}, where \texttt{vel} and \texttt{seis} specify if a file includes velocity maps or seismic data, \texttt{n} represents the number of initial flatten layers for velocity maps generation and \texttt{i} is the index of a file (start from 0) among the ones with the same \texttt{n}. The sizes of training and testing datasets are 48K/6K. Examples are shown in~\Cref{fig:fault_family}.

\subsection{Style Family}
\begin{figure*}[h!]
\centering
\includegraphics[width=1\columnwidth]{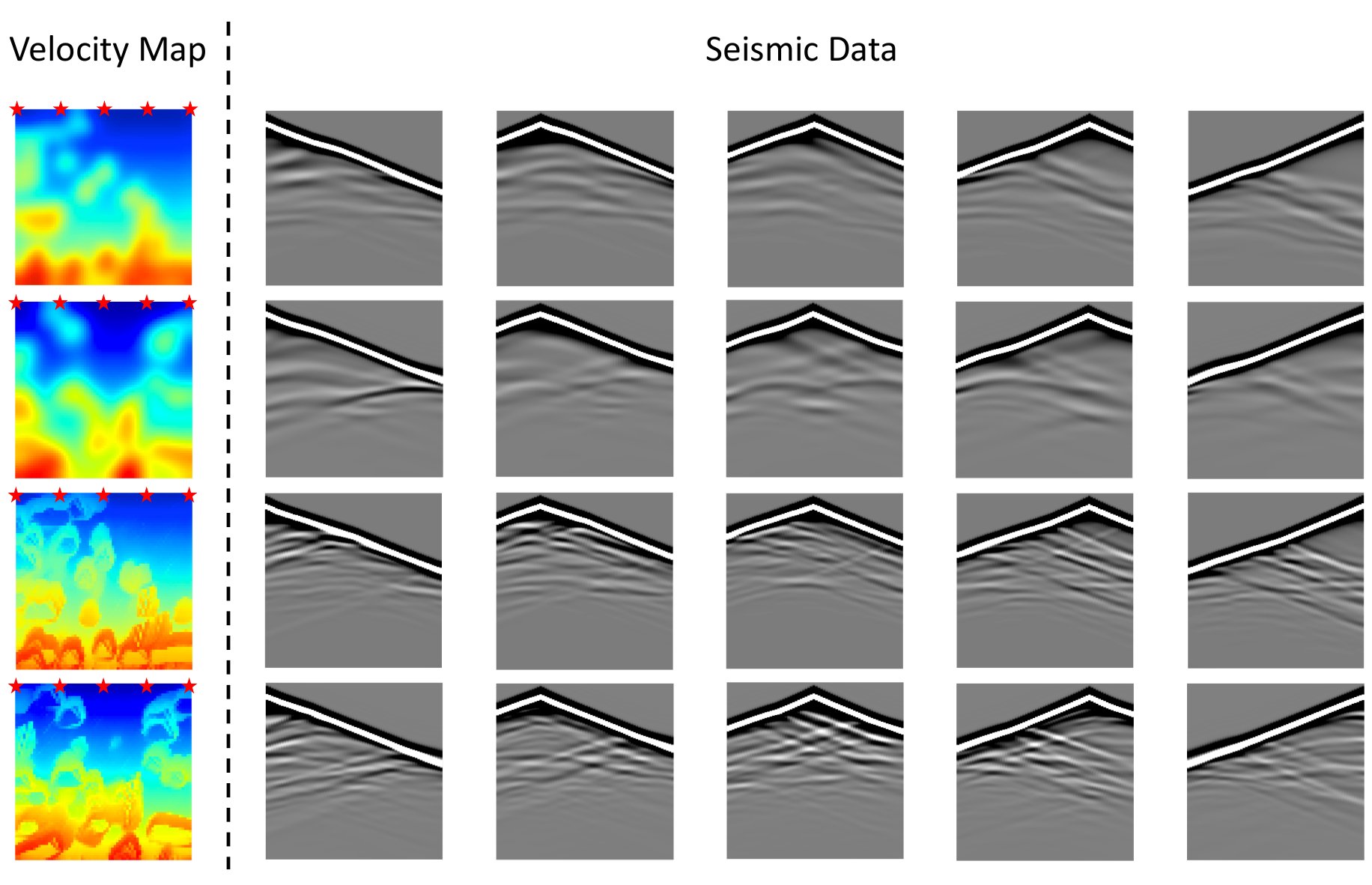}
\caption{\textbf{Example of velocity maps and seismic data in ``\textit{Style Family}''}. The left shows the velocity maps, the first two rows are sampled from Style-A and the last two from Style-B. The right demonstrates all five channels of the seismic data. The red stars in the velocity map indicate the location of sources while the receivers are distributed all over the surface.}
\label{fig:style_family}
\end{figure*}

The saving and naming of ``\textit{Style Family}'' is same with ``\textit{Vel Family}'', seismic data \texttt{\{data\}\{n\}} are saved in \texttt{\{./data\}} while \texttt{\{model\}\{n\}} are saved in \texttt{\{./model\}}, \texttt{n} denotes the index of a file (starting from 1). Each file contains $500$ samples. The training set and testing set are split as 60K/7K, the corresponding training data are \texttt{\{./data1-120.npy\}} (paired with \texttt{\{./model1-120.npy\}}), and testing data are contained in \texttt{\{./data121-134.npy\}} (paired with \texttt{\{./model121-134.npy\}}). Examples are shown in~\Cref{fig:style_family}.
\vspace{0.5em}

\subsection{Kimberlina Family}
\begin{figure*}[h!]
\centering
\includegraphics[width=1\columnwidth]{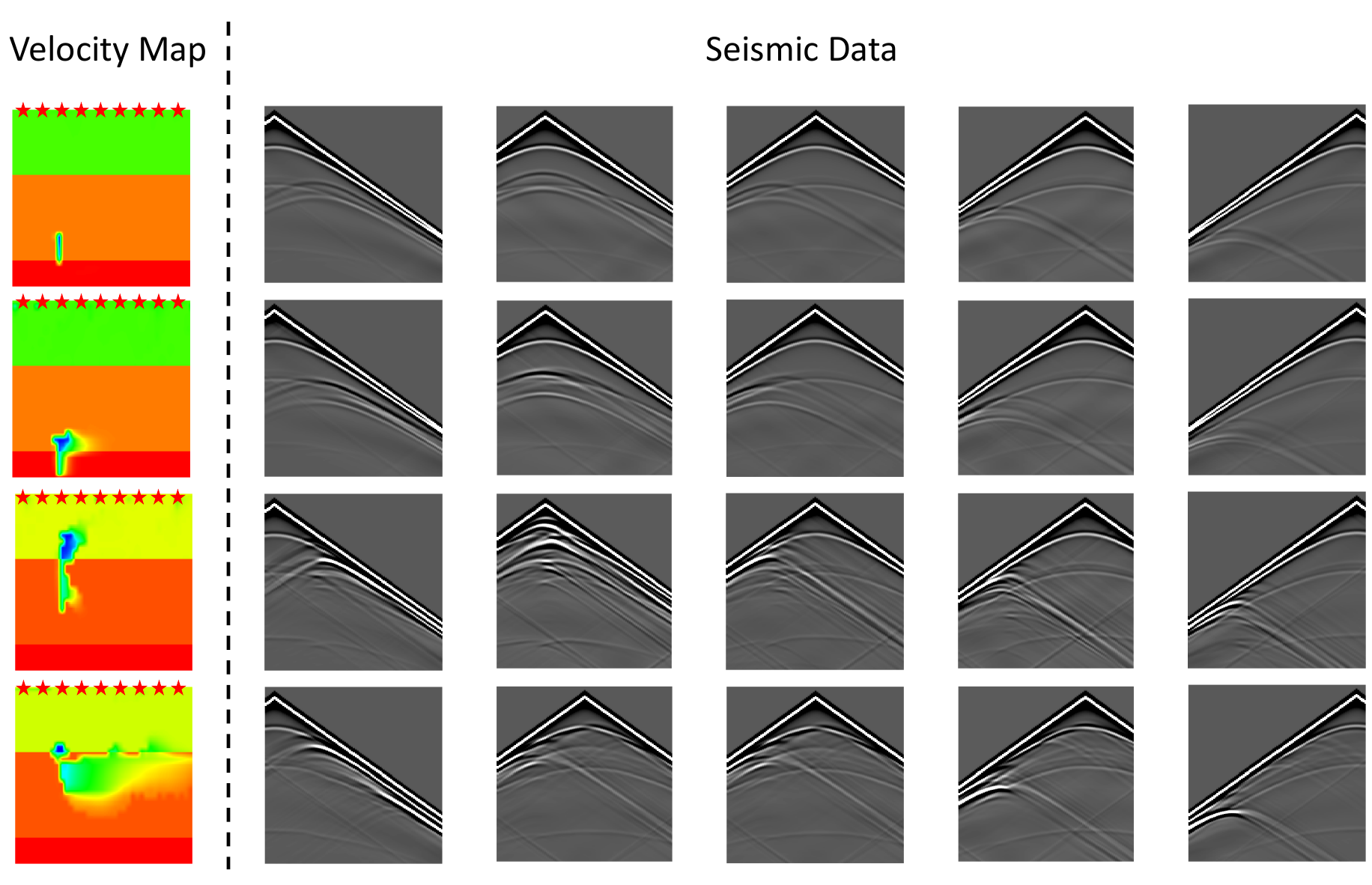}
\caption{\textbf{Example of velocity maps and seismic data in Kimberlina-CO$_2$ dataset}. The left shows the velocity maps with leakage in different levels (First two rows: small; Third row: medium; Last row: Large). The right demonstrates five channels among all nine channels of the seismic data. The red stars in the velocity map indicate the location of sources while the receivers are distributed all over the surface.}
\label{fig:kim_co2}
\end{figure*}

The seismic data and velocity maps of Kimberlina-CO$_2$ dataset are saved in four folders,\texttt{\{./kimberlina\_co2\_\emph{\{phase\}}\_data\}}, \texttt{\{./kimberlina\_co2\_\emph{\{phase\}}\_label\}}, where \emph{\{phase\}} denotes ``train'' or ``test''. 
We use a naming convention of \texttt{\{data\}\_sim\{n\}\_t\{m\}} for seismic data and \texttt{\{label\}\_sim\{n\}\_t\{m\}} for velocity maps, \texttt{n} denotes the 4-digit index of a simulation (starting from 0), and \texttt{m} represents the timesteps from 10 to 200 (at every 10 years). Each file contains one sample, and for the same \texttt{n} and \texttt{m}, data and model become a pair. Examples of Kimberlina-CO$_2$ are shown in \Cref{fig:kim_co2}.

\begin{figure*}[h!]
\centering
\includegraphics[width=1\columnwidth]{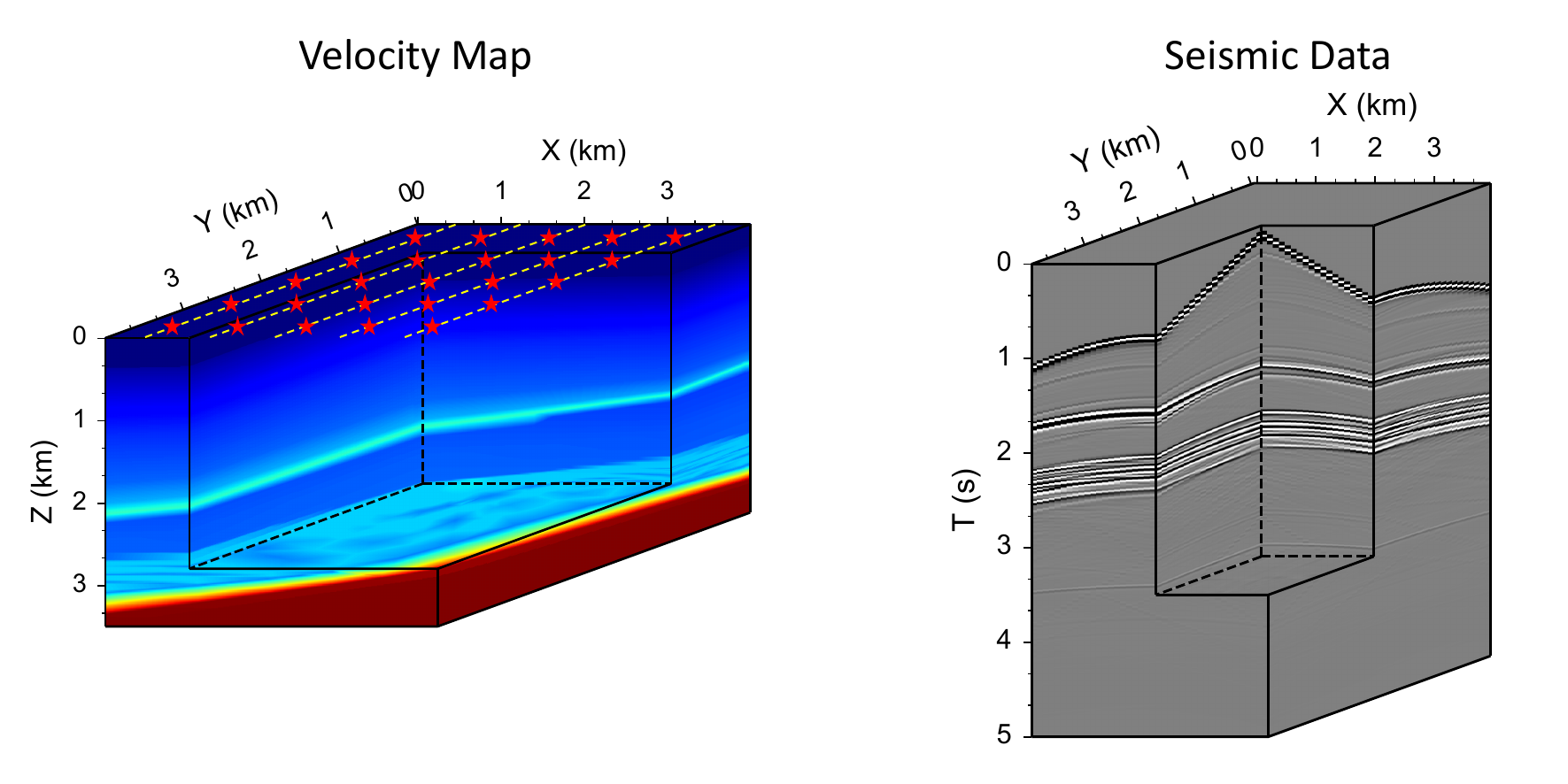}
\caption{\textbf{Example of velocity map and seismic data in 3D Kimberlina-V1 dataset}. The left is the velocity map and the right is seismic data with the source located in the center of the surface. The red stars in the velocity map indicate the location of sources and the yellow dash lines are the survey lines. The receivers are distributed every 100 $m$ over the surface.}
\label{fig:kim_3d}
\end{figure*}

In 3D Kimberlina-V1\footnote{3D Kimberlina-V1 will be released at \url{https://edx.netl.doe.gov/} upon approval by Los Alamos National Laboratory and U.S. Department of Energy.}. The velocity maps are stored in \texttt{\{./velocity\_model.tar.gz\}} as \texttt{\{year\{m\}\_cut\{n\}.npy\}}.~\texttt{m} represents the injection year and \texttt{n} denotes the index of a file (starting from 1). Since the seismic data is too large, they are split into 12 files \texttt{\{seismic\_data.tar.gz.partaa$\sim$al\}}. Examples of 3D Kimberlina-V1 are shown in \Cref{fig:kim_3d}.

\section{\textsc{OpenFWI} Benchmarks: Network Architecture}
\label{sup:network}
The \textsc{OpenFWI} benchmarks are established upon four deep learning methods: InversionNet~\cite{wu2019inversionnet}, VelocityGAN~\cite{zhang2020data}, and UPFWI~\cite{Jin-2021-Unsupervised} for 2D and InversionNet3D~\cite{zeng2021inversionnet3d} for 3D FWI. All details of these methods can be retrieved from their original papers, while in this section we describe the network architecture adopted particularly for the \textsc{OpenFWI} datasets. Note that the ten datasets in the ``\textit{Vel}'', ``\textit{Fault}'' and ``\textit{Style}'' family have the same size, thus share identical network architectures. The Kimberlina-CO$_2$ data requires a minor change on the convolution kernel parameters.

\subsection{InversionNet}
InversionNet is an encoder-decoder structural CNN network. The encoder extracts the hyper-features of the seismic input, and the decoder estimates the corresponding velocity map from the compressed latent vector. We stack $14$ CNN layers in the encoder where the first layer has a $7*1$ kernel size, and the following six layers have a $3*1$ kernel size. Stride $2$ is applied every the other layer to reduce the data dimension to the velocity map dimension. Then six $3*3$ CNN layers are used to extract spatial-temporal features in the compressed data, in which the data is down-sampled every the other layer using stride $2$. After that, a CNN layer with an $8*9$ kernel size is stacked to flatten the feature maps to the output latent vector size, which is $512$ in our implementation. The decoder first applies a deconvolutional layer on the latent vector to generate a $5*5*512$ tensor with a kernel size $5$, followed by a convolutional layer with the same number of input and output channels. The deconvolution-convolution process is then duplicated for $4$ times with a kernel size of $4$ in the deconvolutional layers, resulting in a feature map with a size of $80*80*32$. Finally, we center-crop the feature map by a $70*70$ window and apply a $3*3$ convolution layer to output a single channel velocity map. Thus, there are $14$ CNN layers in the encoder and $11$ layers in the decoder. All the aforementioned convolutional and deconvolutional layers are followed by batch normalization and leakyReLU as the activation function.

\subsection{VelocityGAN}
VelocityGAN is a generative adversarial network, its generator has the same architecture as the encoder-decoder network in InversionNet. The discriminator is a $9$-layer CNN. First, eight $3*3$ CNN layers are used to extract spatial-temporal features in the velocity maps, in which the data is down-sampled every the other layer using stride $2$. Then, a CNN layer with a $5*5$ kernel size and zero padding is used as the output layer. Similarly, all the aforementioned convolutional layers are followed by batch normalization and leakyReLU as the activation function. To train the GAN, we use Wasserstein loss with a gradient penalty besides the pixel-wise $\ell_1$-norm and $\ell_2$-norm losses to distinguish the real and generated velocity maps.

\subsection{UPFWI}
UPFWI is an unsupervised learning method that utilizes convolutional layers in an encoder-decoder architecture. The general network architecture is the same as the one in InversionNet, except for the decoder uses CNN layers with nearest neighbor upsampling as the deconvolutional layers. \par

Unlike the other benchmark methods, UPFWI minimizes the data difference by the following loss function:
\begin{equation}
\ell(p,\tilde{p}) = \lambda_1 \ell_{1}(p,\tilde{p}) + \lambda_2 \ell_{2}(p,\tilde{p}) + \lambda_3 \ell_{per1}(\phi(p),\phi(\tilde{p})) + \lambda_4 \ell_{per2}(\phi(p),\phi(\tilde{p})),
\label{upfwi_loss}
\end{equation}
where $p$ and $\tilde{p}$ are the input and reconstructed seismic data from the forward operator as in section~\ref{sup:forward}. $\ell_1$ and $\ell_2$ are the $\ell_1$-norm and $\ell_2$-norm losses measuring the pixel-wise data losses, respectively. And the $\ell_{per1}$ and $\ell_{per2}$ are the $\ell_1$- and $\ell_2$-norm distance of the perceptual losses that we extract data features from $conv3$ in a VGG-16 network pre-trained on ImageNet. $\phi(\cdot)$ represents the output features of the VGG-16 network. $\lambda_i$ are the hyper-parameters that balance the four parts of the total loss.

\subsection{InversionNet3D}
As a natural extension of InversionNet in the 3D domain, InversionNet3D was constructed in a similar topological structure. The shallowest version, from which the results in this paper were obtained, was built upon 13 layers in both encoder and decoder with each layer being a 3D convolution or deconvolution followed by batch normalization and LeakyReLU activation. In order to reduce memory consumption and computational complexity, two of the most important barriers of 3D FWI, the filter size and stride of each layer were deliberately chosen and another two special components, group convolution, and invertible layers were also employed in certain stages of the network. As a result, this baseline network has $14.42$ million parameters and consumes $9.93$GB of memory with a batch size of one. It is recommended to refer to the original paper for a more accurate and detailed description of the network architecture.

\section{\textsc{OpenFWI} Benchmarks: Training Configurations}
\label{sup:training}
In this section, we provide the details of training configurations to guarantee reproducibility.  All the experiments are implemented on NVIDIA Tesla P100 GPUs. We have released the pretrained model via Google drive: \url{https://tinyurl.com/bddzkxfz}. The codes and related information are available on Github: \url{https://github.com/lanl/OpenFWI}. 

In 2D benchmarks, we use identical hyperparameters across all datasets for InversionNet and VelocityGAN, while the training of UPFWI varies a little on different datasets. In particular, we employ AdamW~\cite{loshchilov2018decoupled} optimizer with a weight decay of $1\times10^{-4}$ and momentum parameters $\beta_1=0.9$, $\beta_2=0.999$ to update all models. 
For InversionNet, the learning rate is $1\times10^{-4}$, and no decay is applied. The size of a mini-batch is 256. We train all InversionNet models for 120 epochs. 
For VelocityGAN, the learning rate of both generator and discriminator is $1\times10^{-4}$, and no decay is applied. We set the size of a mini-batch to $64$. Following the strategy of \cite{arjovsky2017wasserstein}, we perform three discriminator iterations per generator update. All VelocityGAN models are trained for 480 epochs. 
For UPFWI, the initial learning rate is $3.2\times10^{-4}$, and we reduce the learning rate by a factor of $10$ at epochs $150$ and $175$, except for the experiments on CurveVel-A and CurveVel-B where no decay is applied. The size of a mini-batch is $128$ in the experiments on CurveVel-A and CurveVel-B and $256$ in all the other experiments. We follow \cite{Jin-2021-Unsupervised} and set the weight of each loss term to $1$. Due to relatively high computational cost, we train UPFWI models for as many epochs as we can. The number of epochs in different experiments ranges from $200$ to $500$. 
In 3D benchmarks, we keep the same training configurations of InversionNet3D in all experiments, and details can be found in \cite{zeng2021inversionnet3d}. 

During training, we apply min-max normalization to rescale velocity maps and seismic data to $[-1, 1]$. The velocity values range from $1500$ $m/s$ to $4500$ $m/s$, and the amplitude of seismic data is between $-20$ and $60$ in ``\textit{Vel Family}'', ``\textit{Fault Family}'' and ``\textit{Style Family}''. For ``\textit{Kim Family}'', velocity values range from $947$ $m/s$ to $2545$ $m/s$ in Kimberlina-CO$_2$ and from $1975$ $m/s$ to $3892$ $m/s$ in 3D Kimberlina-V1.

\section{Illustration of Inversion Results}
\label{sup:inversion}
\begin{figure*}[h!]
\centering
\includegraphics[width=1\columnwidth]{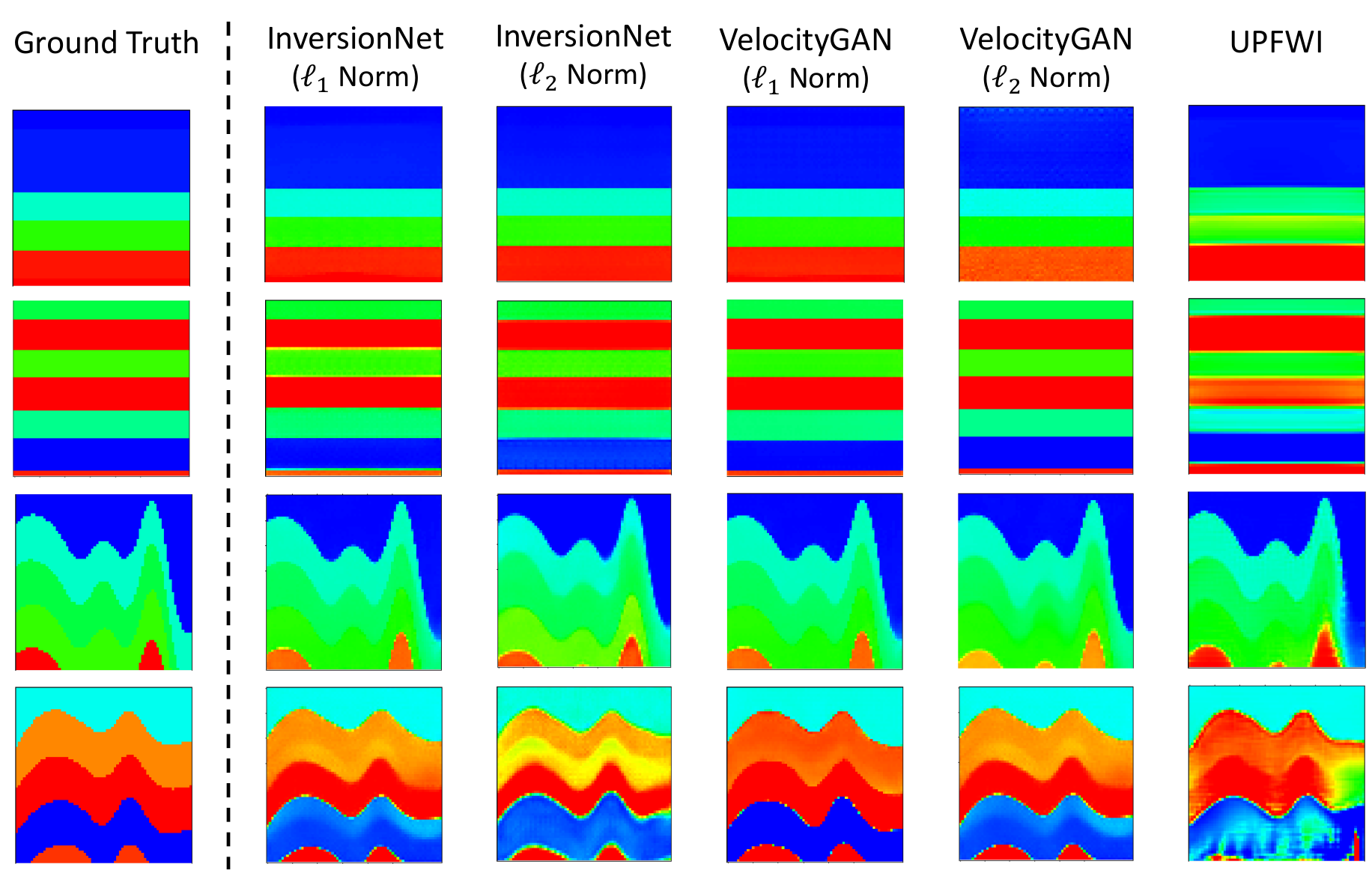}
\caption{\textbf{Example of ground truth and inversion results in ``\textit{Vel Family}''}. The left part shows the ground truth velocity map, and the right demonstrates the inversion results with InversionNet, VelocityGAN, and UPFWI. From the first row to the last: FlatVel-A, FlatVel-B, CurveVel-A, CurveVel-B.}
\label{fig:vel_result}
\end{figure*}

\begin{figure*}[h!]
\centering
\includegraphics[width=1\columnwidth]{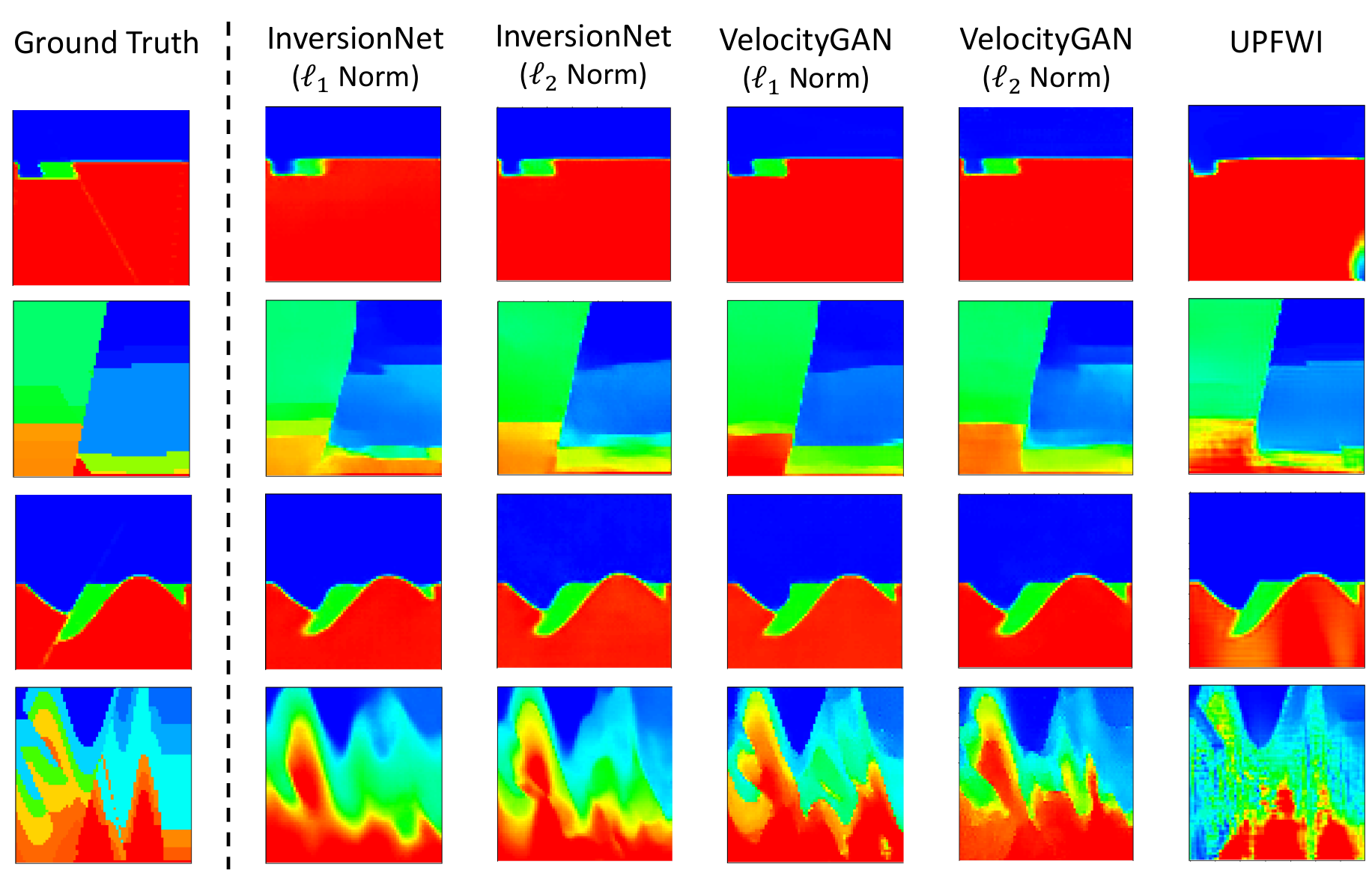}
\caption{\textbf{Example of ground truth and inversion results in ``\textit{Fault Family}''}. The left part shows the ground truth velocity map, and the right demonstrates the inversion results with InversionNet, VelocityGAN, and UPFWI. From the first row to the last: FlatFault-A, FlatFault-B, CurveFault-A, CurveFault-B.}
\label{fig:fault_result}
\end{figure*}

\begin{figure*}[h!]
\centering
\includegraphics[width=1\columnwidth]{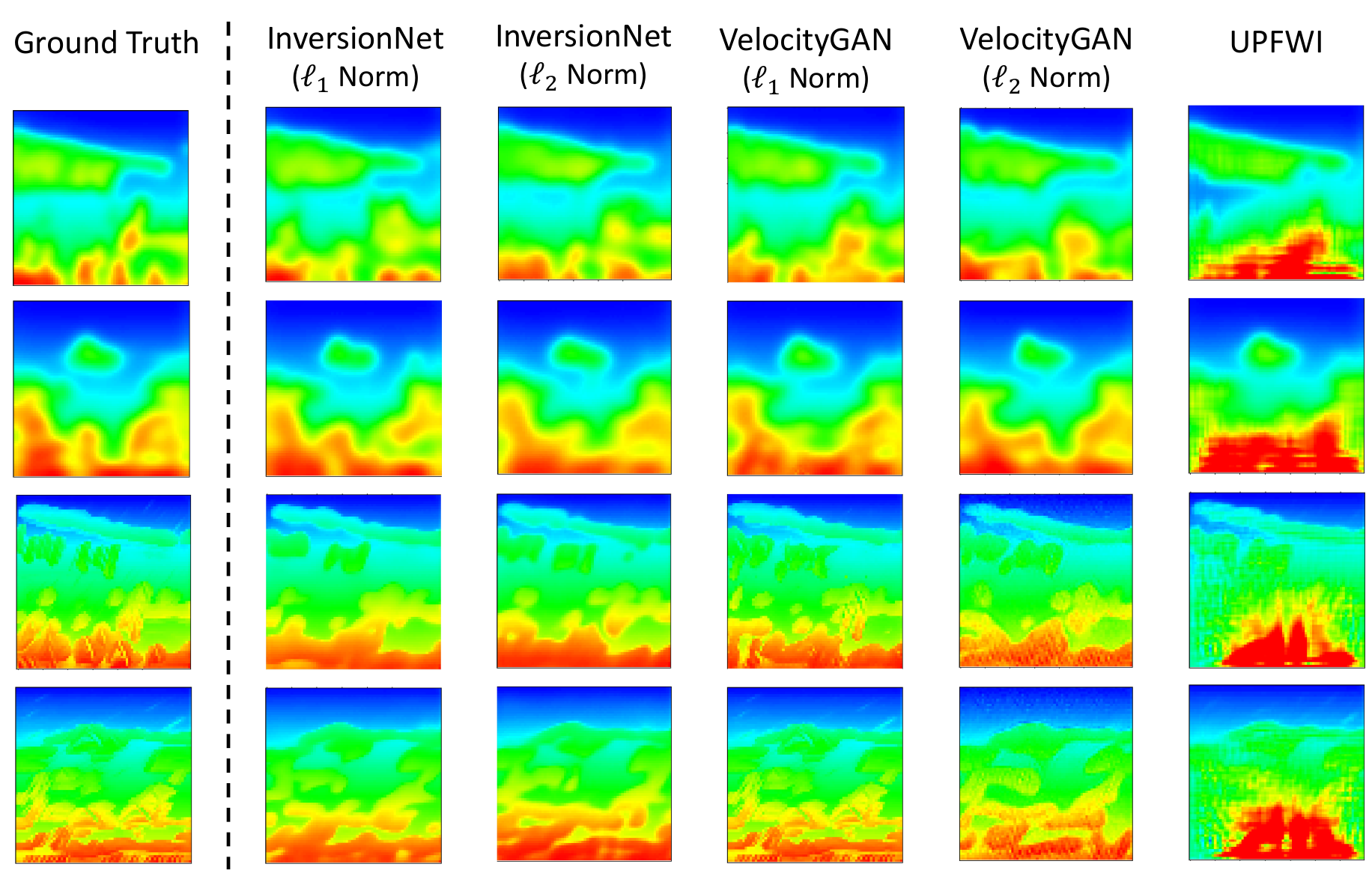}
\caption{\textbf{Example of ground truth and inversion results in ``\textit{Style Family}''}. The left part shows the ground truth velocity map, and the right demonstrates the inversion results with InversionNet, VelocityGAN, and UPFWI. First two rows: Style-A, last two rows: Style-B.}
\label{fig:style_result}
\end{figure*}

\begin{figure*}[h!]
\centering
\includegraphics[width=1\columnwidth]{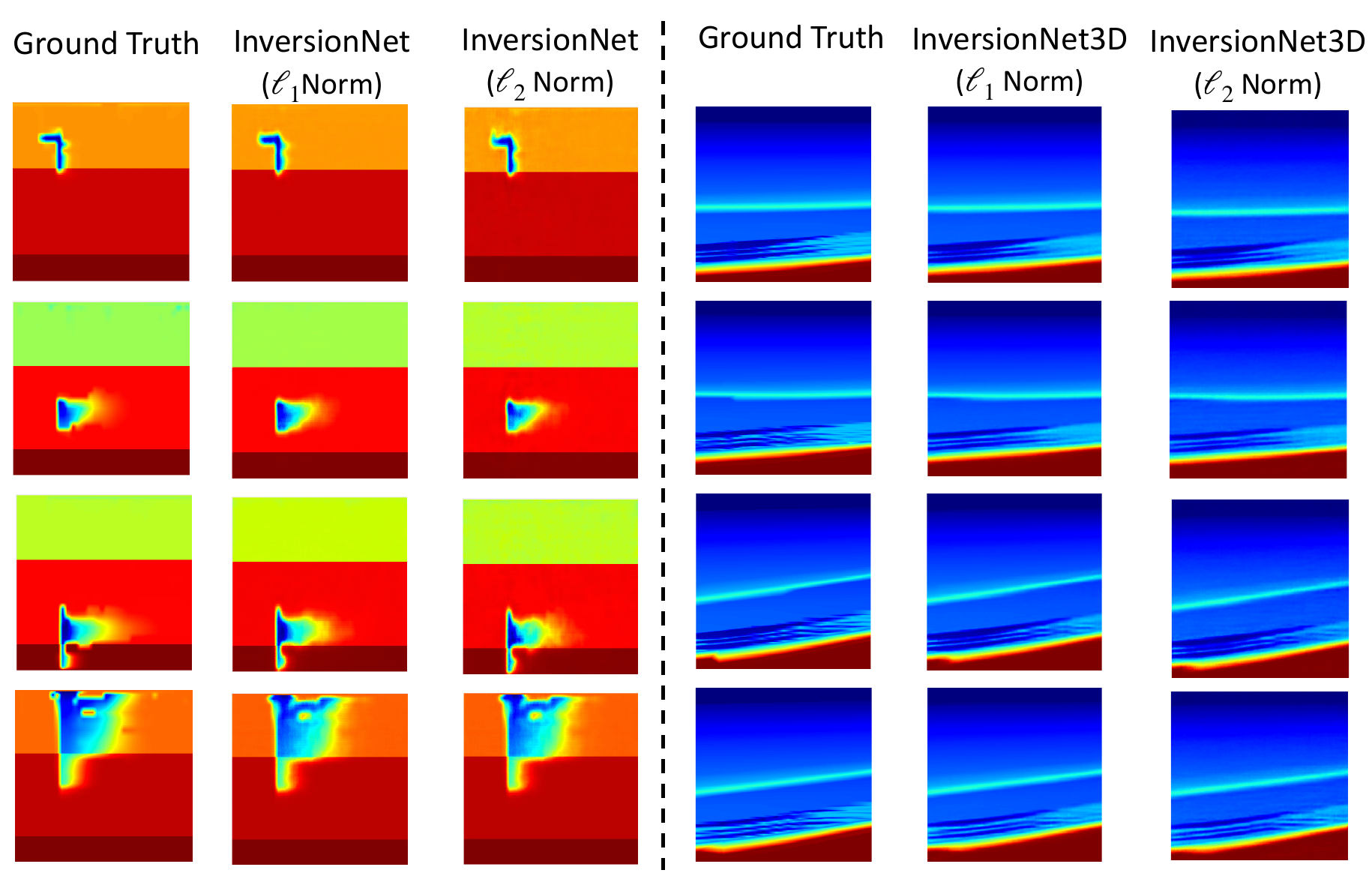}
\caption{\textbf{Example of ground truth and inversion results in ``\textit{Kim Family}''}. The left part shows the ground truth and the InversionNet inversion results with the Kimberlina-CO$_2$ dataset. The right part shows the 2D slices of the ground truth and the InversionNet3D inversion results with the 3D Kimberlina-V1 dataset. }
\label{fig:kim_result}
\end{figure*}

The benchmarks adopt numerical evaluations of the inversion results by SSIM. Here we present illustrations across different methods on all datasets in~\Cref{fig:vel_result,fig:fault_result,fig:style_result,fig:kim_result}. We notice that VelocityGAN usually provides the best results, but it requires more time for training. UPFWI uses the difference between predicted and observed seismic data as the loss function so that it is sensitive to the boundaries of the layers, which are the causes of the reflection waves in the data. Moreover, the deep regions of the velocity maps are hard to invert in UPFWI due to the limited seismic illuminations~\cite{schuster2017seismic}. 

\section{Velocity Map Complexity}
\label{sup:complexity}
In this section, we introduce three metrics applied to measure the velocity map complexity: spatial information~\cite{yu2013image}, gradient sparsity index~\cite{li2016sparse}, and Shannon entropy~\cite{lin1991divergence}. Shannon entropy is the most widely-applied quantifier of ``the amount of information'', and by definition, it is the expectation of the logarithm of the variable in the dataset. Spatial information (SI) is an estimator of the border magnitude since it is obtained from the Sober operator~\cite{kanopoulos1988design}, which is an edge detection filter. The gradient sparsity index (GSI) calculates the percentage of non-smooth pixels in an image also by applying the Sober operator. Specifically, let $G_x$ and $G_y$ denote the gradient magnitude on horizontal and vertical coordinates $(x,y)$ obtained via the Sober filter, $G$ denotes the matrix with $G_p$ as the element on pixel $p$, and $P$ denotes the total number of pixels, and then we have the following definitions:
\begin{eqnarray}
    &SI_{mean} = \mathbb{E}\sqrt{(G_{x}^2 + G_{y}^2)},\\
    &GSI = \frac{\|vec(G)\|_0}{P}.
\end{eqnarray}

\begin{figure*}[h!]
\centering
\includegraphics[width=1\columnwidth]{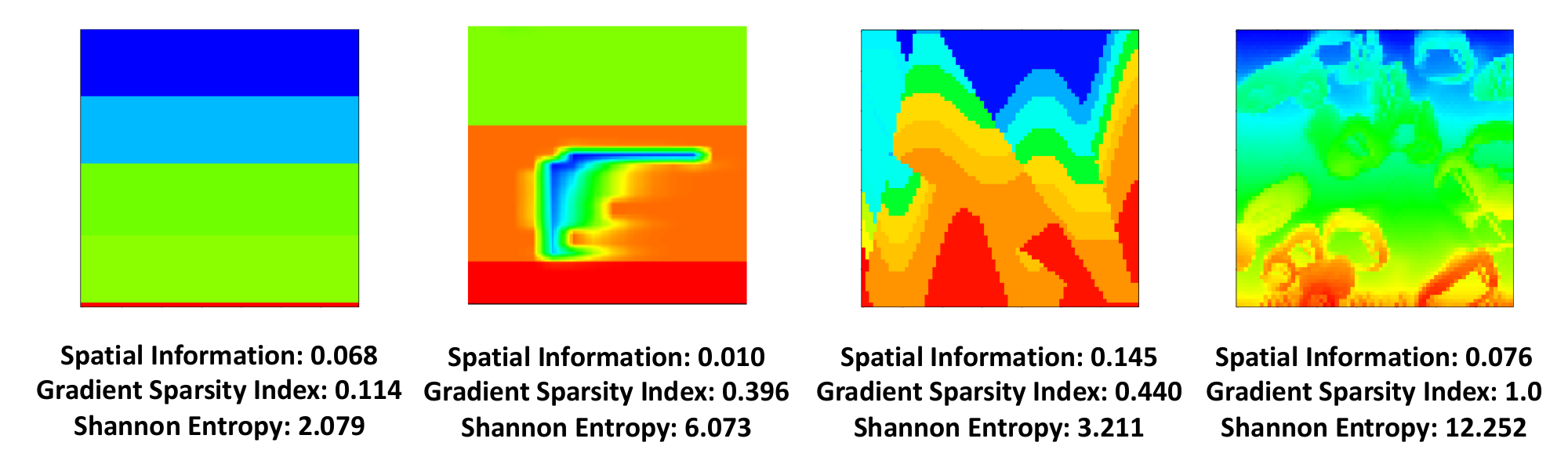}
\caption{\textbf{Examples of different levels of velocity map complexity from each dataset family}.}
\label{fig:com_vis}
\end{figure*}

The measurement of velocity map complexity sharpens our understanding of the datasets and hopefully becomes guidance for researchers to choose appropriate datasets. The numerical results are presented in \Cref{tab:complexity}. For more intuition, \Cref{fig:com_vis} illustrates velocity maps with the values of three metrics of their complexity. Based on the complexity, benchmark results, and our experience, beginning users may benefit from simple datasets (FlatVel-A, FlatVel-B, CurveVel-A, FlatFault-A, Curve-Fault-A, and Kimberlina-CO$_2$) while advanced solutions should be evaluated on challenging datasets (CurveVel-B, FlatFault-B, CurveFault-B, Style-A, and Style-B).  \par 

\begin{table}[h]
    \centering
    \footnotesize	
    \caption{\textbf{Velocity Map Complexity} of 2D Datasets}
    \vspace{0.5em}
    \begin{adjustbox}{width=1\textwidth}
    \setlength{\tabcolsep}{5.2mm}{
    \begin{tabular}{c|ccc}
    \specialrule{.15em}{.05em}{.05em} 
        {Dataset} &  {Spatial Information}  & {Gradient Sparsity Index} & {Shannon Entropy}  \\
    \specialrule{.1em}{.05em}{.05em} 
    FlatVel-A & 0.07 & 0.12 & 2.30 \\
    FlatVel-B & 0.34  & 0.13 & 2.32 \\
    CurveVel-A & 0.10  & 0.24 & 2.38\\
    CurveVel-B & 0.46  & 0.24 & 2.40 \\
    \specialrule{.05em}{.05em}{.05em}
    FlatFault-A & 0.10  & 0.26 & 2.92\\
    FlatFault-B & 0.16  & 0.26 & 3.45 \\
    CurveFault-A & 0.11  & 0.32 & 3.50\\
    CurveFault-B & 0.24  & 0.45 & 3.50 \\
    \specialrule{.05em}{.05em}{.05em}
    Style-A & 0.04  & 1.00 & 12.20 \\
    Style-B & 0.07  & 1.00 & 12.22\\
    \specialrule{.05em}{.05em}{.05em}
    Kimberlina-CO$_2$ & 0.01 & 0.54 & 7.35 \\
    \specialrule{.15em}{.05em}{.05em} 
    \end{tabular}}
    \label{tab:complexity}
    \end{adjustbox}
\end{table} 

We have several observations from \Cref{tab:complexity}. First, the three metrics are mostly consistent with each other for each dataset. Second, the datasets in version B always have higher complexity than version A, which meets our expectations that version B is the hard version. Third, the datasets with curve structures show higher complexity than the corresponding datasets with flat structures (e.g. CurveVel-A versus FlatVel-A).

The major limitation of this measurement is that none of the three metrics is able to provide a coherent evaluation across all datasets. For example, although ``\textit{Style Family}'' datasets have more complicated details than ``\textit{Vel}'' and ``\textit{Fault}'' family datasets, which is consistent with the Shannon-entropy, we cannot imply it from the spatial information. We remark that the discrepancy is due to the difference in data generation methods. In short, we emphasize that although we can justify the numerical results by cross-referencing between three metrics, it would be better to have one comprehensive analysis that utilizes all three metrics.

\section{Generalization Test}
\label{sup:gen}

In this section, we provide detailed results of the generalization test on InversionNet, VelocityGAN and UPFWI in ~\Cref{tab:gen_test_inv,tab:gen_test_vel,tab:gen_test_upfwi}, respectively. ``FVA'' is short for ``FlatVel-A'', and the same naming rule applies to the rest datasets. The blue, orange and green boxes indicate the \textit{intra-domain} tests with ``\textit{Vel Family}'', ``\textit{Fault Family}'' and ``\textit{Style Family}'', respectively. The lower triangle entries always have larger values than the upper entries for all three inversion methods, except there are a few outliers with InversionNet. The performance of VelocityGAN and UPFWI is better than the InversionNet, which is consistent with the results in~\cite{Jin-2021-Unsupervised,zhang2020data}. Moreover, InversionNet uses a larger batch size than the other two methods, which may degrade its generalization ability.

\begin{table}[h!]
\centering
\caption{\textbf{Generalization performance on 10 2D datasets with InversionNet}. The blue, orange and green boxes indicates the \textit{intra-domain} tests with ``\textit{Vel Family}'', ``\textit{Fault Family}'' and ``\textit{Style Family}''.}
\vspace{0.5em}
\label{tab:gen_test_inv}
\arrayrulecolor{black}
\begin{tabular}{c|cccc!{\color[rgb]{0.859,0.557,0}\vrule}cccccc} 
    \hline
\multicolumn{1}{c!{\color{blue}\vrule}}{\backslashbox{Source}{Target}}    & FVA              & FVB              & CVA              & \multicolumn{1}{c}{CVB}                                   & FFA              & FFB              & CFA              & CFB                                                                     & STA              & STB                                                                  \\ 
\cline{1-1}\arrayrulecolor{blue}\cline{2-5}\arrayrulecolor{black}\cline{6-11}
\multicolumn{1}{c!{\color{blue}\vrule}}{FVA} & \cellcolor{blue!30}-- & \cellcolor{blue!30}0.51             & \cellcolor{blue!30}0.55             & \multicolumn{1}{c!{\color{blue}\vrule}}{\cellcolor{blue!30}0.30}             & 0.84             & 0.52             & 0.66             & 0.43                                                                    & 0.51             & 0.50                                                                 \\
\multicolumn{1}{c!{\color{blue}\vrule}}{FVB} & \cellcolor{blue!30}0.42             & \cellcolor{blue!30}-- & \cellcolor{blue!30}0.48             & \multicolumn{1}{c!{\color{blue}\vrule}}{\cellcolor{blue!30}0.22}             & 0.34             & 0.51             & 0.36             & 0.42                                                                    & 0.27             & 0.20                                                                 \\
\multicolumn{1}{c!{\color{blue}\vrule}}{CVA} & \cellcolor{blue!30}0.44             & \cellcolor{blue!30}0.46             & \cellcolor{blue!30}-- & \multicolumn{1}{c!{\color{blue}\vrule}}{\cellcolor{blue!30}0.25}             & 0.39             & 0.54             & 0.42             & 0.42                                                                    & 0.39             & 0.29                                                                 \\
\multicolumn{1}{c!{\color{blue}\vrule}}{CVB} & \cellcolor{blue!30}0.69             & \cellcolor{blue!30}0.53             & \cellcolor{blue!30}0.52             & \multicolumn{1}{c!{\color{blue}\vrule}}{\cellcolor{blue!30}--} & 0.78             & 0.50             & 0.37             & 0.41                                                                    & 0.50             & 0.46                                                                 \\ 
\arrayrulecolor{blue}\cline{2-5}\arrayrulecolor[rgb]{0.859,0.557,0}\cline{6-9}
FFA                                                    & 0.86             & 0.17             & 0.23             & 0.22                                                               & \cellcolor{orange!30}-- & \cellcolor{orange!30}0.23             & \cellcolor{orange!30}0.46             & \multicolumn{1}{c!{\color[rgb]{1,0.659,0}\vrule}}{\cellcolor{orange!30}0.23}             & 0.30             & 0.34                                                                 \\
FFB                                                    & 0.42             & 0.45             & 0.50             & 0.19                                                               & \cellcolor{orange!30}0.40             & \cellcolor{orange!30}-- & \cellcolor{orange!30}0.41             & \multicolumn{1}{c!{\color[rgb]{1,0.659,0}\vrule}}{\cellcolor{orange!30}0.52}             & 0.39             & 0.33                                                                 \\
CFA                                                    & 0.09             & 0.10             & 0.15             & 0.05                                                               & \cellcolor{orange!30}0.14             & \cellcolor{orange!30}0.13             & \cellcolor{orange!30}-- & \multicolumn{1}{c!{\color[rgb]{1,0.659,0}\vrule}}{\cellcolor{orange!30}0.14}             & 0.56             & 0.51                                                                 \\
CFB                                                    & 0.38             & 0.40             & 0.50             & 0.19                                                               & \cellcolor{orange!30}0.38             & \cellcolor{orange!30}0.59             & \cellcolor{orange!30}0.42             & \multicolumn{1}{c!{\color[rgb]{1,0.659,0}\vrule}}{\cellcolor{orange!30}--} & 0.36             & 0.31                                                                 \\ 
\cline{6-9}\arrayrulecolor[rgb]{0,0.502,0}\cline{10-11}
STA                                                    & 0.44             & 0.38             & 0.59             & \multicolumn{1}{c}{0.20}                                           & 0.46             & 0.51             & 0.62             & \multicolumn{1}{c!{\color[rgb]{0,0.502,0}\vrule}}{0.42}             & \cellcolor{green!30}-- & \cellcolor{green!30}0.55                                                                 \\
STB                                                    & 0.48             & 0.28             & 0.42             & \multicolumn{1}{c}{0.23}                                           & 0.47             & 0.42             & 0.50             & \multicolumn{1}{c!{\color[rgb]{0,0.502,0}\vrule}}{0.35}             & \cellcolor{green!30}0.59             & \multicolumn{1}{c!{\color[rgb]{0,0.502,0}}}{\cellcolor{green!30}--}  \\
\arrayrulecolor{black}\cline{1-9}\arrayrulecolor[rgb]{0,0.502,0}\cline{10-11}
\end{tabular}
\arrayrulecolor{black}
\end{table}

\begin{table}[h!]
\centering
\caption{\textbf{Generalization performance on 10 2D datasets with VelocityGAN}. The blue, orange and green boxes indicates the \textit{intra-domain} tests with ``\textit{Vel Family}'', ``\textit{Fault Family}'' and ``\textit{Style Family}''.}
\vspace{0.5em}
\label{tab:gen_test_vel}
\arrayrulecolor{black}
\begin{tabular}{c|cccc!{\color[rgb]{0.859,0.557,0}\vrule}cccccc} 
    \hline
\multicolumn{1}{c!{\color{blue}\vrule}}{\backslashbox{Source}{Target}}    & FVA              & FVB              & CVA              & \multicolumn{1}{c}{CVB}                                   & FFA              & FFB              & CFA              & CFB                                                                     & STA              & STB                                                                  \\ 
\cline{1-1}\arrayrulecolor{blue}\cline{2-5}\arrayrulecolor{black}\cline{6-11}
\multicolumn{1}{c!{\color{blue}\vrule}}{FVA} & \cellcolor{blue!30}-- & \cellcolor{blue!30}0.50             & \cellcolor{blue!30}0.63             & \multicolumn{1}{c!{\color{blue}\vrule}}{\cellcolor{blue!30}0.36}             & 0.84             & 0.54             & 0.69             & 0.42                                                                    & 0.52             & 0.48                                                                 \\
\multicolumn{1}{c!{\color{blue}\vrule}}{FVB} & \cellcolor{blue!30}0.94             & \cellcolor{blue!30}-- & \cellcolor{blue!30}0.58             & \multicolumn{1}{c!{\color{blue}\vrule}}{\cellcolor{blue!30}0.38}             & 0.82             & 0.49             & 0.68             & 0.37                                                                    & 0.54             & 0.45                                                                 \\
\multicolumn{1}{c!{\color{blue}\vrule}}{CVA} & \cellcolor{blue!30}0.88             & \cellcolor{blue!30}0.50             & \cellcolor{blue!30}-- & \multicolumn{1}{c!{\color{blue}\vrule}}{\cellcolor{blue!30}0.44}             & 0.89             & 0.58             & 0.83             & 0.46                                                                    & 0.61             & 0.53                                                                 \\
\multicolumn{1}{c!{\color{blue}\vrule}}{CVB} & \cellcolor{blue!30}0.79             & \cellcolor{blue!30}0.75             & \cellcolor{blue!30}0.75             & \multicolumn{1}{c!{\color{blue}\vrule}}{\cellcolor{blue!30}--} & 0.86             & 0.54             & 0.79             & 0.42                                                                    & 0.60             & 0.51                                                                 \\ 
\arrayrulecolor{blue}\cline{2-5}\arrayrulecolor[rgb]{0.859,0.557,0}\cline{6-9}
FFA                                                    & 0.83             & 0.47             & 0.63             & 0.38                                                               & \cellcolor{orange!30}-- & \cellcolor{orange!30}0.56             & \cellcolor{orange!30}0.75             & \multicolumn{1}{c!{\color[rgb]{1,0.659,0}\vrule}}{\cellcolor{orange!30}0.44}             & 0.58             & 0.51                                                                 \\
FFB                                                    & 0.86             & 0.49             & 0.67             & 0.40                                                               & \cellcolor{orange!30}0.91             & \cellcolor{orange!30}-- & \cellcolor{orange!30}0.82             & \multicolumn{1}{c!{\color[rgb]{1,0.659,0}\vrule}}{\cellcolor{orange!30}0.53}             & 0.63             & 0.55                                                                 \\
CFA                                                    & 0.79             & 0.38             & 0.66             & 0.34                                                               & \cellcolor{orange!30}0.92             & \cellcolor{orange!30}0.58             & \cellcolor{orange!30}-- & \multicolumn{1}{c!{\color[rgb]{1,0.659,0}\vrule}}{\cellcolor{orange!30}0.48}             & 0.58             & 0.50                                                                 \\
CFB                                                    & 0.70             & 0.43             & 0.67             & 0.38                                                               & \cellcolor{orange!30}0.79             & \cellcolor{orange!30}0.64             & \cellcolor{orange!30}0.60             & \multicolumn{1}{c!{\color[rgb]{1,0.659,0}\vrule}}{\cellcolor{orange!30}--} & 0.62            & 0.53                                                                 \\ 
\cline{6-9}\arrayrulecolor[rgb]{0,0.502,0}\cline{10-11}
STA                                                    & 0.66             & 0.37             & 0.56             & \multicolumn{1}{c}{0.32}                                           & 0.67             & 0.53             & 0.63             & \multicolumn{1}{c!{\color[rgb]{0,0.502,0}\vrule}}{0.43}             & \cellcolor{green!30}-- & \cellcolor{green!30}0.66                                                                \\
STB                                                    & 0.52             & 0.29             & 0.47             & \multicolumn{1}{c}{0.25}                                           & 0.51             & 0.47             & 0.50             & \multicolumn{1}{c!{\color[rgb]{0,0.502,0}\vrule}}{0.39}             & \cellcolor{green!30}0.70             & \multicolumn{1}{c!{\color[rgb]{0,0.502,0}}}{\cellcolor{green!30}--}  \\
\arrayrulecolor{black}\cline{1-9}\arrayrulecolor[rgb]{0,0.502,0}\cline{10-11}
\end{tabular}
\arrayrulecolor{black}
\end{table}

\begin{table}[h!]
\centering
\caption{\textbf{Generalization performance on 10 2D datasets with UPFWI}. The blue, orange and green boxes indicates the \textit{intra-domain} tests with ``\textit{Vel Family}'', ``\textit{Fault Family}'' and ``\textit{Style Family}''.}
\vspace{0.5em}
\label{tab:gen_test_upfwi}
\arrayrulecolor{black}
\begin{tabular}{c|cccc!{\color[rgb]{0.859,0.557,0}\vrule}cccccc} 
    \hline
\multicolumn{1}{c!{\color{blue}\vrule}}{\backslashbox{Source}{Target}}    & FVA              & FVB              & CVA              & \multicolumn{1}{c}{CVB}                                   & FFA              & FFB              & CFA              & CFB                                                                     & STA              & STB                                                                  \\ 
\cline{1-1}\arrayrulecolor{blue}\cline{2-5}\arrayrulecolor{black}\cline{6-11}
\multicolumn{1}{c!{\color{blue}\vrule}}{FVA} & \cellcolor{blue!30}-- & \cellcolor{blue!30}0.49             & \cellcolor{blue!30}0.63             & \multicolumn{1}{c!{\color{blue}\vrule}}{\cellcolor{blue!30}0.28}             & 0.84             & 0.54             & 0.70             & 0.41                                                                    & 0.58             & 0.53                                                                 \\
\multicolumn{1}{c!{\color{blue}\vrule}}{FVB} & \cellcolor{blue!30}0.64             & \cellcolor{blue!30}-- & \cellcolor{blue!30}0.56             & \multicolumn{1}{c!{\color{blue}\vrule}}{\cellcolor{blue!30}0.28}             & 0.49             & 0.49             & 0.47             & 0.37                                                                   & 0.45             & 0.34                                                                 \\
\multicolumn{1}{c!{\color{blue}\vrule}}{CVA} & \cellcolor{blue!30}0.71             & \cellcolor{blue!30}0.48             & \cellcolor{blue!30}-- & \multicolumn{1}{c!{\color{blue}\vrule}}{\cellcolor{blue!30}0.38}             & 0.68             & 0.55             & 0.69             & 0.45                                                                    & 0.52             & 0.43                                                                 \\
\multicolumn{1}{c!{\color{blue}\vrule}}{CVB} & \cellcolor{blue!30}0.70             & \cellcolor{blue!30}0.57             & \cellcolor{blue!30}0.67             & \multicolumn{1}{c!{\color{blue}\vrule}}{\cellcolor{blue!30}--} & 0.78             & 0.53             & 0.75             & 0.44                                                                    & 0.58             & 0.49                                                                 \\ 
\arrayrulecolor{blue}\cline{2-5}\arrayrulecolor[rgb]{0.859,0.557,0}\cline{6-9}
FFA                                                    & 0.90             & 0.50             & 0.60             & 0.37                                                               & \cellcolor{orange!30}-- & \cellcolor{orange!30}0.53             & \cellcolor{orange!30}0.72             & \multicolumn{1}{c!{\color[rgb]{1,0.659,0}\vrule}}{\cellcolor{orange!30}0.42}             & 0.54             & 0.49                                                                 \\
FFB                                                    & 0.66             & 0.50             & 0.61             & 0.34                                                               & \cellcolor{orange!30}0.62             & \cellcolor{orange!30}-- & \cellcolor{orange!30}0.62             & \multicolumn{1}{c!{\color[rgb]{1,0.659,0}\vrule}}{\cellcolor{orange!30}0.47}             & 0.55             & 0.45                                                                 \\
CFA                                                    & 0.76             & 0.50             & 0.63             & 0.38                                                               & \cellcolor{orange!30}0.81             & \cellcolor{orange!30}0.56             & \cellcolor{orange!30}-- & \multicolumn{1}{c!{\color[rgb]{1,0.659,0}\vrule}}{\cellcolor{orange!30}0.46}             & 0.60             & 0.50                                                                 \\
CFB                                                    & 0.52             & 0.34             & 0.48             & 0.29                                                               & \cellcolor{orange!30}0.53             & \cellcolor{orange!30}0.43             & \cellcolor{orange!30}0.52             & \multicolumn{1}{c!{\color[rgb]{1,0.659,0}\vrule}}{\cellcolor{orange!30}--} & 0.51            & 0.39                                                                 \\ 
\cline{6-9}\arrayrulecolor[rgb]{0,0.502,0}\cline{10-11}
STA                                                    & 0.54             & 0.36             & 0.53             & \multicolumn{1}{c}{0.28}                                           & 0.51             & 0.48             & 0.53             & \multicolumn{1}{c!{\color[rgb]{0,0.502,0}\vrule}}{0.41}             & \cellcolor{green!30}-- & \cellcolor{green!30}0.58                                                                \\
STB                                                    & 0.49             & 0.30             & 0.45             & \multicolumn{1}{c}{0.23}                                           & 0.48             & 0.42             & 0.48             & \multicolumn{1}{c!{\color[rgb]{0,0.502,0}\vrule}}{0.35}             & \cellcolor{green!30}0.63             & \multicolumn{1}{c!{\color[rgb]{0,0.502,0}}}{\cellcolor{green!30}--}  \\
\arrayrulecolor{black}\cline{1-9}\arrayrulecolor[rgb]{0,0.502,0}\cline{10-11}
\end{tabular}
\arrayrulecolor{black}
\end{table}


\section{Uncertainty Quantification}
\label{sup:uncertainty}
\begin{figure*}[h]
\centering
\includegraphics[width=1.0\textwidth]{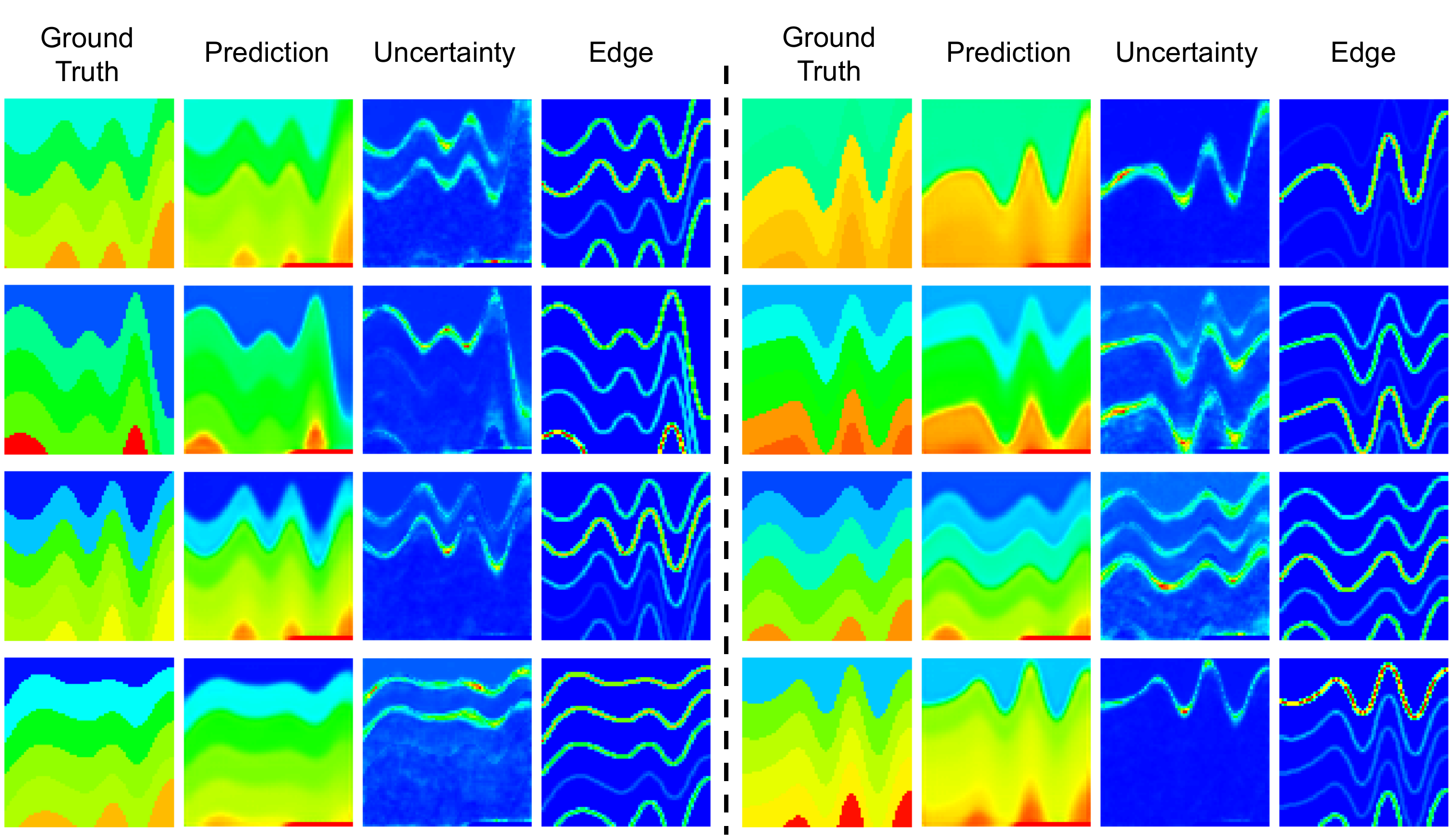}
\caption{\textbf{Uncertainty visualization}. The uncertainty is higher on the boundaries compared with other regions.}
\label{fig:uncertainty}
\end{figure*}

We further conduct experiments on CurveVel-A to quantify uncertainty in InversionNet as a case study. Following \cite{kendall2017uncertainties}, we modify the network architecture by adding a dropout layer with dropout rate $p=0.2$ after each convolutional layer except the last one.
As shown in \Cref{fig:uncertainty}, The uncertainty on boundaries is higher than in other regions, which implies the prediction sensitivity around the boundaries. To quantify the correlation between the uncertainty and boundaries, we calculate the Pearson correlation coefficient between the uncertainty value and the gradient magnitude on the edge. The value is 0.5462, indicating a moderate positive correlation as shown in \Cref{fig:uncertain_scatter}. As illustrated in Table~\ref{tab:uncertain_table}, the uncertainty increases gradually when increasing the noise level. We also include the average peak-to-noise ratio (PSNR) in the table. The PSNR of a sample is defined as 
\begin{equation}
    \text{PSNR} = 10\log_{10}\frac{(p_{max} - p_{min})^2}{\ell_2(p - p')},
\end{equation}
where $p_{max}$ and $p_{min}$ denote the maximum and minimum possible values of the seismic data in a dataset, $p$ is the clean seismic data, and $p'$ is the noisy data. 
Moreover, the uncertainty values of cross datasets are much higher than training and testing on the same dataset, which indicates that domain shifts lead to an increase in uncertainty as compared in Table~\ref{table:uq_generalize}. 

\begin{minipage}{\textwidth}
  \begin{minipage}[h]{0.4\textwidth}
    \centering
    \includegraphics[width=1\textwidth]{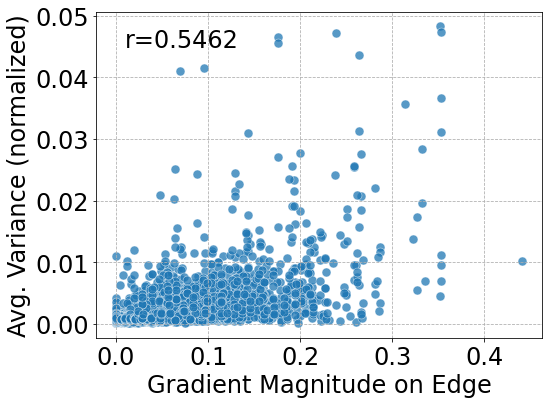}
    \captionof{figure}{\textbf{The correlation between the mean variance and the gradient magnitude on velocity edge}.}
    \label{fig:uncertain_scatter}
  \end{minipage}
  \hfill
  \begin{minipage}[h]{0.55\textwidth}
    \centering
    \captionof{table}{\textbf{Quantitative results of uncertainty quantification on noisy seismic input.} Gaussian noise with different standard deviation is added to seismic data during testing ($\sigma_{test}$). }
    \renewcommand{\arraystretch}{1.2}
    \begin{adjustbox}{width=1\textwidth}
    \begin{tabular}{c | c | c  }
    \hline
    $\sigma_{test}$ ($10^{-4}$) & PSNR (dB) & Variance ($10^{-3}$) \\
    \hline
    0 & 100.00 & 1.3903 \\
    \hline
    0.1 & 70.41 & 1.3970 \\
    \hline
    0.5 & 63.40 & 1.4269 \\
    \hline
    1.0 & 60.38 & 1.4479 \\
    \hline
    5.0 & 53.32 & 1.5621 \\
    \hline
    \end{tabular}
    \label{tab:uncertain_table}
    \end{adjustbox}
    \end{minipage}
 \end{minipage}

\begin{table}
\renewcommand{\arraystretch}{1.5}
\setlength{\tabcolsep}{3pt}
\footnotesize
\centering
\caption{\textbf{Quantitative results of mean variance on 2D datasets.} The model is trained on CurveVel-A. }
\vspace{0.5em}
\begin{adjustbox}{width=1\textwidth}
\begin{tabular}{c|c|c|c|c|c|c|c|c|c|c}
\hline
Dataset & FlatVel-A & FlatVel-B & CurveVel-A & CurveVel-B & FlatFault-A & FlatFault-B & CurveFault-A & CurveFault-B & Style-A & Style-B \\
\hline
Variance ($10^{-3}$) & 2.6065 & 4.1899 & 1.3903 &3.5000 & 4.2857 & 2.4679 & 2.3639 & 2.3142 & 2.9280 & 2.9946 \\
\hline
\end{tabular}
\end{adjustbox}
\vspace{0.5em}
\label{table:uq_generalize}
\vspace{-1em}
\end{table}

\section{Robustness Test}
\label{sup:robust}
In this section, we provide the quantitative results, visualization, and detailed analysis of the robustness test. Table~\ref{table:noise_testing} shows averaged performance across all 2D datasets (Kimberlina-CO$_2$ excluded) on both clean and noisy data. It is illustrated that InversionNets trained with $\ell_1$ loss and $\ell_2$ loss are both the most sensitive to noise compared to other models. 

The results of 3D robustness test are also summarized in Table~\ref{table:noise_testing}. We average the performance of the three-channel selections in the benchmark result. We observe that both the $\ell_1$- and $\ell_2$-trained InversionNet3D models are robust to Gaussian noise. The MAE/RMSE increase by $175\%$ on average compared to clean data benchmark results, mainly due to the clean MAE/RMSE being already small. The SSIM slightly decreases by a maximum of $1.76\%$ due to the background velocity of 3D Kimberlina-V1 remaining unchanged for all the samples which have more weight than the reservoir region when calculating SSIM.



\begin{figure*}[t]
\centering
\includegraphics[width=1\columnwidth]{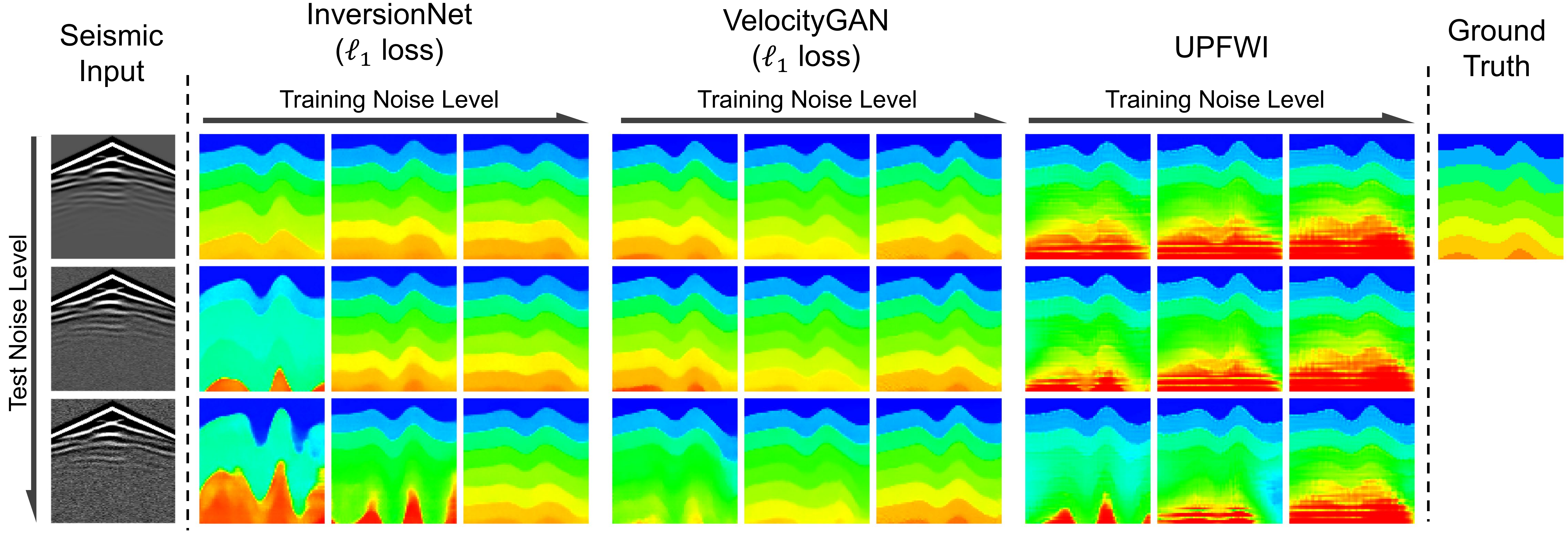}
\caption{\textbf{Comparison of inversion results generated by different models trained with various levels of noise on CurveVel-A}. When facing noisy seismic input, the models trained with corresponding level of noise yield more accurate inversion results compared to those trained on clean data. }
\label{fig:noise_training}
\end{figure*}

In addition to adding noise during testing, we are also interested in whether training models on noisy data can improve robustness. We train our models on CurveVel-A with two levels of Gaussian noise ($\sigma_{train}=1\times10^{-4}$ and $\sigma_{train}=5\times10^{-4}$) added separately to seismic data. Table~\ref{table:noise_training} shows quantitative results of those models on both clean data and noisy data. We observe that introducing noise into training effectively improves model robustness for all three models when the testing noise level is high. When the testing noise level is lower than the training noise level, results are mixed. Specifically, InversionNet trained with noisy data achieves better performance than the one trained on clean data, while the other two models (VelocityGAN and UPFWI) have a slight performance drop. Visualization of results is shown in~\Cref{fig:noise_training}.

\begin{table}
\renewcommand{\arraystretch}{1.5}
\renewcommand{\bfdefault}{sb}
\setlength{\tabcolsep}{3pt}
\centering
\caption{\textbf{Quantitative results of different models on 2D and 3D datasets with noisy seismic input (Kimberlina-CO$_2$ excluded)}. Gaussian noise with different standard deviation $\sigma_{test}$ is added to seismic data during testing. Performance in all three metrics is averaged across all datasets for 2D models. Performance is averaged across all channel selections for InversionNet3D. The least degradation is highlighted.}
\vspace{0.5em}
\begin{adjustbox}{width=1\textwidth}
\begin{tabular}{p{3cm} | c c c | c c c  | c c c |c c c | c c c} 
\thickhline

\multirow{3}{*}{Method} & 
\multicolumn{3}{c|}{\multirow{2}{*}{$\sigma_{test}=0$}} & \multicolumn{3}{c|}{$\sigma_{test}=1\times10^{-5}$} & \multicolumn{3}{c|}{$\sigma_{test}=5\times10^{-5}$} & \multicolumn{3}{c|}{$\sigma_{test}=1\times10^{-4}$} & \multicolumn{3}{c}{$\sigma_{test}=5\times10^{-4}$} \\
& & & & 
\multicolumn{3}{c|}{PSNR=70.41dB} & 
\multicolumn{3}{c|}{PSNR=63.40dB} & 
\multicolumn{3}{c|}{PSNR=60.38dB} & 
\multicolumn{3}{c}{PSNR=53.32dB} \\

\cline{2-16}
& MAE$\downarrow$ & RMSE$\downarrow$ & SSIM$\uparrow$ & MAE$\downarrow$ & RMSE$\downarrow$ & SSIM$\uparrow$ & MAE$\downarrow$ & RMSE$\downarrow$ & SSIM$\uparrow$ & MAE$\downarrow$ & RMSE$\downarrow$ & SSIM$\uparrow$ & MAE$\downarrow$ & RMSE$\downarrow$ & SSIM$\uparrow$ \\ 
\thickhline
InversionNet-$\ell_1$   & 0.0708 & 0.1316 & 0.8187 & 0.0769 & 0.1381 & 0.8131 & 0.1142 & 0.1879 & 0.7779 & 0.1519 & 0.2362 & 0.7419 & 0.2418 & 0.3530 & 0.6387\\ 
\multicolumn{1}{r|}{Degradation(\%)} & $\backslash$ & $\backslash$ & $\backslash$ & -8.66 & -4.97 & -0.69 & -61.32 & -42.78 & -4.98 & -114.70 & -79.48 & -9.38 & -241.70 & -168.26 & -21.99\\ 
\hline

InversionNet-$\ell_2$  & 0.0727 & 0.1211 & 0.8333 & 0.0788 & 0.1280 & 0.8290 & 0.1028 & 0.1609 & 0.8076 & 0.1348 & 0.2048 & 0.7784 & 0.2301 & 0.3309 & 0.6802\\ 
\multicolumn{1}{r|}{Degradation(\%)} & $\backslash$ & $\backslash$ & $\backslash$ & -8.45 & -5.72 & -0.51 & -41.48 & -32.91 & -3.08 & -85.54 & -69.14 & -6.59 & -216.63 & -173.35 & -18.37\\ 
\hline

VelocityGAN-$\ell_1$  & 0.0713 & 0.1286 & 0.8341 & 0.0725 & 0.1292 & 0.8329 & 0.0825 & 0.1416 & 0.8247 & 0.0951 & 0.1581 & 0.8147 & 0.1644 & 0.2526 & 0.7451\\ 
\multicolumn{1}{r|}{Degradation(\%)} & $\backslash$ & $\backslash$ & $\backslash$ & -1.73 & -0.47 & \textbf{-0.14} & -15.65 & -10.11 & -1.12 & -33.35 & -22.89 & -2.33 & -130.60 & -96.35 & \textbf{-10.67}\\ 
\hline

VelocityGAN-$\ell_2$  & 0.0723 & 0.1209 & 0.8380 & 0.0722 & 0.1208 & 0.8364 & 0.0773 & 0.1273 & 0.8297 & 0.0858 & 0.1380 & 0.8212 & 0.1818 & 0.2722 & 0.7444\\ 
\multicolumn{1}{r|}{Degradation(\%)} & $\backslash$ & $\backslash$ & $\backslash$ & +0.11 & +0.06 & -0.19 & \textbf{-6.91} & \textbf{-5.33} & -\textbf{0.99} & \textbf{-18.60} & \textbf{-14.18} & \textbf{-2.01} & -151.38 & -125.17 & -11.17\\ 
\hline

UPFWI & 0.1326 & 0.2252 & 0.7716 & 0.1324 & 0.2241 & 0.7700 & 0.1437 & 0.2395 & 0.7560 & 0.1639 & 0.2680 & 0.7360 & 0.2286 & 0.3503 & 0.6725\\ 
\multicolumn{1}{r|}{Degradation(\%)} & $\backslash$ & $\backslash$ & $\backslash$ & \textbf{+0.14} & \textbf{+0.48} & -0.21 & -8.43 & -6.33 & -2.01 & -23.66 & -18.97 & -4.61 & \textbf{-72.45} & \textbf{-55.51} & -12.85\\ 
\hline

InversionNet3D-$\ell_1$ & 0.0107 & 0.0281 & 0.9837 & 0.0319 & 0.0719 & 0.9665 & 0.0320 & 0.0720 & 0.9665 & 0.0319 & 0.0719 & 0.9664 & 0.0340 & 0.0759 & 0.9731\\ 
\multicolumn{1}{r|}{Degradation(\%)} & $\backslash$ & $\backslash$ & $\backslash$ & -199.21 & -155.41 & -1.75 & -199.78 & -155.86 & -1.75 & -199.37 & -155.51 & -1.76 & -218.59 & -169.73 & -1.08\\ 
\hline

InversionNet3D-$\ell_2$ & 0.0155 & 0.0306 & 0.9462 & 0.0384 & 0.0867 & 0.9355 & 0.0388 & 0.0877 & 0.9353 & 0.0390 & 0.0879 & 0.9352 & 0.0411 & 0.0924 & 0.9346\\ 
\multicolumn{1}{r|}{Degradation(\%)} & $\backslash$ & $\backslash$ & $\backslash$ & -148.51 & -182.93 & -1.13 & -151.00 & -186.12 & -1.15 & -152.21 & -186.85 & -1.16 & -165.80 & -201.41 & -1.22\\ 
\thickhline

\end{tabular}
\end{adjustbox}
\vspace{0.5em}
\label{table:noise_testing}
\vspace{-1em}
\end{table}


\begin{table}
\renewcommand{\arraystretch}{1.5}
\renewcommand{\bfdefault}{sb}
\setlength{\tabcolsep}{3pt}
\centering
\caption{\textbf{Quantitative results of different models trained with noisy seismic data on CurveVel-A.} Gaussian noise with different standard deviation is added to seismic data during training ($\sigma_{train}$) and testing ($\sigma_{test}$)  }
\vspace{0.5em}
\begin{adjustbox}{width=1\textwidth}
\begin{tabular}{l|c|c c c|c c c|c c c|c c c|c c c} 
\thickhline
\multirow{2}{*}{Method} & \multirow{2}{*}{\makecell{$\sigma_{train}$\\($10^{-4}$)}} & \multicolumn{3}{c|}{$\sigma_{test}=0$} & \multicolumn{3}{c|}{$\sigma_{test}=1\times10^{-5}$} & \multicolumn{3}{c|}{$\sigma_{test}=5\times10^{-5}$} & \multicolumn{3}{c|}{$\sigma_{test}=1\times10^{-4}$} & \multicolumn{3}{c}{$\sigma_{test}=5\times10^{-4}$} \\
\cline{3-17}
& & MAE$\downarrow$ & RMSE$\downarrow$ & SSIM$\uparrow$ & MAE$\downarrow$ & RMSE$\downarrow$ & SSIM$\uparrow$ & MAE$\downarrow$ & RMSE$\downarrow$ & SSIM$\uparrow$ & MAE$\downarrow$ & RMSE$\downarrow$ & SSIM$\uparrow$ & MAE$\downarrow$ & RMSE$\downarrow$ & SSIM$\uparrow$ \\ 
\thickhline
\multirow{3}{*}{InversionNet-$\ell_1$} & 0 & 0.0685 & 0.1273 & 0.8074 & 0.0721 & 0.1315 & 0.8006 & 0.1530 & 0.2536 & 0.7248 & 0.1609 & 0.3600 & 0.6751 & 0.2401 & 0.3387 & 0.6081 \\
& 1 & 0.0675 & 0.1238 & 0.8242 & 0.0672 & 0.1236 & 0.8245 & 0.0674 & 0.1238 & 0.8246 & 0.0676 & 0.1241 & 0.8246 & 0.0780 & 0.1391 & 0.8036 \\
& 5 & \textbf{0.0657} & \textbf{0.1228} & \textbf{0.8340} & \textbf{0.0659} & \textbf{0.1230} & \textbf{0.8338} & \textbf{0.0662} & \textbf{0.1234} & \textbf{0.8336} & \textbf{0.0663} & \textbf{0.1234} & \textbf{0.8338} & \textbf{0.0637} & \textbf{0.1205} & \textbf{0.8376} \\
\hline

\multirow{3}{*}{InversionNet-$\ell_2$} & 0 & 0.0691 & 0.1202 & 0.8223 & 0.0759 & 0.1273 & 0.8178 & 0.0830 & 0.1364 & 0.7989 & 0.1185 & 0.1839 & 0.7523 & 0.2530 & 0.3599 & 0.5887 \\
& 1 & \textbf{0.0665} & \textbf{0.1165} & \textbf{0.8402} & \textbf{0.0662} & \textbf{0.1162} & \textbf{0.8405} & \textbf{0.0656} & \textbf{0.1155} & \textbf{0.8412} & \textbf{0.0655} & \textbf{0.1155} & \textbf{0.8409} & 0.0894 & 0.1449 & 0.8024 \\
& 5 & 0.0687 & 0.1198 & 0.8360 & 0.0686 & 0.1197 & 0.8362 & 0.0683 & 0.1194 & 0.8366 & 0.0680 & 0.1189 & 0.8372 & \textbf{0.0668} & \textbf{0.1175} & \textbf{0.8391} \\
\hline

\multirow{3}{*}{VelocityGAN-$\ell_1$} & 0& \textbf{0.0483} & \textbf{0.1034} & \textbf{0.8625} & \textbf{0.0490} & \textbf{0.1042} & \textbf{0.8615} & 0.0518 & 0.1070 & 0.8583 & 0.0557 & 0.1111 & 0.8546 & 0.1331 & 0.2166 & 0.7969 \\
& 1 & 0.0534 & 0.1083 & 0.8630 & 0.0529 & 0.1078 & 0.8632 & \textbf{0.0517} & \textbf{0.1065} & \textbf{0.8638} & \textbf{0.0512} & \textbf{0.1060} & \textbf{0.8639} & 0.0591 & 0.1156 & 0.8534 \\
& 5 & 0.0524 & 0.1081 & 0.8618 & 0.0524 & 0.1081 & 0.8619 & 0.0523 & 0.1079 & 0.8621 & 0.0520 & 0.1077 & 0.8624 & \textbf{0.0516} & \textbf{0.1073} & \textbf{0.8627} \\
\hline

\multirow{3}{*}{VelocityGAN-$\ell_2$} & 0 & \textbf{0.0510} & \textbf{0.0976} & \textbf{0.8759} & \textbf{0.0519} & \textbf{0.0985} & \textbf{0.8752} & \textbf{0.0559} & \textbf{0.1027} & \textbf{0.8721} & 0.0651 & 0.1136 & 0.8652 & 0.2025 & 0.3103 & 0.7643 \\
& 1 & 0.0627 & 0.1122 & 0.8333 & 0.0622 & 0.1117 & 0.8333 & 0.0608 & 0.1104 & 0.8335 & \textbf{0.0603} & \textbf{0.1098} & \textbf{0.8333} & 0.0666 & 0.1170 & 0.8258 \\
& 5 & 0.0618 & 0.1072 & 0.8083 & 0.0617 & 0.1070 & 0.8084 & 0.0613 & 0.1065 & 0.8089 & 0.0610 & 0.1061 & 0.8093 & \textbf{0.0606} & \textbf{0.1054} & \textbf{0.8100} \\
\hline

\multirow{3}{*}{UPFWI} & 0 & \textbf{0.0805} & \textbf{0.1411} & \textbf{0.8443} & \textbf{0.0798} & \textbf{0.1411} & \textbf{0.8437} & \textbf{0.0835} & \textbf{0.1481} & \textbf{0.8379} & 0.0927 & 0.1620 & 0.8292 & 0.1897 & 0.2988 & 0.7683 \\
& 1 & 0.0869 & 0.1604 & 0.8278 & 0.0870 & 0.1608 & 0.8279 & 0.0877 & 0.1632 & 0.8280 & \textbf{0.0894} & \textbf{0.1676} & \textbf{0.8272} & 0.1206 & 0.2241 & 0.8057 \\
& 5 & 0.0923 & 0.1671 & 0.8233 & 0.0922 & 0.1671 & 0.8233 & 0.0921 & 0.1673 & 0.8233 & 0.0920 & 0.1676 & 0.8233 & 0.\textbf{0953} & \textbf{0.1785} & \textbf{0.8188} \\
\thickhline

\end{tabular}
\end{adjustbox}
\vspace{0.5em}
\label{table:noise_training}
\vspace{-1em}
\end{table}

\section{Comparison with Physics-driven Methods}
\label{sup:method-compare}
In this section, we select $500$ samples from each 2D dataset and $1$ samples from the 3D dataset to test the performance of physics-driven methods. Multi-scale full waveform inversion~\cite{bunks1995multiscale,brossier2009seismic} is performed with the six different low-pass ﬁlters that the cutoff frequencies equal to $(1, 3, 5, 10, 20, 30)$ $Hz$. We use three different initial maps in the inversion: homogeneous maps, linear increasing maps, and smoothed maps. Homogeneous maps are built with a constant velocity equal to minimum velocity on the surface. The velocity increases linearly with depth in linear increasing maps. Smoothed maps are obtained by applying mean filters to the ground truth velocity maps. $\ell_2$-norm loss function and the conjugated gradient method are used for the iterative updating of the velocity maps. The inversion stops when the loss change is less than $0.1\%$. 

\begin{figure*}[h!]
\centering
\includegraphics[width=1\columnwidth]{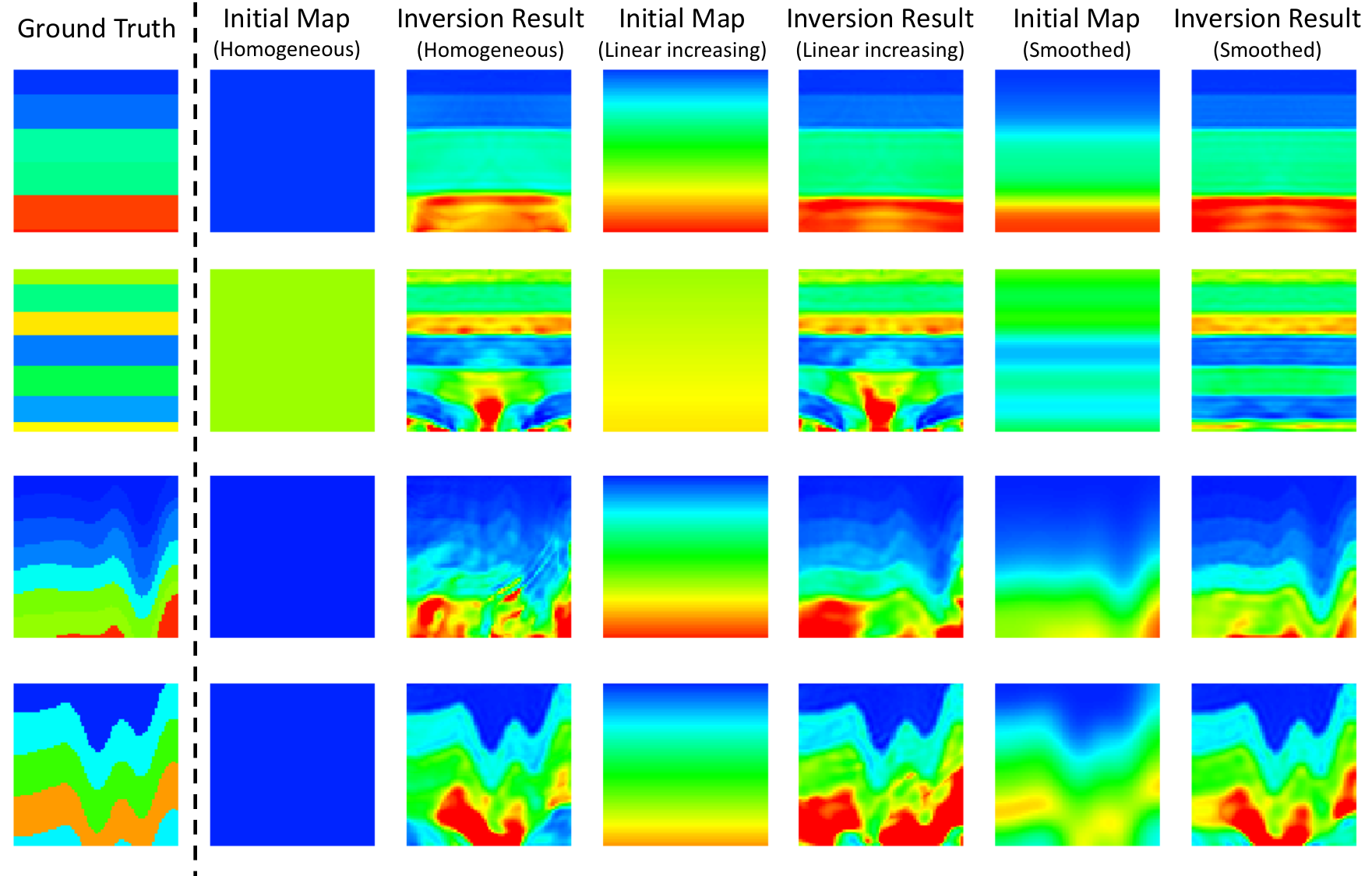}
\caption{\textbf{Example of ground truth and physics-driven inversion results in ``\textit{Vel Family}''}. The left part shows the ground truth velocity map, and the right demonstrates the initial maps and inversion results with homogeneous, linear increasing, and smoothed initial maps. From the first row to the last: FlatVel-A, FlatVel-B, CurveVel-A, CurveVel-B.  }
\label{fig:vel_fwi}
\end{figure*}

\begin{figure*}[h!]
\centering
\includegraphics[width=1\columnwidth]{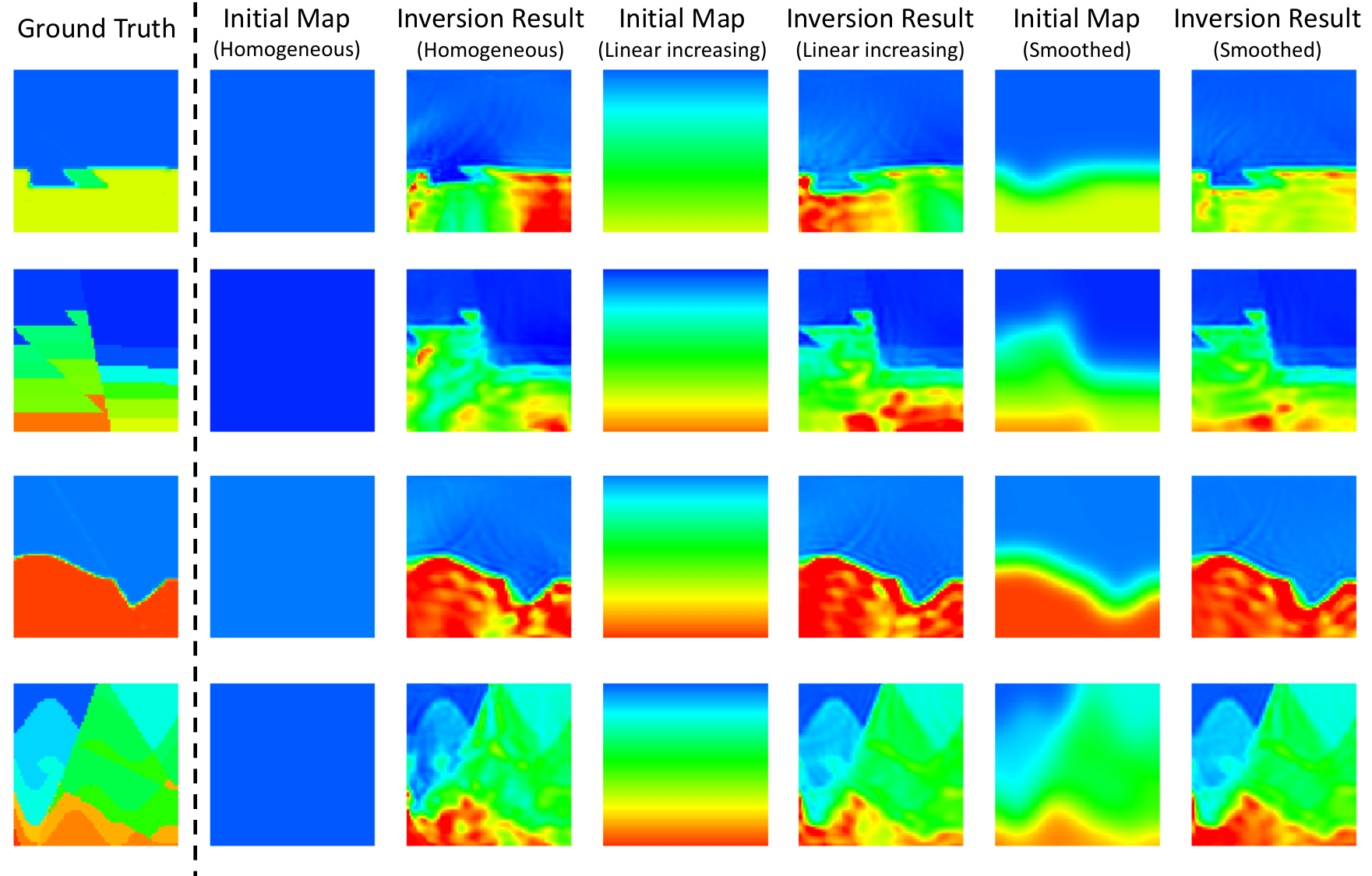}
\caption{\textbf{Example of ground truth and physics-driven inversion results in ``\textit{Fault Family}''}. The left part shows the ground truth velocity map, and the right demonstrates the initial maps and inversion results with homogeneous, linear increasing and smoothed initial maps. From the first row to the last: FlatFault-A, FlatFault-B, CurveFault-A, CurveFault-B.  }
\label{fig:fault_fwi}
\end{figure*}

\begin{figure*}[h!]
\centering
\includegraphics[width=1\columnwidth]{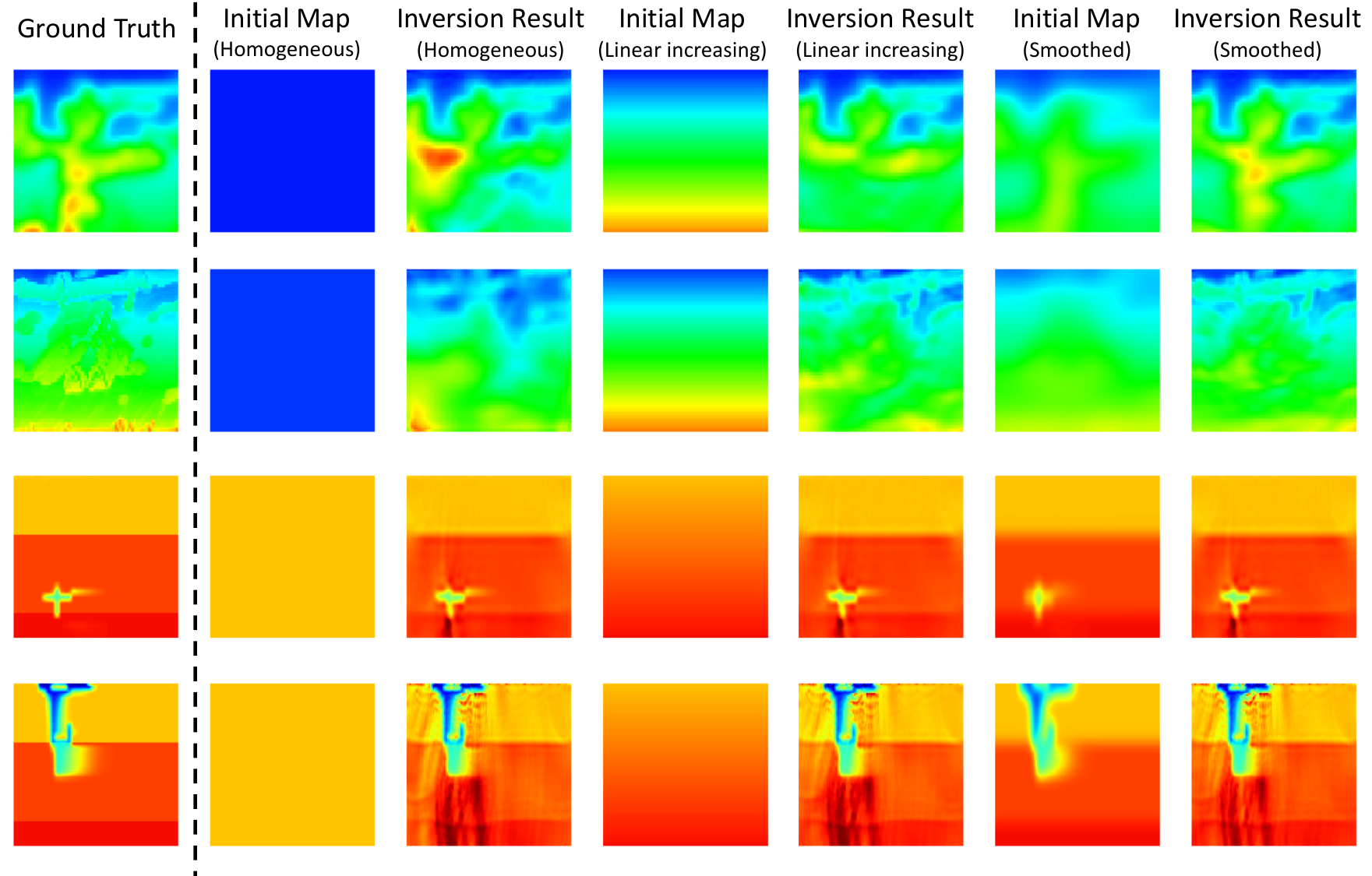}
\caption{\textbf{Example of ground truth and physics-driven inversion results in ``\textit{Style Family}''}. The left part shows the ground truth velocity map, and the right demonstrates the initial maps and inversion results with homogeneous, linear increasing, and smoothed initial maps. First two rows: Style-A and Style-B, last two rows: Kimberlina-CO$_2$.  }
\label{fig:style_fwi}
\end{figure*}

\begin{figure*}[h!]
\centering
\includegraphics[width=1\columnwidth]{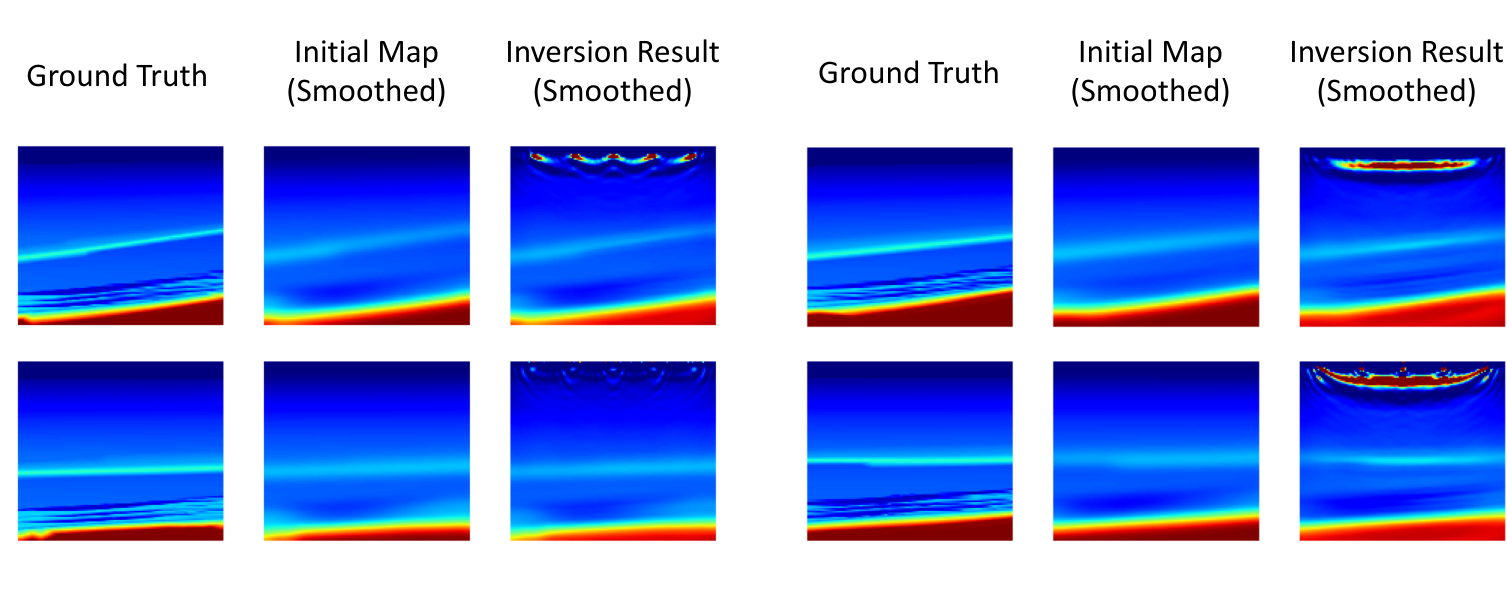}
\caption{\textbf{Example of ground truth and physics-driven inversion results in ``\textit{3D Kimberlina-V1}''}. The ground truth velocity map, the smoothed initial maps and inversion results. }
\label{fig:kim_fwi_3d}
\end{figure*}

Quantitative results of the physics-driven methods are given in~\Cref{table:physicsFWI_Benchmark} and their illustrations are presented in~\Cref{fig:vel_fwi,fig:fault_fwi,fig:style_fwi,fig:kim_fwi_3d}. Compared with data-driven FWI results in~\Cref{fig:vel_result,fig:fault_result,fig:style_result,fig:kim_result}, physics-driven FWI results have more artifacts and the performance of physics-driven FWI strongly depends on the quality of initial maps. The initial map with higher SSIM leads to FWI results with higher SSIM. The building of the initial maps is very important and how to obtain a good initial map is still an open research topic~\cite{prieux2013building,datta2016estimating,teodor2021challenges}. For 3D Kimberlina-V1 data, the loss is hard to converge with homogeneous and linear increasing initial maps due to their poor qualities. The inversion with smoothed initial maps is greatly affected by the acquisition footprint~\cite{chopra2000acquisition}. As a result, the inversion results have strong artifacts at the surface in~\Cref{fig:kim_fwi_3d} and their SSIM is even lower than the initial maps in~\Cref{table:physicsFWI_Benchmark}.

\begin{figure*}[h!]
\centering
\includegraphics[width=1\columnwidth]{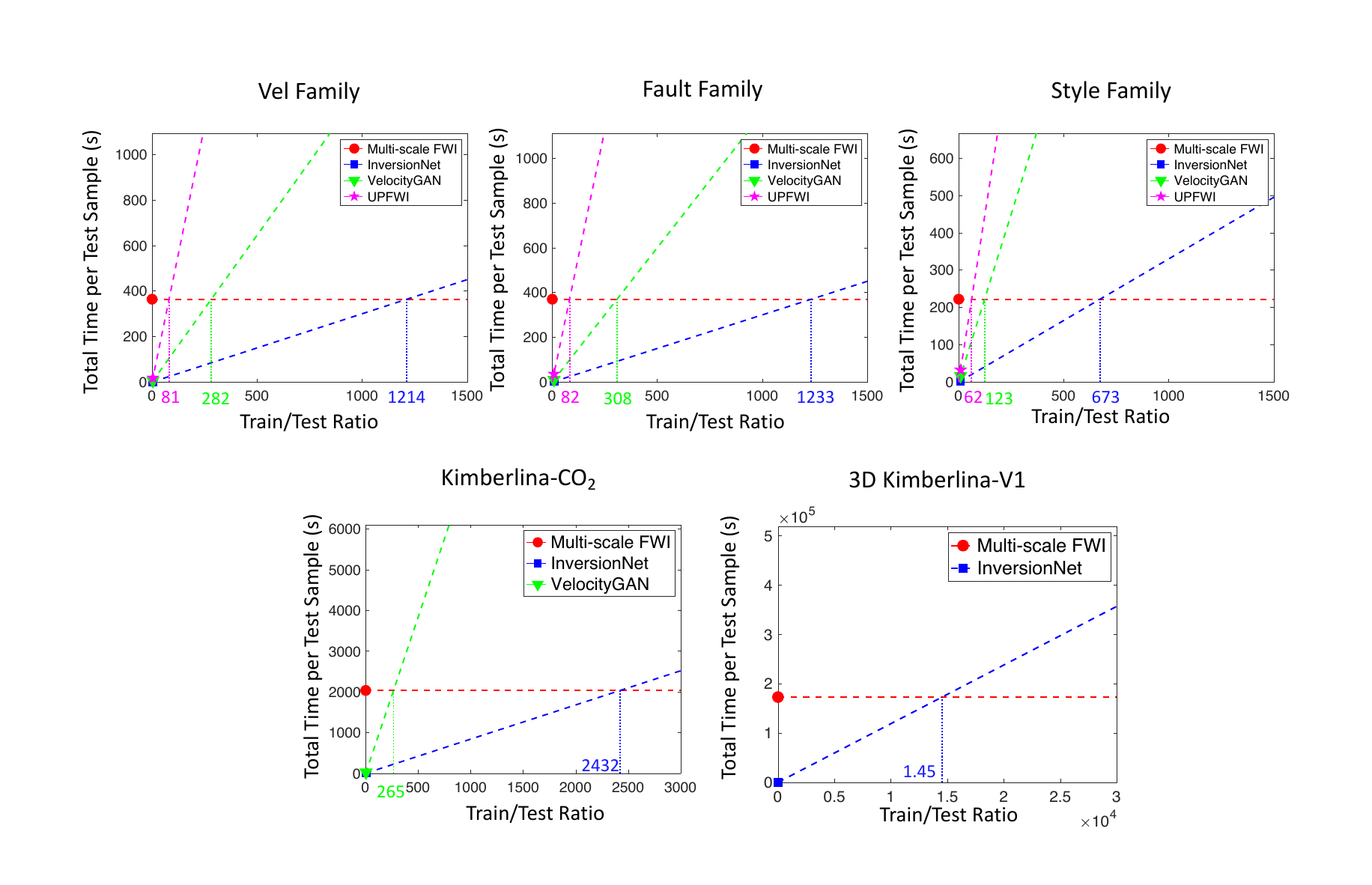}
\caption{ \textbf{The relationship between total computation time per test sample and train/test ratio}. The total computation time of data-driven methods is the summation of the training and test time.   The purple, green and blue numbers are the train/test ratio when the computation time of the data-driven methods (UPFWI, VelocityGAN and InversionNet) is equal to the physics-driven method (Multi-scale FWI). }
\label{fig:time_physics}
\end{figure*}

\textbf{Computational cost comparison:} The per-sample computation time of physics-driven methods and the average training time of data-driven methods for each sample is given in~\Cref{tab:time_data}.  Recall that all the experiments are implemented on NVIDIA Tesla P100 GPUs. Physic-driven methods and data-driven methods measure computational cost differently. Physics-driven methods do not require a training stage but directly optimize the testing data for multiple epochs to generate the inversion results, while data-driven methods need to be trained for several epochs and then applied to the test data for a one-time inference. Thus, the computational comparison varies over different ratios between the number of training and test samples. The relationship between total computation time per test sample and train/test ratio is given in~\Cref{fig:time_physics}. The total computation time of data-driven methods is the summation of the training and test time. The physics-driven methods do not need training, so the total computation time is their computation time on a test set only.

\begin{figure*}[h!]
\centering
\includegraphics[width=1\columnwidth]{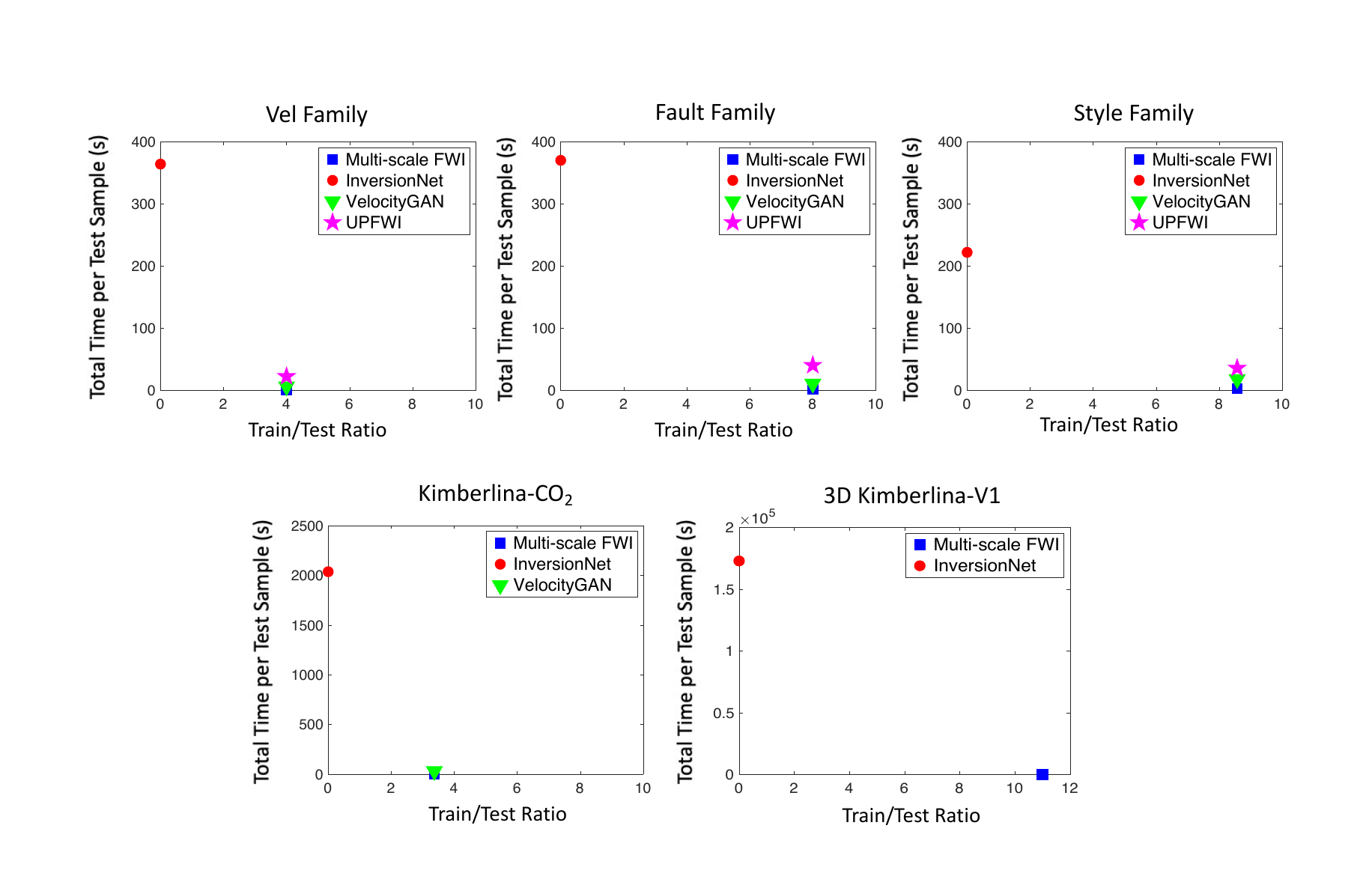}
\caption{\textbf{The comparison between total computation time per test sample of physics-driven methods and data-driven methods in~\textsc{OpenFWI} regarding to train/test ratio}. The computation time of data-driven methods (square, triangle and star markers) is much less than the physics-driven method (circle marker) in \textsc{OpenFWI} dataset.}
\label{fig:time_physics_zoom}
\end{figure*}

All three data-driven methods are faster when the ratio between the number of training and test samples is less than 62. Typically, the ratios for all twelve \textsc{OpenFWI} datasets are significantly lower (about 10) as in~\Cref{fig:time_physics_zoom}. Therefore, data-driven methods speed up 6 times more than physics-driven methods. 

 The time complexity of physics-driven methods is $O(N^3)$ for 2D cases and $O(N^4)$ for 3D cases. Their computational costs grow much faster than those of data-driven methods when the sizes of the velocity maps and seismic data increase. For example, the total computation time of the physics-driven method with 3D Kimberlina-V1 dataset is $475$ times more than that of Vel Family. For InversionNet with 3D Kimberlina-V1 dataset, the total computation time is only $40$ times more than that of Vel Family. In conclusion, data-driven methods have obvious computational advantages on large velocity maps.

\begin{table}
\footnotesize
\renewcommand{\arraystretch}{1.5}
\centering
\caption{\textbf{Quantitative results} of physics-driven FWI on~\textsc{OpenFWI} datasets with homogeneous, linear increasing, smoothed maps as initial maps. }
    \vspace{0.5em}
\begin{adjustbox}{width=1\textwidth}
\label{table:physicsFWI_Benchmark}
\begin{tabular}{c|cc|ccc|ccc|ccc} 
\thickhline
\multirow{2}{*}{Dataset}           & \multicolumn{2}{c|}{\multirow{2}{*}{Map}} & \multicolumn{3}{c}{Homogeneous Map}                                                  & \multicolumn{3}{c}{Linear Increasing Map}                                                & \multicolumn{3}{c}{Smoothed Map}                                   \\ 
\cline{4-12}
                                   & \multicolumn{2}{c|}{}                     & \multicolumn{1}{c}{MAE$\downarrow$} & RMSE$\downarrow$ & \multicolumn{1}{c|}{SSIM$\uparrow$} & \multicolumn{1}{c}{MAE$\downarrow$} & RMSE$\downarrow$ & \multicolumn{1}{c|}{SSIM$\uparrow$} & \multicolumn{1}{c}{MAE$\downarrow$} & RMSE$\downarrow$ & SSIM$\uparrow$   \\ 
\thickhline
\multirow{2}{*}{FlatVel-A}         & Initial &                                 & 0.5553                               & 0.7536           & 0.2249                             & 0.1841                               & 0.2438           & 0.4657                             & 0.0641                               & 0.1089           & 0.7087           \\
                                   &         & Result                          & 0.0643                               & 0.1318           & \textbf{0.8009}                    & 0.0426                               & 0.0768           & \textbf{0.8370}                    & 0.0324                               & 0.0621           & \textbf{0.8689}  \\ 
\hline
\multirow{2}{*}{FlatVel-B}         & Initial &                                 & 0.5222                               & 0.7186           & 0.1568                             & 0.4899                               & 0.6752           & 0.1547                             & 0.1752                               & 0.2634           & 0.4850           \\
                                   &         & Result                          & 0.1920                                    & 0.3712                & \textbf{0.5495}                         & 0.2112                                    & 0.4042                & \textbf{0.5293}                         & 0.0938                                    & 0.1902                & \textbf{0.6908}       \\ 
\hline
\multirow{2}{*}{CurveVel-A}        & Initial &                                 & 0.5599                               & 0.7578           & 0.1968                             & 0.2397                               & 0.3178           & 0.3415                             & 0.0871                               & 0.1405           & 0.6349           \\
                                   &         & Result                          & 0.1173                               & 0.2120           & \textbf{0.6447}                    & 0.0966                               & 0.1666           & \textbf{0.6654}                    & 0.0721                               & 0.1282           & \textbf{0.7347}  \\ 
\hline
\multirow{2}{*}{CurveVel-B}        & Initial &                                 & 0.5129                               & 0.7199           & 0.1528                             & 0.4901                               & 0.6699           & 0.1395                             & 0.2166                               & 0.3150           & 0.3991           \\
                                   &         & Result                          & 0.2465                               & 0.4084           & \textbf{0.4157}                    & 0.2670                               & 0.4402           & \textbf{0.3966}                    & 0.1734                               & 0.3020           & \textbf{0.5142}  \\ 
\hline
\multirow{2}{*}{FlatFault-A}       & Initial &                                 & 0.5123                               & 0.7737           & 0.1224                             & 0.2728                               & 0.3613           & 0.3309                             & 0.0619                               & 0.1543           & 0.7846           \\
                                   &         & Result                          & 0.0559                               & 0.1102           & \textbf{0.8222}                    & 0.0560                               & 0.1095           & \textbf{0.8244}                    & 0.0412                               & 0.0834           & \textbf{0.8607}  \\ 
\hline
\multirow{2}{*}{FlatFault-B}       & Initial &                                 & 0.8476                               & 1.0324           & 0.0028                             & 0.2881                               & 0.3682           & 0.2819                             & 0.1082                               & 0.1705           & 0.5790           \\
                                   &         & Result                          & 0.1442                               & 0.2366           & \textbf{0.5547}                    & 0.1247                               & 0.1992           & \textbf{0.5907}                    & 0.0961                               & 0.1599           & \textbf{0.6596}  \\ 
\hline
\multirow{2}{*}{CurveFault-A}      & Initial &                                 & 0.5064                               & 0.7567           & 0.1288                             & 0.2730                               & 0.3604           & 0.3381                             & 0.0655                               & 0.1533           & 0.7886           \\
                                   &         & Result                          & 0.0764                               & 0.1397           & \textbf{0.7652}                    & 0.0742                               & 0.1369           & \textbf{0.7716}                    & 0.0563                               & 0.1078           & 0.8190           \\ 
\hline
\multirow{2}{*}{CurveFault-B}      & Initial &                                 & 0.8493                               & 1.0297           & 0.0063                             & 0.3084                               & 0.3931           & 0.2186                             & 0.1484                               & 0.2191           & 0.4672           \\
                                   &         & Result                          & 0.1783                               & 0.2832           & \textbf{0.4837}                    & 0.1637                               & 0.2537           & \textbf{0.5078}                    & 0.1311                               & 0.2066           & \textbf{0.5765}  \\ 
\hline
\multirow{2}{*}{Style-A}           & Initial &                                 & 0.7571                               & 0.8812           & 0.1395                             & 0.2513                               & 0.3195           & 0.2881                             & 0.0775                               & 0.1042           & 0.6703           \\
                                   &         & Result                          & 0.1229                               & 0.2007           & \textbf{0.6018}                    & 0.0903                               & 0.1491           & \textbf{0.6866}                    & 0.0552                               & 0.0948           & \textbf{0.7986}  \\ 
\hline
\multirow{2}{*}{Style-B}           & Initial &                                 & 0.8165                               & 0.9219           & 0.0710                             & 0.1564                               & 0.1927           & 0.3578                             & 0.0910                               & 0.1080           & 0.4931           \\
                                   &         & Result                          & 0.1344                               & 0.1914           & \textbf{0.4599}                    & 0.0764                               & 0.1102           & \textbf{0.6521}                    & 0.0821                               & 0.1168           & \textbf{0.6132}  \\ 
\hline
\multirow{2}{*}{Kimberlina-CO$_2$} & Initial &                                 & 0.2033                                    & 0.2656                & 0.8198                                  & 0.0792                                    & 0.1512                & 0.8873                                  & 0.0689                                    & 0.1061                & 0.8924                \\
                                   &         & Result                         & 0.0434                                    & 0.0914                & \textbf{0.9137}                                  & 0.0419                                    & 0.0904                & \textbf{0.9125}                                  & 0.0404                                    & 0.0896                & \textbf{0.9122}                \\
 \hline
 \multirow{2}{*}{3D Kimberlina-V1} & Initial &                                 & N.A.                                    & N.A.                & N.A.                                  & N.A.                                    & N.A.                & N.A.                                  & 0.1174                                    & 0.2891                & \textbf{0.7975}                \\
                                   &         & Result                          & N.A.                                    & N.A.                & N.A.                         & N.A.                                    & N.A.                & N.A.                         & 0.5248                                    & 0.7581                & 0.5861       \\

\thickhline
\end{tabular}
\end{adjustbox}
\vspace{0.5em}
\end{table}

\begin{table}
\centering
\renewcommand{\arraystretch}{1.5}
\caption{\textbf{Pre sample computation time of physics-driven and data-driven FWI}.}
    \vspace{0.5em}
\begin{adjustbox}{width=1\textwidth}
\label{tab:time_data}
\begin{tabular}{c|c|ccccc} 
\thickhline
Computation Time                                                                           & Method          & Vel Family & Fault Family & Style Family & Kimberlina-CO$_2$ & 3D Kimberlina-V1  \\ 
\thickhline
Physics-driven FWI $t_{p}$                                                                    & Multi-scale FWI & 364s       & 370s         & 222s         & 2036s             & 48h               \\ 
\hline
\multirow{3}{*}{Data-driven FWI $t_{d}$}                                                      & InversionNet    & 0.3s       & 0.3s         & 0.33s        & 0.84s             & 11.90s            \\
                                                                                           & VelocityGAN     & 1.29s      & 1.2s         & 1.8s         & 7.68s             & N.A.              \\
                                                                                           & UPFWI           & 4.5s       & 4.5s         & 3.6s         & N.A.              & N.A.              \\ 
\thickhline
\end{tabular}
\end{adjustbox}
\end{table}

\section{Comparison between InversionNet and InversionNet3D}
\label{sup:inv2d_3d}
In this section, we compare the performance of the InversionNet when dealing with 3D data as 2D slices with the benchmark performance of InversionNet3D. Treating the Earth as 2D planes is a common way to deal with real data. To quantify the difference between 2D and 3D training strategies, we perform the following two 2D experiments and compare their results with the InversionNet3D benchmark results on the 3D Kimberlina-V1 dataset.
\newline
In the first experiment (3D simulation, 2D slices), we slice each ``velocity/seismic'' data sample in the 3D Kimberlina-V1 dataset into five vertical 2D slices along both inline ($X$) and cross-line ($Y$) directions according to the source locations. Thus, each 3D sample generates $10$ 2D velocity/seismic data pairs as training/validation samples.
In the second experiment (2D simulation), we slice each ``velocity'' sample in the 3D Kimberlina-V1 dataset into $40$ vertical 2D slices along the inline ($X$) direction according to the receiver locations. We then use the benchmark 2D finite difference forward operator to generate seismic data in the 2D planes and pair it with the corresponding 2D velocity slices as training/validation samples.
The sources and receivers are in the same 2D vertical planes. In total, $18$K 2D slice pairs are generated in the first experiment and $73$K are generated in the second experiment. The samples are used to train an InversionNet under a $90/10$ training/validation separation ratio. Once trained, we test the networks' performance and compare the results with the InversionNet3D benchmark. The performance comparison is listed in Table~\ref{tab:3dvs2d}. The table shows that the InversionNet performance is comparable to the InversionNet3D performance. For the 3D simulation 2D slices experiment, the MAE loss and the RMSE loss of InversionNet are about $20\%$ less than the ones of InversionNet3D using an $\ell_1$ training loss, and are about $50\%$ larger when using an $\ell_2$ training loss. The SSIM of InversionNet is $1.86\%$ lower than the one of InversionNet3D using an $\ell_1$ training loss, and is about $0.38\%$ higher when using an $\ell_2$ training loss. For the 2D simulation experiment, the MAE loss and the RMSE loss of InversionNet are about $68\%$ less than the ones of InversionNet3D using an $\ell_1$ training loss, and are about $87\%$ smaller when using an $\ell_2$ training loss. The SSIM of InversionNet is $0.79\%$ higher than the one of InversionNet3D using an $\ell_1$ training loss, and is about $4.55\%$ higher when using an $\ell_2$ training loss. 
\newline
To conclude, the performance of an end-to-end InversionNet that assumes the 3D Earth as 2D planes is comparable to that of the InversionNet3D. However, 3D training returns the whole volume of the region while 2D training returns only a few slices of the 3D volume. People may need further processing, for example, interpolation between slices, to obtain the whole 3D velocity volume.
\begin{table}
\centering
\renewcommand{\arraystretch}{1.2}
\caption{\textbf{Quantitative results} of two InversionNet training strategies compared with the averaged InversionNet3D benchmark on the 3D Kimberlina-V1 dataset.}
    \vspace{0.5em}
\begin{adjustbox}{width=0.8\textwidth}
\label{tab:3dvs2d}
\begin{tabular}{c|c|cccc} 
\thickhline
Network Architecture & Data Generator & MAE$\downarrow$ & RMSE$\downarrow$ & SSIM$\uparrow$  \\ 
\thickhline
\multirow{2}{*} {InversionNet-$\ell_1$}
    & 3D simulation, 2D slices & 0.0080	&	0.0215	& 0.9652 \\
    & 2D simulation & 0.0028 & 0.0093 & 0.9917 \\
\hline
\multirow{2}{*} {InversionNet-$\ell_2$}
    & 3D simulation, 2D slices & 0.0300	&	0.0401	& 0.9500 \\
    & 2D simulation & 0.0019 & 0.0042 & 0.9974 \\
\hline
InversionNet3D-$\ell_1$
    & 3D simulation & 0.0105 &		0.0276 &	0.9838 \\
\hline
InversionNet3D-$\ell_2$
    & 3D simulation & 0.0155 & 0.0306 & 0.9462 \\

\thickhline
\end{tabular}
\end{adjustbox}
\end{table}


\section{Test Strategy in Real-world Situation}
\label{sup:field}
The selection of the best inversion model is to minimize the discrepancy between the predicted velocity map and ground truth, which can be written as 
\begin{equation}
\underset{m}{\arg\min}~D(c_{pred},c_{true})
\label{eqn:selection_true}
\end{equation}
where $D$ is the discrepancy, $c_{pred}$ is the predicted velocity map given by inversion model ${m}$ and $c_{true}$ is the ground truth. The discrepancy $D$ may be obtained by calculation ($\ell_1$ or $\ell_2$ norm), or may be obtained from domain experts.

In the real-world case, $c_{true}$ may not be available . If we have a prior velocity map $c_{prior}$, we   will be able to calculate $\underset{m}{\arg\min} D(c_{pred},c_{prior})
$
to approximate~\Cref{eqn:selection_true}. However, the inversion models must be trained to give $c_{pred}$. In order to save computational cost, we can minimize the discrepancy between the training set and the prior velocity map:
\begin{equation}
\underset{m}{\arg\min}\sum_{train_i} D(c_{train_i},c_{prior})
\label{eqn:selection_train_prior}
\end{equation}
where $train_i$ is the index of the training samples corresponding to $m$. 

If there is no prior information, we would suggest using the seismic loss as the discrepancy, which is similar to physics-driven FWI:
\begin{equation}
\underset{m}{\arg\min} ||f(c_{pred})-d_{true}||^2
\label{eqn:selection_prior}
\end{equation}
where $f$ is the forward modeling operator and $d_{true}$ is the observed seismic data. Below is the demonstration of this scheme.

\begin{figure*}[h]
\centering
\includegraphics[width=1.0\textwidth]{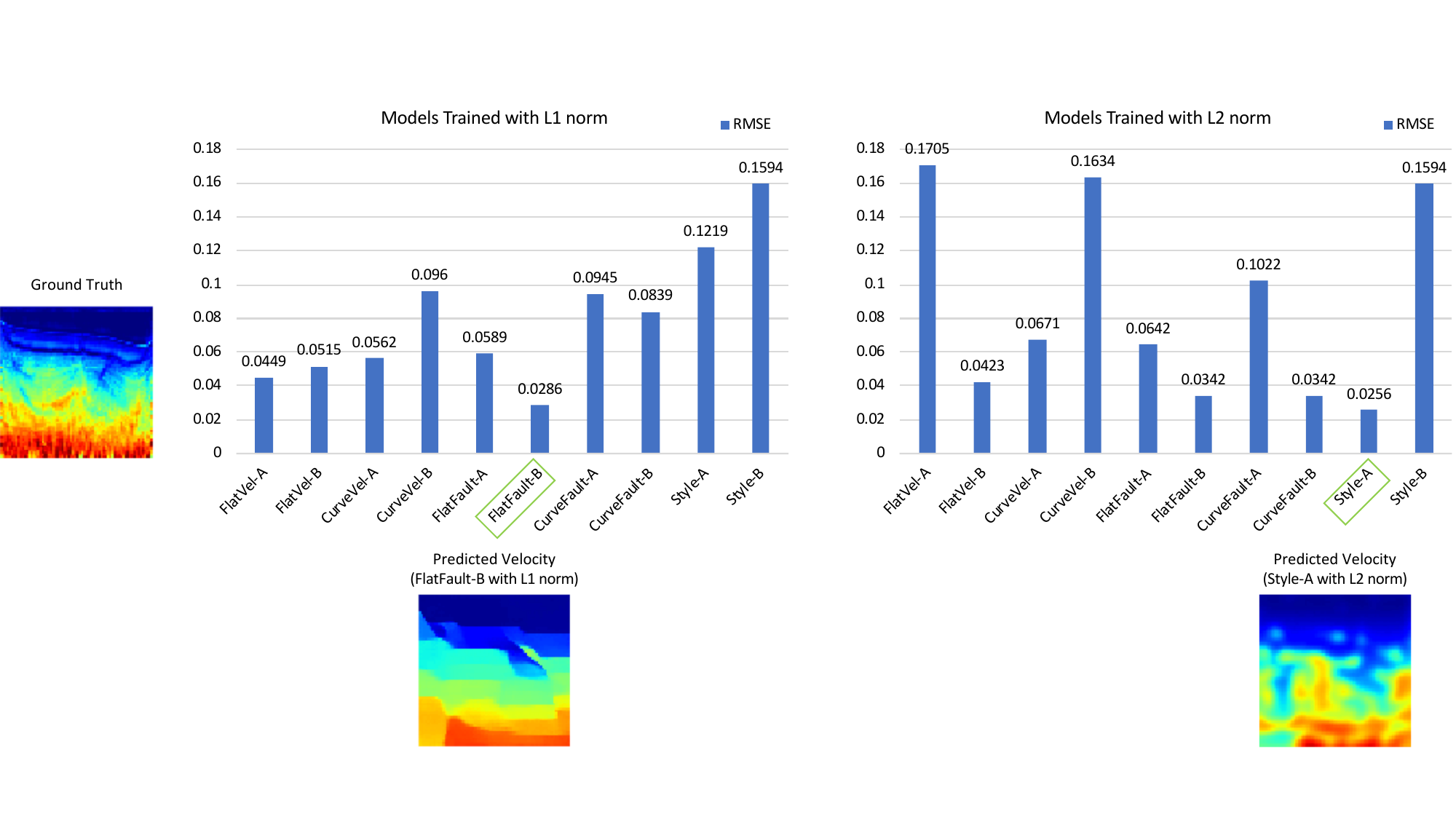}
\caption{\textbf{ Seismic data loss between predicted seismic data and pseudo-real data}. Among all the models trained with $\ell$1 norm and $\ell$2 norm, models trained with FlatFault-B and Style-A have the lowest RMSE, respectively. The predicted velocity maps with these two models are shown. }
\label{fig:real_strategy}
\end{figure*}

 Here we conduct an experiment to demonstrate how to choose a dataset for the target in the real scenario. We choose a real velocity map in the Gulf of Mexico~\cite{huang2018full} and simulate the seismic data as the pseudo-real data. Then we apply all twenty models trained across ten datasets (two models trained per dataset using $\ell_1$ and $\ell_2$ norm, respectively). We choose the two models with the minimal seismic loss. 
 The RMSE between the predicted seismic data and the pseudo-real data is given in~\Cref{fig:real_strategy}. The best model trained using $\ell_1$ is from the FlatVel-B dataset and the best model trained using $\ell_2$ is from the Style-A dataset.  Both models generate reasonably good velocity maps.

All of the methods mentioned above would require some additional effort to select the best datasets to use. If we would like to choose one with minimal effort, we would suggest using the "Style Family" in that this particular dataset yields highly diversified features obtained from natural images. That would enable the inversion of field data in general cases. You may choose version A or version B depending on the requirement of the velocity resolution.

\section{Passive Picking}
\label{sup:passive}
\begin{figure*}[h]
\centering
\includegraphics[width=1\columnwidth]{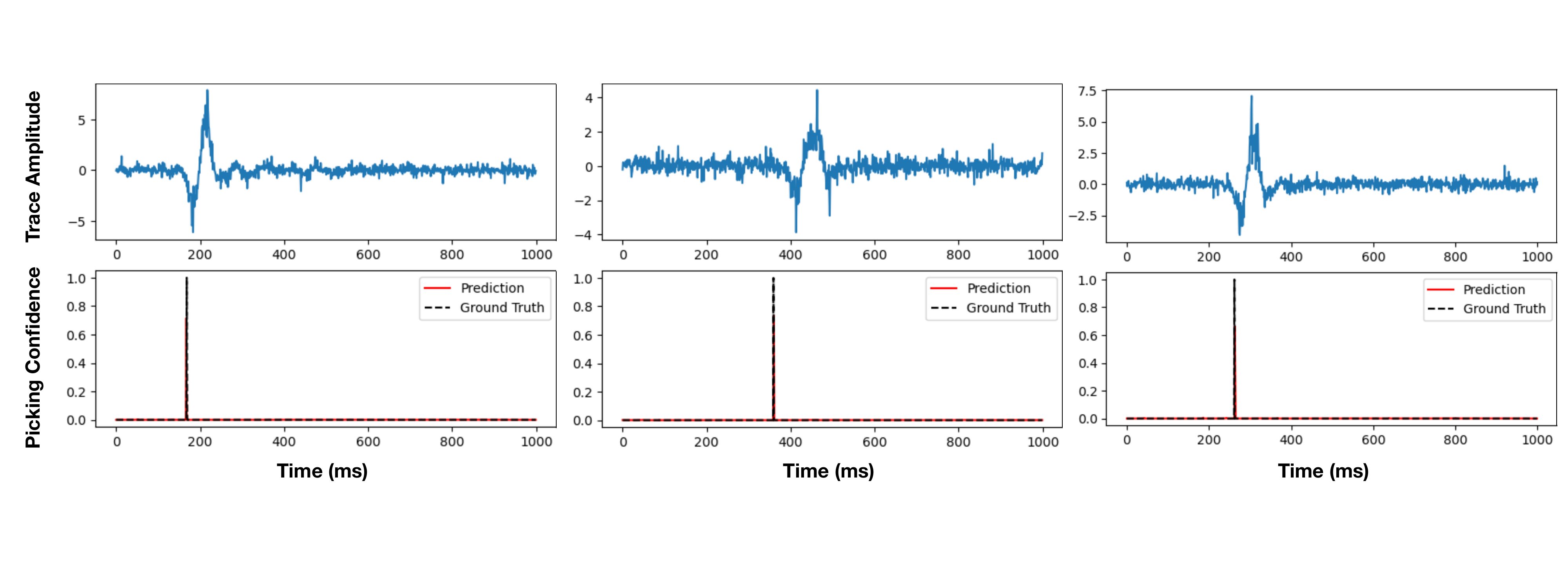}
\caption{\textbf{P-wave arrival pickings} by InversionNet trained on the Style-B dataset. Arrival labels manually generated and used in the training stage.}
\label{fig:picking}
\vspace{-1em}
\end{figure*}

 The current \textsc{OpenFWI} datasets mainly focus on geophysical imaging problems with active or controlled seismic sources, meaning the seismic sources are usually controlled explosions or generated by vibrators. However, there is another sub-field of geophysical imaging problems, passive seismic imaging, in which the source term is unknown. It is a common problem in global seismology, hydraulic fracturing monitoring, shale oil/gas exploration, CO$_2$ injection monitoring, and many other fields of study. The main objective is to find the passive seismic source information, including source spatial location, onset, and seismic moment tensor, when giving a trustful velocity model. A potential contribution of \textsc{OpenFWI} to passive seismic problems is using the provided large number of simulated traces to train a neural network, such as PhaseNet~\cite{zhu2019phasenet}, that can do arrival picking or event detection, which is an important step for passive seismic imaging problems. We take P-wave arrival picking as an example in this paper. We convert the Style-B dataset by adding (a) the labels of P-wave arrivals, and (b) strong Gaussian noise ($\sigma=5\times10^{-3}$) to the seismic traces. Then we train an InversionNet to predict the P-wave arrivals from noisy seismic traces. The corresponding test results are shown in~\Cref{fig:picking}, which indicates the InversionNet trained on the converted dataset accurately recognizes P-wave arrivals. Besides, people may also manually set the sources in the \textsc{OpenFWI} dataset as unknowns, though they are all located on the top surface, and invert the sources and velocity models in FWI schemes~\cite{wang2009simultaneous,wang2018microseismic} or by neural networks. In a future \textsc{OpenFWI} version, we may open-source passive seismic datasets and benchmarks upon data availability and approval of DOE and LANL.
\section{Discussion}
\label{sup:dis}

\subsection{Past Version}
We remark that several datasets with the same name have been used in previous publications. Specifically, FlatVel-A and CurveVel-A first emerged in \cite{wu2019inversionnet}; FlatFault-A and CurveFault-A were generated in \cite{Jin-2021-Unsupervised}; Style-A and Style-B were proposed in \cite{feng2021multiscale}. \textsc{OpenFWI} unified all datasets in the ``\textit{Vel}'', ``\textit{Fault}'' and the ``\textit{Style}'' families in terms of data size and forward modeling parameters, also training parameters for the benchmarking results. Therefore there is a slight difference from the previously reported experimental results. From \textsc{OpenFWI}, we will maintain the datasets as presented now, and the future comparison should also be conducted with \textsc{OpenFWI} benchmarks.

\subsection{Limitations}

\textbf{\textsc{OpenFWI} datasets: } All datasets are synthesized from only a few prior knowledge~(i.e., mathematical representations, natural images, or geological reservoir), therefore would inherently limit the representativeness and variability of the generated velocity maps. We also note that the ``\textit{Style Family}'' datasets are excellent candidates for the inversion of field data in general cases. However, there may be some specific subsurface structures that are not covered by \textsc{OpenFWI}. Additionally, it would be better if \textsc{OpenFWI} can be validated with some field data for further evaluation. \par 

\textbf{\textsc{OpenFWI} benchmarks: } There are two main limitations with the current benchmarks. One, the 3D FWI has limited literature and our benchmark is almost solitary. Furthermore, our evaluation strategy tests randomly selected channels, therefore is not ubiquitous.   The other concern is that data-driven FWI has been flourishing with new advancements, we may not be able to compare all other most recent methods due to the unavailability of the codes associated.

\end{document}